\documentclass[sigconf]{acmart}

\AtBeginDocument{%
  }

\copyrightyear{2026}
\acmYear{2026}
\setcopyright{cc}
\setcctype{by}
\acmConference[CIKM ’26]{Proceedings of the 35th ACM International Conference on Information and Knowledge Management}{November 07--11, 2026}{Rome, Italy}
\acmBooktitle{Proceedings of the 35th ACM International Conference on Information and Knowledge Management (CIKM ’26), November 07--11, 2026, Rome, Italy}
\acmDOI{10.1145/3799682.3841108}
\acmISBN{979-8-4007-2539-5/2026/11}

\usepackage{booktabs}
\usepackage{multirow}
\usepackage{adjustbox}
\usepackage{array}
\usepackage{algorithm}
\usepackage{algpseudocode}
\usepackage{amsmath}
\usepackage{graphicx}
\usepackage{enumitem}
\usepackage[utf8]{inputenc}
\usepackage{textgreek}
\usepackage{tabularx}
\usepackage{wrapfig}
\usepackage{subcaption}
\usepackage{pifont}
\usepackage{makecell}
\usepackage[table]{xcolor}
\usepackage{ragged2e}
\usepackage{siunitx}

\newlength\myindent
\usepackage{bm}

\def\vr{{\bm{r}}}

\def\vx{{\bm{x}}}

\usepackage{float}
\usepackage{seqsplit}
\usepackage{microtype}
\usepackage{array}
\usepackage{makecell}
\usepackage{xcolor}
\usepackage{adjustbox}

\newcolumntype{C}[1]{>{\centering\arraybackslash}p{#1}}

\newcommand{\spaced}[1]{\textls[40]{#1}}
\newcommand{\Yes}{\textcolor{green!60!black}{\ding{51}}}

\newcommand{\OApp}[2]{Appendix~\ref{#2}}

\renewcommand\footnotetextcopyrightpermission[1]{%
  \footnotetext{%
    This is an extended version of a paper accepted at CIKM~'26.
    The main text is identical to the camera-ready version, while Appendices~\ref{app:metrics}--\ref{app:addition_results} consolidate the appendix and online supplementary material into a single document.
    The paper will appear in the Proceedings of CIKM~'26. \url{https://doi.org/10.1145/3799682.3841108}.
    \copyright~2026 Copyright held by the owner/author(s).
    Published under a Creative Commons Attribution 4.0 International
    (CC~BY~4.0) license.}}

\begin{document}

\title{\textit{Seq2Synth}: Benchmarking Temporal Fidelity in Synthetic Sequential Tabular Data}


\author{Kiwan Kwon}
\authornote{Equal contributions.}
\email{kkgreenbean@unist.ac.kr}
\affiliation{%
  \institution{UNIST}
  \city{Ulsan}
  \country{Republic of Korea}
}

\author{Kangmin Kim}
\authornotemark[1]
\email{kmin1015@unist.ac.kr}
\affiliation{%
  \institution{UNIST}
  \city{Ulsan}
  \country{Republic of Korea}
}

\author{Hojin Lee}
\email{hojin0627@unist.ac.kr}
\affiliation{%
  \institution{UNIST}
  \city{Ulsan}
  \country{Republic of Korea}
}

\author{Yeseong Jung}
\email{y2ahsong@unist.ac.kr}
\affiliation{%
  \institution{UNIST}
  \city{Ulsan}
  \country{Republic of Korea}
}

\author{Hyeongwoo Kong}
\email{gho3283@hufs.ac.kr}
\affiliation{%
  \institution{HUFS}
  \city{Yongin}
  \country{Republic of Korea}
}

\author{Vamsi K. Potluru}
\authornote{The views and opinions expressed in this article are solely those of the author and do not necessarily reflect the official policy or position of JPMorgan Chase \& Co. or its affiliates. This research was conducted independently by the author outside of official work duties and without the use of firm resources.}
\email{ismav29@gmail.com}
\affiliation{%
  \institution{Independent Researcher}
  \city{New York}
  \state{NY}
  \country{USA}
}

\author{Saerom Park}
\authornote{Corresponding authors.}
\email{srompark@unist.ac.kr}
\affiliation{%
  \institution{UNIST}
  \city{Ulsan}
  \country{Republic of Korea}
}

\author{Yongjae Lee}
\authornotemark[3]
\email{yongjaelee@unist.ac.kr}

\affiliation{%
  \institution{UNIST}
  \city{Ulsan}
  \country{Republic of Korea}
}

\affiliation{%
  \institution{LinqAlpha}
  \city{New York}
  \state{NY}
  \country{USA}
}

\renewcommand{\shortauthors}{Kwon et al.}




\begin{abstract}

Synthetic sequential tabular data are increasingly used for privacy-preserving data sharing and research, yet conventional tabular metrics often overlook temporal structure. Existing single-table and relational evaluation protocols largely collapse records into static distributions, leaving key temporal properties insufficiently evaluated. We introduce \textit{Seq2Synth}, a unified benchmark for assessing these properties. Its taxonomy characterizes temporal and schema properties to determine applicable evaluations, covering timestamp, cross-sectional, longitudinal, and structural fidelity, alongside trajectory-aware utility and privacy. Across seven core datasets from a 13-dataset benchmark and eight generators, models with near-perfect static fidelity still violate basic temporal constraints, producing duplicate timestamps, irregular intervals, and incomplete observation grids. Moreover, static and temporal-aware rankings diverge substantially, showing that temporal fidelity must be evaluated directly rather than inferred from static or relational scores. Project page and online appendices are available at: \url{https://seq2synth.github.io/}.
\end{abstract}

\maketitle

\section{Introduction}
The demand for synthetic tabular data has grown rapidly in domains such as finance~\cite{syn_finance, syn_finance2} and healthcare~\cite{syn_healthcare}, driven by privacy regulations and limited access to sensitive records~\cite{syn_bias, syn_bias2}. Sequential tabular data, where timestamped records accumulate over time and are grouped by a persistent entity such as a customer, patient, or account~\cite{zhang2022sequential}, is in fact the dominant data format across many high-stakes domains. Transaction logs in finance, visit records in healthcare, usage histories in retail, and sensor streams in industrial systems all share this structure. Despite its ubiquity, the synthesis and evaluation of such data remain critical yet underexplored.

Existing evaluation frameworks overlook the temporal structure of such data. Single-table benchmarks~\cite{synthcity, syntheval} treat records as independent samples, while relational benchmarks~\cite{syntherela, relbench} evaluate inter-table dependencies but remain agnostic to temporal ordering and population dynamics. Prior work targeting sequential tabular data~\cite{zhang2022sequential, tabdit} relies on aggregate-level similarity, leaving cross-sectional dynamics and trajectory-level fidelity unaddressed.

\begin{figure*}[t]
    \centering
    \includegraphics[width=1.0\textwidth]{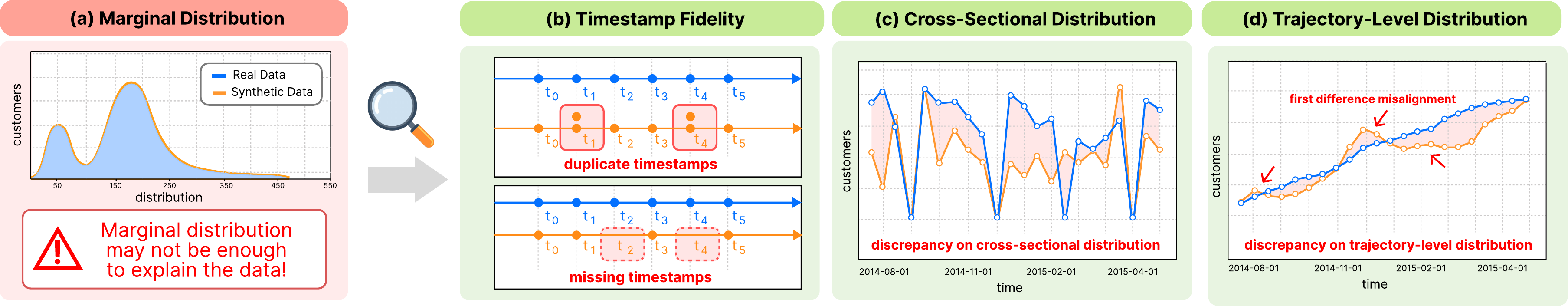}
    \caption{Motivation for temporal fidelity evaluation. (a) Synthetic data may appear similar to real data when evaluated only by marginal distributions. However, temporal views reveal mismatches in (b) timestamp fidelity, (c) cross-sectional distributions across time, and (d) trajectory-level evolution.}
    \Description{
    A four-panel motivation figure showing why marginal distribution matching is insufficient for evaluating temporal synthetic data.
    Panel (a) shows that real and synthetic data can have nearly overlapping marginal distributions, even though a warning box states that marginal distributions may not be enough to explain the data.
    Panel (b) illustrates timestamp fidelity problems, where synthetic sequences contain duplicate timestamps and missing timestamps compared with regular real timestamp sequences.
    Panel (c) compares real and synthetic customer values over calendar time and highlights discrepancies in cross-sectional distributions.
    Panel (d) compares real and synthetic trajectory-level customer trends over time and highlights first-difference misalignment and trajectory-level distribution discrepancies.
    Overall, the figure shows that temporal mismatches can remain hidden under marginal-distribution evaluation but become visible through timestamp, cross-sectional, and trajectory-level fidelity views.
    }
    \label{fig:motivation}
    \vspace{-4mm}  
\end{figure*}

Figure~\ref{fig:motivation} illustrates the consequences of this gap. ClavaDDPM~\cite{clavaddpm}, a state-of-the-art relational generative model produces synthetic records that violate chronological order and contain duplicate timestamps, which is incompatible with the constraints of real-world sequential data. However, they are not detected by existing benchmarks: the synthetic data attains near-perfect marginal fidelity (a), as conventional metrics collapse temporal structure into static distributions. The discrepancy emerges under timestamp (b), cross-sectional (c), and trajectory-level (d) inspection where the synthetic population fails to reflect real-world temporal dynamics. 


To address this limitation, we introduce \textit{Seq2Synth}, a unified benchmark for evaluating temporal fidelity in synthetic sequential tabular data. Static and relational evaluation can be insensitive to temporal validity: generators may match marginal or relational distributions while violating temporal constraints directly determined by the data. Temporal evaluation therefore provides more than an alternative model ranking: for datasets with explicit temporal constraints, properties such as timestamp uniqueness and regularity can be assessed against fixed reference values. Seq2Synth maps dataset characteristics to four applicable fidelity dimensions: timestamp, cross-sectional, longitudinal, and structural fidelity. We further extend standard utility and privacy protocols to trajectory-aware settings, evaluating downstream usefulness and leakage risk under sequential dependencies rather than row-level similarity alone.
Our contributions are as follows:
\begin{itemize}
\item We formulate a taxonomy-guided evaluation protocol for sequential tabular data, where time representation, sampling regularity, cross-trajectory dependence, and schema structure determine which temporal metrics are applicable.
\item We introduce \textit{Seq2Synth}, covering timestamp, cross-sectional, longitudinal, and structural fidelity, with trajectory-aware utility and privacy.
\item We build a 13-dataset benchmark across six domains and evaluate eight generators, showing that high static-distribution scores can mask temporal constraint violations, static and temporal-aware rankings diverge substantially, and architectures exhibit distinct temporal failure modes.
\end{itemize}
\section{Related Work}
\label{sec:related_work}

\paragraph{Synthetic Tabular Data Generation}
Tabular modeling spans heterogeneous numerical and categorical variables~\cite{hwang2025geodesic} and diverse generative approaches. SDV~\cite{sdv} and RCTGAN~\cite{rctgan}, used hierarchical or conditional generation, while ClavaDDPM~\cite{clavaddpm} incorporated inter-table relationships into diffusion. Recent graph-based methods, including RGCLD~\cite{rgcld} and RelDiff~\cite{reldiff}, use GNNs to model structural dependencies and improve
relational consistency. RDBDiff~\cite{rdbdiff} applies structure-aware diffusion to preserve intra-table distributions and inter-table consistency at scale. For sequential tabular data, CPAR~\cite{zhang2022sequential} in the SDV library~\cite{sdvlibrary} and TabularARGN~\cite{tabularargn} use
autoregressive modeling in single-table settings. REaLTabFormer (RTF)~\cite{rtf} employs a Seq2Seq-based framework for linear schemas linking parent entities to child trajectories. TabDiT~\cite{tabdit} adopts an autoregressive diffusion Transformer for time-dependent relational data and covers a broader range of sequential tabular structures.

\paragraph{Evaluation for Synthetic Data}


Various benchmark frameworks have been developed to assess synthetic data
quality across modalities. SynthCity~\cite{synthcity} and
SynthEval~\cite{syntheval} evaluate fidelity, utility, and privacy. For relational data, RelBench~\cite{relbench} evaluates utility by representing tabular records as graphs and applying graph neural networks to downstream prediction tasks, while SyntheRela~\cite{syntherela} measures relational fidelity by aggregating child-table features into parent tables and conducting classifier-based two-sample tests. The SDV library~\cite{sdv} provides metrics that capture step-wise distributional similarities in sequential data. Prior work also evaluates temporal properties beyond marginal similarity, including stylized facts~\cite{kwon2024can}, MLD-TS and MLE-TS~\cite{tabdit}, and successive-step coherence~\cite{tabularargn}. However, these evaluations remain fragmented and lack a systematic framework for assessing temporal characteristics across diverse sequential structures, leaving important failure modes in state-of-the-art models undetected.




\begin{figure*}[t]
    \centering
    \includegraphics[width=0.88\textwidth]{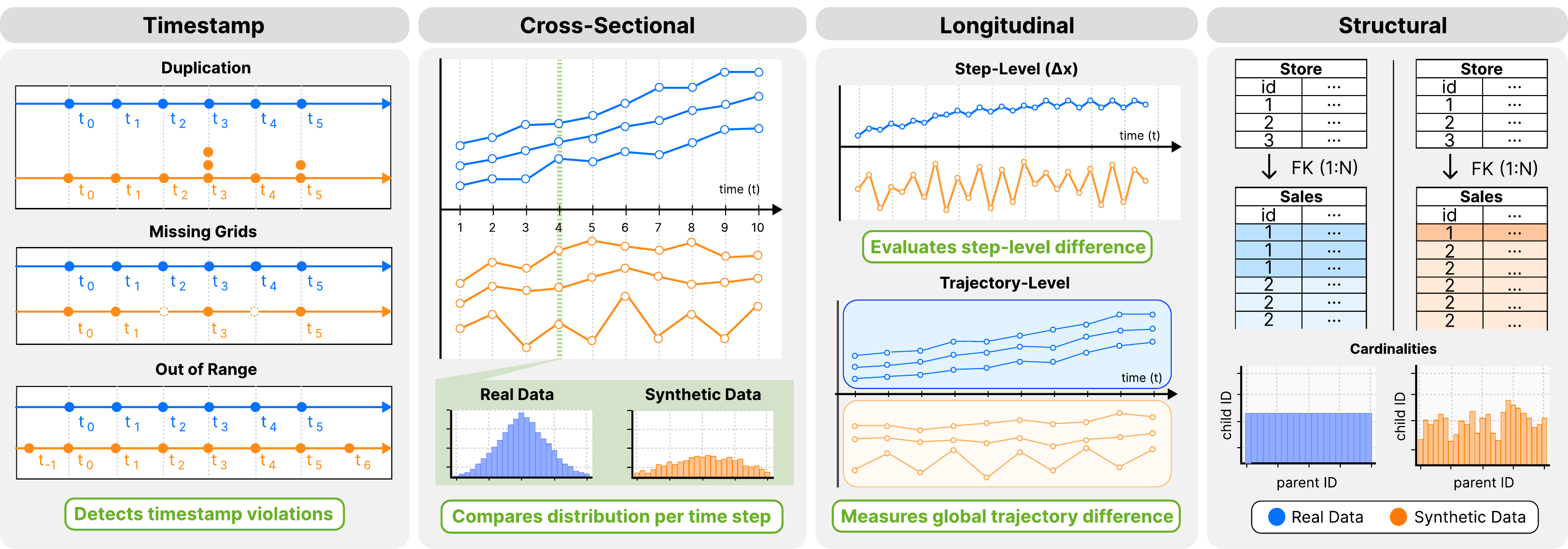}
    \vspace{-0.5em}
    \caption{Overview of proposed temporal fidelity evaluation metrics. \textbf{Timestamp Fidelity} detects temporal violations. \textbf{Cross-Sectional Fidelity} captures distribution differences at each time step. \textbf{Longitudinal Fidelity} compares step-level differences and global trajectory differences. \textbf{Structural Fidelity} checks similarity in table relationship. 
    }
    \vspace{-1em}
    \Description{An overview figure of the proposed temporal fidelity evaluation metrics, organized into four columns. The timestamp column illustrates three types of timestamp violations: duplicated timestamps, missing grid points, and timestamps outside the valid range. The cross-sectional column shows distributions compared at each time step, with real and synthetic values forming different per-time distributions. The longitudinal column shows two trajectory-based comparisons: step-level differences between consecutive time points and trajectory-level differences across entire sequences. The structural column shows parent-child table relationships and compares relational cardinality distributions between real and synthetic data.}
    \label{fig:metrics_overview}
\end{figure*}

\section{Seq2Synth Benchmark}
\label{sec:methodology}

\vspace{-1em}
\newcommand{\Part}{$\triangle$}
\newcommand{\No}{\ding{55}}

\begin{table}[htbp]
\centering
\caption{
Comparison of existing evaluation frameworks for synthetic tabular data.
\Yes: explicitly supported, \Part: partially or indirectly supported, \No: not explicitly addressed.
}
\vspace{-0.5em}
\Description{A comparison table of existing evaluation frameworks for synthetic tabular data. Rows indicate evaluation dimensions, including timestamp validity, cross-sectional dynamics, longitudinal trajectory dynamics, relational or structural fidelity, temporal utility, and temporal privacy. Columns group existing frameworks into single-table benchmarks, relational-table benchmarks, sequential tabular generative models, and the proposed framework. Most existing frameworks address only a subset of these dimensions, while the proposed framework supports all six.}
\label{tab:benchmark_comparison}
\scriptsize
\setlength{\tabcolsep}{3.0pt}
\renewcommand{\arraystretch}{1.08}
\begin{adjustbox}{max width=\columnwidth}
\begin{tabular}{
p{2.35cm}
ccc
cc
ccc
>{\columncolor{blue!7}}c
}
\toprule
\multirow{2}{*}{\textbf{Evaluation dimension}}
& \multicolumn{3}{c}{\textbf{Single-table}}
& \multicolumn{2}{c}{\textbf{Relational}}
& \multicolumn{3}{c}{\textbf{Sequential}}
& \multicolumn{1}{c}{\textbf{Ours}} \\
\cmidrule(lr){2-4}
\cmidrule(lr){5-6}
\cmidrule(lr){7-9}
& \makecell{\textbf{Synth}\\\textbf{City}\\[-0.15em]\cite{synthcity}}
& \makecell{\textbf{Synth}\\\textbf{Eval}\\[-0.15em]\cite{syntheval}}
& \makecell{\textbf{Syn}\\\textbf{Meter}\\[-0.15em]\cite{synmeter}}
& \makecell{\textbf{Rel}\\\textbf{Bench}\\[-0.15em]\cite{relbench}}
& \makecell{\textbf{Synthe}\\\textbf{Rela}\\[-0.15em]\cite{syntherela}}
& \makecell{\textbf{SDV}\\[-0.15em]\cite{sdvlibrary}}
& \makecell{\textbf{Tabular}\\\textbf{ARGN}\\[-0.15em]\cite{tabularargn}}
& \makecell{\textbf{Tab}\\\textbf{DiT}\\[-0.15em]\cite{tabdit}}
& \makecell{\textbf{Seq2}\\\textbf{Synth}} \\
\midrule

Timestamp validity
& \No & \No & \No
& \No & \No
& \No & \No & \No
& \textbf{\Yes} \\

Cross-sectional dynamics
& \No & \No & \No
& \No & \No
& \No & \No & \No
& \textbf{\Yes} \\

Trajectory dynamics
& \No & \No & \No
& \No & \No
& \Yes & \Part & \Yes
& \textbf{\Yes} \\

Structural fidelity
& \No & \No & \No
& \Yes & \Yes
& \Yes & \No & \No
& \textbf{\Yes} \\

Temporal utility
& \No & \No & \No
& \No & \No
& \No & \Yes & \Yes
& \textbf{\Yes} \\

Temporal privacy
& \No & \No & \No
& \No & \No
& \No & \No & \No
& \textbf{\Yes} \\

\bottomrule
\end{tabular}
\end{adjustbox}
\vspace{-1em}
\end{table}

To address the limitations of existing frameworks
(See Table~\ref{tab:benchmark_comparison}.), \seqsplit{Seq2Synth} evaluates synthetic
sequential tabular data through a taxonomy-guided protocol rather than a
fixed list of universally applied metrics. Given a real relational dataset
$\mathcal{T}$ and a synthetic counterpart $\hat{\mathcal{T}}$, the benchmark
first identifies the temporal and structural properties of the data, selects
the applicable evaluation dimensions, normalizes metric inputs when required,
and reports fidelity, utility, and privacy scores under this common protocol.

\subsection{Data Model and Taxonomy}
\label{subsec:data_model}

\textit{Sequential tabular data} consist of timestamped records that accumulate over time and are grouped by persistent entities. Let \(\mathcal{T} = \{T_1, \ldots, T_D\}\) denote a relational dataset with temporal and referential dependencies, and let \(T \in \mathcal{T}\) be a target table containing \(N\) timestamped records:
\[
T = \{(x^{key}_i,\, x^{time}_i,\, \vr_i)\}_{i=1}^{N},
\]
where \(x^{key}_i\) identifies the entity, \(x^{time}_i \in \Omega\) its
timestamp, and \(\vr_i = (\vx^{num}_i, \vx^{cat}_i)\) its numerical and
categorical attributes. For each entity \(k\), rows with \(x^{key}_i=k\),
ordered by timestamp, form a trajectory
\[
T(k) = \{(x^{time}_{k,j},\, \vr_{k,j})\}_{j=1}^{L_k},
\qquad x^{time}_{k,1} \leq \cdots \leq x^{time}_{k,L_k}.
\]

\paragraph{Taxonomy}
Seq2Synth classifies each dataset along four axes that determine metric
applicability:
(1) \textit{time representation}, absolute calendar or relative per-trajectory time,
(2) \textit{sampling regularity}, regular grids or irregular event intervals,
(3) \textit{trajectory dependence}, population-level dependent trajectories or isolated entities, and
(4) \textit{schema structure}, single, linear, multi-child, or multi-parent schemas.
\footnote{\textit{Single} schemas reside in one table where time-invariant attributes are implicit in the entity identifier, while \textit{linear} schemas separate them into one parent table linked to a child trajectory table.}
Table~\ref{tab:metric_applicability} summarizes the taxonomy-to-metric mapping and metric provenance (new, adapted, or adopted); formal taxonomy definitions are in Appendix~\ref{app:taxonomy}, while metric definitions and provenance details are in Appendices~\ref{app:metrics} and~\ref{app:metric_provenance}.

\subsection{Benchmark Protocol}
\label{sec:benchmark_protocol}


\paragraph{Step 1: classify the dataset}
The dataset is first assigned to the taxonomy above, which determines whether aligned population-level, grid-based, and relational-structure evaluations are applicable.

\paragraph{Step 2: select applicable dimensions}
The benchmark selects among four temporal fidelity dimensions.
\textbf{Timestamp} (\textsc{ts}) and \textbf{Longitudinal} fidelity (\textsc{lg}) apply broadly to the time axis and within-entity trajectories. \textbf{Cross-sectional} fidelity (\textsc{cs})
requires absolute time and dependent trajectories to compare population
distributions at each time point. \textbf{Structural} fidelity (\textsc{st}) depends on
the schema axis, with additional metrics for multi-child and multi-parent schemas.


\paragraph{Step 3: prepare metric inputs}
Timestamp metrics are computed on raw generator outputs so that timestamp validity violations remain fully exposed. Other metrics use post-processed inputs when needed, including sorting, grid alignment, and completion where applicable, to prevent basic timestamp-validity errors already captured by the timestamp dimension from mechanically propagating into other fidelity scores. Full pipeline details are in Appendix~\ref{app:online_postprocessing}, with imputation ablations in Appendix~\ref{app:postprocessing-ablation}.

\paragraph{Step 4: report dimension-level scores}

Scores are aggregated within each dimension using only metrics applicable to each dataset. Each dimension score is the mean of its applicable sub-metrics, with univariate and bivariate scores averaged over applicable features or feature pairs detailed in Appendix~\ref{app:feature_scope}. This aggregation avoids imposing an a priori importance ordering across sub-metrics, while their individual profiles provide diagnostic detail in Section~\ref{sec:results}.

\subsection{Temporal Fidelity Dimensions}
\label{sec:temp_fidelity}

The four fidelity dimensions are summarized in Figure~\ref{fig:metrics_overview}.

\paragraph{Timestamp fidelity}
Timestamp fidelity evaluates the generated time axis independently of feature
values. It checks temporal validity through
\texttt{Temporal Order Consistency}, \texttt{\seqsplit{Timestamp Uniqueness}}, and
\texttt{Temporal Range Compliance}; sampling behavior through
\texttt{Regularity Consistency} and \texttt{Grid Completeness} for regular
series or \texttt{Time Interval Distribution Similarity} for irregular series;
and trajectory span through \texttt{Trajectory Duration Similarity}. These
metrics are computed before post-processing. Full definitions are provided in
Appendix~\ref{app:timestamp_metrics}.

\paragraph{Cross-sectional fidelity}
Cross-sectional fidelity evaluates whether feature distributions are preserved
at aligned time points rather than only after pooling all records. It applies
when the dataset uses absolute time and contains dependent trajectories. We
partition the temporal domain into intervals and compute distributional
similarity within each interval. The instantiated metrics cover numerical
marginals (\texttt{CS-KSComplement}), categorical marginals
(\texttt{CS-TVComplement}), summary statistics, range and category coverage,
and bivariate dependencies through correlation and contingency similarity.
Full definitions are provided in Appendix~\ref{app:cross_sectional_metrics}.

\paragraph{Longitudinal fidelity}
Longitudinal fidelity evaluates whether within-entity dynamics are preserved
over time. Step-level metrics compare local changes using
\texttt{FirstDiffKSComplement} for numerical increments and
\texttt{TransMatrixTVDComplement} for categorical state transitions.
Trajectory-level metrics compare longer-range sequential structure using
\texttt{AutoCorrSimilarity}, \texttt{\seqsplit{MLD-TS}}~\citep{tabdit}, and
\texttt{TT-Wasserstein Distance}~\citep{ttwasserstein} where fixed-length trajectories are
available. Full definitions are provided in
Appendix~\ref{app:longitudinal_metrics}.

\paragraph{Structural fidelity}
Structural fidelity evaluates whether temporal relational structure is preserved.
\texttt{\seqsplit{Sequence Length Similarity}} applies to all schemas by comparing
trajectory-length distributions. Multi-child schemas add
\texttt{\seqsplit{Temporal Cardinality Shape Similarity}} and
\texttt{Dynamic K-Hop Correlation Similarity} to assess whether child-record
counts and cross-table correlations evolve correctly over time. Full definitions
are provided in Appendix~\ref{app:structural_metrics}.

\subsection{Trajectory-aware Utility and Privacy}
\label{sec:util_privacy}

\paragraph{Utility}
Fidelity scores alone do not determine whether synthetic data supports
downstream analysis. Seq2Synth adopts Machine Learning Efficiency, training
models on synthetic data and testing them on real data. We evaluate
trajectory-level utility with \texttt{MLE-TS}~\citep{tabdit} and report \texttt{MLE-Temporal}, which includes historical context in the input. Details are provided in Appendix~\ref{app:utility_metric}.

\paragraph{Privacy}
Standard row-level privacy metrics such as
\texttt{\seqsplit{DCR}} and \texttt{\seqsplit{NNDR}}~\citep{syntheval,synmeter} ignore whether a
synthetic record matches a real record at the same time point or within the
same trajectory. Seq2Synth adds cross-sectionally restricted nearest-neighbor metrics, \texttt{\seqsplit{CS-DCR}} and \texttt{\seqsplit{CS-NNDR}}, and an $n$-gram privacy score, \(\texttt{\seqsplit{NGP}}(n)=1-\texttt{\seqsplit{NgramExposure}}(n)\). Full definitions are provided in Appendix~\ref{app:privacy_metrics}.


\section{Experiments}
\label{sec:experiments}

\subsection{Experimental Setup}
\label{sec:setup}

We evaluate 8 models on 13 datasets spanning six domains described in Table~\ref{tab:dataset_taxonomy}. We designate 7 core datasets, each supported by at least six of the seven broadly applicable baselines, enabling cross-model comparison. The models are ClavaDDPM~\cite{clavaddpm}, RCTGAN~\cite{rctgan}, RDBDiff~\cite{rdbdiff}, RTF~\cite{rtf}, RGCLD~\cite{rgcld}, RelDiff~\cite{reldiff}, SDV~\cite{sdv}, and TabDiT~\cite{tabdit}.  Static-distribution (\textsc{sd}) and \textsc{ts} use raw outputs, \textsc{cs} and \textsc{st} the sparse variant, and \textsc{lg} interpolation-fill when complete trajectories are required.


\subsection{Results}
\label{sec:results}

\begin{table}[thbp]
\centering
\caption{Taxonomy of sequential tabular datasets. Abs./Rel., Reg./Irreg., and Dep./Indep. denote time representation, periodicity, and trajectory independence. M-Child, M-Parent, and M-C\&P denote schema types. The first seven rows are the core set.}

\Description{A taxonomy table of thirteen sequential tabular datasets used in the benchmark. Each row lists a dataset and characterizes it by domain, time representation, periodicity, trajectory independence, and data structure. The first seven rows form the core evaluation set used for main cross-model comparison. The datasets span retail, healthcare, finance, industrial, transportation, and IT systems domains, and include both absolute and relative time representations, regular and irregular periodicity, dependent and independent trajectories, and linear, multi-child, multi-parent, and single-table structures.}
\label{tab:dataset_taxonomy}

\vspace{-0.8em}
\small
\setlength{\tabcolsep}{3.2pt}
\renewcommand{\arraystretch}{1.04}

\begin{tabular*}{\linewidth}{@{\extracolsep{\fill}}llcccc@{}}
\toprule
\textbf{Dataset} 
& \textbf{Domain} 
& \makecell{\textbf{Time}\\\textbf{Rep.}} 
& \textbf{Period.} 
& \makecell{\textbf{Traj.}\\\textbf{Indep.}} 
& \makecell{\textbf{Data}\\\textbf{Struct.}} \\
\midrule

\textbf{Rossmann}~\cite{tabdit}       & Retail    & Abs. & Reg.   & Dep.   & Linear   \\
\textbf{Berka}~\cite{tabdit}         & Finance   & Abs. & Irreg. & Dep.   & Linear   \\
\textbf{Fannie Mae} ~\cite{fanniemae}
     & Finance   & Abs. & Reg.   & Dep.   & Linear   \\
\textbf{Walmart}~\cite{syntherela}        & Retail    & Abs. & Reg.   & Dep.   & M-Child  \\
\textbf{AirBnB}~\cite{tabdit}          & Retail    & Rel. & Irreg. & Dep.   & Linear   \\
\textbf{PTB-XL}~\cite{ptbxl}         & Health    & Rel. & Reg.   & Indep. & Linear   \\
\textbf{Freddie Mac}~\cite{freddiemac}    & Finance   & Abs. & Reg.   & Dep.   & Linear   \\
\midrule
\textbf{Citi Bike}~\cite{citibike_nyc}      & Transport & Abs. & Irreg. & Dep.   & Single   \\
\textbf{CMAPSS}~\cite{cmapss}         & Industry  & Rel. & Reg.   & Indep. & Single   \\
\textbf{Coupon}~\cite{coupon}         & Retail    & Abs. & Irreg. & Dep.   & M-C\&P   \\
\textbf{Google Cluster}~\cite{googlecluster}  & IT        & Rel. & Both   & Indep. & M-Child  \\
\textbf{H\&M}~\cite{hm_kaggle}           & Retail    & Abs. & Irreg. & Dep.   & M-Parent \\
\textbf{Home Credit}~\cite{homecredit}    & Finance   & Rel. & Reg.   & Dep.   & M-Child  \\

\bottomrule
\end{tabular*}

\vspace{-1.2em}
\end{table}

We organize the results around three empirical findings following the benchmark
design in Section~\ref{sec:benchmark_protocol}. First, the temporal
dimensions are not interchangeable with static-distribution evaluation.
Model rankings change substantially when timestamp, longitudinal, and structural
fidelity are measured directly. Second, the failures are architecture-coherent.
Autoregressive models tend to preserve within-trajectory order but struggle with
relational cardinality. Joint and relational generators better preserve some
schema-level structure but often fail on timestamp validity and trajectory
dynamics. Third, temporal evaluation changes the interpretation of both privacy
and downstream utility. Appendix~\ref{app:addition_results} reports complete
sub-metric tables for the seven core datasets.


\label{sec:overview}

\begin{table*}[t]
\centering
\caption{Per-dataset fidelity scores on the seven core datasets across five evaluation dimensions.
Higher is better; ``--'' denotes inapplicable cases. Bold and underline indicate the
best and second-best scores per dataset--metric row.}
\label{tab:per-dataset}

\footnotesize
\setlength{\tabcolsep}{2.6pt}
\renewcommand{\arraystretch}{0.9}

\begin{tabular*}{0.98\textwidth}{@{\extracolsep{\fill}}llcccccccc@{}}
\toprule
\textbf{Dataset}
& \textbf{Metric}
& \textbf{ClavaDDPM}
& \textbf{RCTGAN}
& \textbf{RDBDiff}
& \textbf{RTF}
& \textbf{RGCLD}
& \textbf{RelDiff}
& \textbf{SDV}
& \textbf{TabDiT} \\
\midrule

\multirow[c]{5}{*}{\textbf{Rossmann}}
& \textsc{sd} & 0.850 & 0.810 & $\underline{0.858}$ & 0.809 & 0.821 & 0.799 & 0.734 & $\mathbf{0.932}$ \\
& \textsc{ts} & 0.607 & 0.562 & 0.608 & $\underline{0.823}$ & 0.598 & 0.602 & 0.565 & $\mathbf{0.824}$ \\
& \textsc{cs} & $\underline{0.912}$ & 0.808 & $\underline{0.912}$ & 0.882 & 0.897 & 0.817 & 0.635 & $\mathbf{0.965}$ \\
& \textsc{lg} & 0.917 & 0.872 & 0.915 & $\underline{0.948}$ & 0.902 & 0.882 & 0.789 & $\mathbf{0.975}$ \\
& \textsc{st} & $\mathbf{1.000}$ & $\mathbf{1.000}$ & $\mathbf{1.000}$ & $\mathbf{1.000}$ & $\mathbf{1.000}$ & $\mathbf{1.000}$ & $\mathbf{1.000}$ & $\mathbf{1.000}$ \\

\midrule

\multirow[c]{5}{*}{\textbf{Berka}}
& \textsc{sd} & 0.948 & 0.865 & $\mathbf{0.992}$ & $\underline{0.951}$ & 0.922 & 0.898 & 0.695 & 0.909 \\
& \textsc{ts} & 0.594 & 0.620 & 0.759 & $\mathbf{0.907}$ & 0.731 & 0.603 & 0.535 & $\underline{0.777}$ \\
& \textsc{cs} & $\underline{0.691}$ & 0.632 & $\mathbf{0.829}$ & 0.665 & 0.671 & 0.626 & 0.448 & 0.651 \\
& \textsc{lg} & 0.877 & 0.858 & $\mathbf{0.949}$ & $\underline{0.896}$ & 0.885 & 0.868 & 0.686 & 0.890 \\
& \textsc{st} & 0.993 & 0.952 & $\mathbf{1.000}$ & $\mathbf{1.000}$ & 0.993 & 0.952 & 0.814 & 0.958\\

\midrule

\multirow[c]{5}{*}{\textbf{Fannie Mae}}
& \textsc{sd} & $\underline{0.900}$ & 0.860 & 0.876 & 0.854 & $\mathbf{0.906}$ & 0.899 & 0.814 & -- \\
& \textsc{ts} & 0.567 & 0.550 & $\underline{0.571}$ & 0.474 & 0.503 & $\mathbf{0.598}$ & 0.481 & -- \\
& \textsc{cs} & $\mathbf{0.965}$ & 0.852 & 0.870 & 0.777 & $\underline{0.876}$ & 0.849 & 0.772 & -- \\
& \textsc{lg} & 0.820 & 0.787 & 0.781 & 0.804 & $\underline{0.858}$ & 0.822 & $\mathbf{0.864}$ & -- \\
& \textsc{st} & $\mathbf{0.998}$ & 0.953 & 0.822 & 0.817 & $\underline{0.990}$ & 0.750 & 0.848 & -- \\

\midrule

\multirow[c]{5}{*}{\textbf{Walmart}}
& \textsc{sd} & 0.796 & 0.861 & 0.815 & 0.719 & $\mathbf{0.950}$ & $\underline{0.898}$ & 0.857 & -- \\
& \textsc{ts} & $\mathbf{0.610}$ & 0.526 & $\underline{0.608}$ & 0.455 & 0.600 & 0.598 & 0.522 & -- \\
& \textsc{cs} & 0.867 & 0.785 & $\mathbf{0.896}$ & 0.605 & $\underline{0.891}$ & 0.828 & 0.746 & -- \\
& \textsc{lg} & $\mathbf{0.887}$ & 0.844 & 0.868 & 0.690 & $\underline{0.872}$ & 0.852 & 0.814 & -- \\
& \textsc{st} & 0.864 & 0.731 & $\mathbf{0.887}$ & 0.518 & $\underline{0.882}$ & 0.804 & 0.786 & -- \\

\midrule

\multirow[c]{5}{*}{\textbf{Airbnb}}
& \textsc{sd} & $\mathbf{0.967}$ & 0.909 & $\underline{0.963}$ & 0.879 & 0.925 & 0.937 & 0.711 & 0.923 \\
& \textsc{ts} & $\mathbf{0.875}$ & 0.840 & 0.824 & 0.777 & 0.843 & $\underline{0.860}$ & 0.811 & $\underline{0.860}$ \\
& \textsc{cs} & -- & -- & -- & -- & -- & -- & -- & -- \\
& \textsc{lg} & 0.964 & $\mathbf{0.970}$ & 0.965 & 0.967 & $\underline{0.969}$ & $\underline{0.969}$ & 0.909 & 0.965 \\
& \textsc{st} & 0.987 & 0.924 & $\mathbf{1.000}$ & 0.631 & 0.981 & $\mathbf{1.000}$ & 0.938 & 0.912 \\

\midrule

\multirow[c]{5}{*}{\textbf{PTB-XL}}
& \textsc{sd} & $\underline{0.886}$ & 0.576 & $\mathbf{0.896}$ & 0.625 & 0.692 & 0.669 & 0.758 & -- \\
& \textsc{ts} & 0.594 & 0.330 & 0.577 & $\mathbf{1.000}$ & 0.534 & 0.591 & $\underline{0.675}$ & -- \\
& \textsc{cs} & -- & -- & -- & -- & -- & -- & -- & -- \\
& \textsc{lg} & $\mathbf{0.927}$ & 0.841 & 0.909 & 0.915 & 0.906 & $\underline{0.922}$ & 0.920 & -- \\
& \textsc{st} & $\mathbf{1.000}$ & $\mathbf{1.000}$ & $\mathbf{1.000}$ & $0.999$ & $\mathbf{1.000}$ & $\mathbf{1.000}$ & $\mathbf{1.000}$ & -- \\

\midrule

\multirow[c]{5}{*}{\textbf{Freddie Mac}}
& \textsc{sd} & 0.944 & 0.882 & $\mathbf{0.986}$ & 0.853 & $\underline{0.975}$ & 0.920 & 0.810 & -- \\
& \textsc{ts} & 0.713 & 0.688 & 0.660 & 0.581 & 0.736 & $\mathbf{0.752}$ & $\underline{0.747}$ & -- \\
& \textsc{cs} & 0.924 & 0.843 & $\mathbf{0.982}$ & 0.777 & $\underline{0.930}$ & 0.847 & 0.740 & -- \\
& \textsc{lg} & 0.842 & 0.797 & $\mathbf{0.915}$ & 0.784 & $\underline{0.875}$ & 0.841 & 0.780 & -- \\
& \textsc{st} & 0.995 & 0.972 & $\mathbf{1.000}$ & 0.695 & 0.993 & $\mathbf{1.000}$ & 0.965 & -- \\

\bottomrule
\end{tabular*}

\vspace{-1.5mm}
\end{table*}

\paragraph{Finding 1: Temporal-aware dimensions are not interchangeable with static-distribution evaluation.} Table~\ref{tab:per-dataset} reports fidelity scores for the 7 core datasets under \textsc{sd} from prior benchmarks~\cite{synthcity,syntherela} and our four temporal fidelity dimensions. The results show that fidelity is strongly dimension-dependent: models with high \textsc{sd} scores do not consistently dominate the temporal-aware dimensions, and the best-performing model often changes across \textsc{ts}, \textsc{lg}, and \textsc{st}. Based on this observation, we analyze rank disagreement between static-distribution and temporal evaluation.

\begin{table}[ht]
\centering
\caption{Rank disagreement between \textsc{sd} and temporal dimensions.
Spearman $\rho$ and inversion rates are computed across models within each
dataset; pooled $\rho$ concatenates within-dataset ranks. ``--'' denotes
inapplicable.}
\Description{A table reporting rank disagreement between static-distribution evaluation and temporal-aware fidelity dimensions. For each dataset, the table compares static-distribution rankings against timestamp, cross-sectional, longitudinal, and structural rankings using Spearman correlation and the fraction of inverted model pairs. Lower or negative correlations and higher inversion fractions indicate stronger disagreement between static and temporal-aware evaluations. The table shows that static-distribution rankings often disagree with temporal-aware rankings, especially for timestamp and longitudinal dimensions, supporting the need for separate temporal fidelity evaluation.}
\label{tab:rank_disagreement}

\vspace{-0.1em}
\small
\setlength{\tabcolsep}{2.4pt}
\renewcommand{\arraystretch}{1.02}

\begin{tabular*}{\linewidth}{@{\extracolsep{\fill}}lcccccccc@{}}
\toprule
\multirow{2}{*}{\textbf{Dataset}}
 & \multicolumn{2}{c}{\footnotesize\textsc{sd} vs \textsc{ts}}
 & \multicolumn{2}{c}{\footnotesize\textsc{sd} vs \textsc{cs}}
 & \multicolumn{2}{c}{\footnotesize\textsc{sd} vs \textsc{lg}}
 & \multicolumn{2}{c}{\footnotesize\textsc{sd} vs \textsc{st}} \\
\cmidrule(lr){2-3}
\cmidrule(lr){4-5}
\cmidrule(lr){6-7}
\cmidrule(lr){8-9}
 & $\rho$ & Inv. & $\rho$ & Inv. & $\rho$ & Inv. & $\rho$ & Inv. \\
\midrule
\textbf{Airbnb}       &  $0.714$ & $0.21$ & --       & --     & $-0.071$ & $0.57$ & $0.755$ & $0.22$ \\
\textbf{Berka}        &  $0.595$ & $0.29$ & $0.905$  & $0.11$ &  $0.905$ & $0.11$ & $0.497$ & $0.28$ \\
\textbf{Fannie Mae}   &  $0.500$ & $0.33$ & $0.857$  & $0.14$ &  $0.071$ & $0.48$ & $0.429$ & $0.38$ \\
\textbf{Freddie Mac}  & $-0.107$ & $0.52$ & $1.000$  & $0.00$ &  $1.000$ & $0.00$ & $0.775$ & $0.20$ \\
\textbf{PTB-XL}       &  $0.143$ & $0.48$ & --       & --     &  $0.393$ & $0.33$ & $0.408$ & $0.17$ \\
\textbf{Rossmann}     &  $0.571$ & $0.25$ & $0.929$  & $0.07$ &  $0.690$ & $0.21$ & --      & --     \\
\textbf{Walmart}      &  $0.107$ & $0.43$ & $0.321$  & $0.33$ &  $0.250$ & $0.38$ & $0.286$ & $0.38$ \\
\midrule
\textit{Mean (per-ds)} & $0.360$ & $0.36$ & $0.802$ & $0.13$ & $0.463$ & $0.30$ & $0.525$ & $0.27$ \\
\textit{Pooled}        & $0.403$ & --     & $0.819$ & --     & $0.477$ & --     & $0.497$ & --     \\
\bottomrule
\end{tabular*}

\vspace{-1.4em}
\end{table}

Table~\ref{tab:rank_disagreement} shows that \textsc{sd}-based
evaluation is a poor proxy for temporal fidelity. Across the seven core
datasets, the rankings induced by timestamp, longitudinal, and structural
fidelity diverge substantially from the \textsc{sd}-based ranking, with
mean Spearman correlations of $0.36$, $0.46$, and $0.53$, respectively.
The inversion rates are high: $36\%$, $30\%$, and $27\%$ of model pairs reverse their ordering relative to \textsc{sd}. These disagreements show that model selection based on static-distribution fidelity does not reliably transfer to temporal fidelity, motivating direct evaluation of temporal properties.

\textbf{Cross-sectional} fidelity is the only dimension that remains
closely aligned with \textsc{sd} evaluation, with pooled $\rho=0.82$ and a
mean inversion rate of $13\%$. This alignment arises because \textsc{cs} evaluates distributional similarity within individual time slices and aggregates the slice-wise scores over time. Therefore, cross-sectional fidelity largely preserves the global ranking captured by static-distribution evaluation while diagnosing whether
distributional fidelity remains stable over time. A model can achieve a strong pooled \textsc{sd} score while still failing on specific time slices, and the cross-sectional
dimension is designed to expose such temporally localized distributional
failures. 
Rank disagreement shows that the dimensions are not interchangeable, but does not imply that temporal-aware rankings are inherently more correct, since
different metric families capture different properties. We therefore turn to
failure modes whose reference properties are determined by the real data and
examine timestamp, cross-sectional, longitudinal, and structural failure modes.



\vspace{-0.5em}
\paragraph{Finding 2: Temporal sub-metrics expose model-specific failures.}
\label{sec:timestamp}

\begin{table}[ht]
\centering
\caption{Timestamp sub-metric scores on Rossmann before post-processing. All scores lie in $[0,1]$, with higher values indicating better timestamp fidelity. Bold and underline mark the best and second-best scores per column.}
\Description{A results table reporting timestamp sub-metric scores on the Rossmann dataset before post-processing. Rows compare synthetic data generation models, and columns report timestamp order consistency, timestamp uniqueness, duration similarity, regularity consistency, and grid completeness. Scores range from zero to one, with higher values indicating better timestamp fidelity.}
\label{tab:ts-rossmann}

\vspace{-0.8em}
\small
\setlength{\tabcolsep}{2.8pt}
\renewcommand{\arraystretch}{1.02}

\begin{tabular*}{\linewidth}{@{\extracolsep{\fill}}lccccc@{}}
\toprule
\textbf{\spaced{Model}}
  & {\footnotesize\spaced{\texttt{Order}}}
  & {\footnotesize\spaced{\texttt{Unique}}}
  & {\footnotesize\spaced{\texttt{Duration}}}
  & {\footnotesize\spaced{\texttt{Regularity}}}
  & {\footnotesize\spaced{\texttt{Grid}}} \\
\midrule
\textbf{\spaced{ClavaDDPM}} & $0.507$ & $0.629$ & $0.462$ & $0.402$ & $0.644$ \\
\textbf{\spaced{RCTGAN}}    & $\mathbf{0.512}$ & $0.594$ & $0.287$ & $0.372$ & $0.611$ \\
\textbf{\spaced{RDBDiff}}   & $\underline{0.510}$ & $0.628$ & $0.466$ & $0.402$ & $0.644$ \\
\textbf{\spaced{RTF}}       & $0.001$ & $\mathbf{0.999}$ & $\underline{0.942}$ & $\mathbf{0.997}$ & $\mathbf{0.999}$ \\
\textbf{\spaced{RGCLD}}     & $0.507$ & $0.619$ & $0.435$ & $0.392$ & $0.636$ \\
\textbf{\spaced{RelDiff}}   & $0.509$ & $0.616$ & $0.507$ & $0.369$ & $0.618$ \\
\textbf{\spaced{SDV}}       & $0.508$ & $0.594$ & $0.350$ & $0.364$ & $0.596$ \\
\textbf{\spaced{TabDiT}}    & $0.023$ & $\underline{0.978}$ & $\mathbf{1.000}$ & $\underline{0.959}$ & $\underline{0.983}$ \\
\bottomrule
\end{tabular*}
\vspace{-1em}
\end{table}

\noindent\textbf{Timestamp.}
Timestamps encode trajectory-level structure that ordinary marginal fidelity
metrics cannot capture. We first examine whether the eight models generate
valid timestamp structure on Rossmann, including uniqueness, regular spacing,
grid coverage, and trajectory duration.
Table~\ref{tab:ts-rossmann} reports timestamp sub-metric scores on Rossmann,
a regular time-series dataset. The columns denote \texttt{\seqsplit{TemporalOrderConsistency}},
\texttt{\seqsplit{Timestamp Uniqueness}}, \texttt{\seqsplit{Trajectory Duration Similarity}}, \texttt{\seqsplit{Regularity Consistency}},
and \texttt{\seqsplit{Grid Completeness}}, respectively.
Together, these metrics distinguish misordered records, duplicate timestamps,
unrealistic durations, irregular intervals, and incomplete observation grids.
Rossmann follows a daily grid with unique timestamps at one-day intervals: \texttt{\seqsplit{Timestamp Uniqueness}} and \texttt{\seqsplit{Regularity Consistency}} equal $1.0$ by construction rather than by measurement. 
The scores below are therefore read against a known reference value, not against another metric's ranking. 
RTF and TabDiT consistently outperform the joint-generation models on timestamp
properties that remain meaningful after reordering. This pattern reflects their
generation mechanisms. Autoregressive models condition each event on the
preceding sequence, helping preserve uniqueness and regular spacing, whereas
joint-generation models lack this sequential constraint when generating
timestamps with other attributes. The joint models show
substantial violations even for timestamp properties that are fixed by
construction in the real data. Their relatively strong \textsc{sd} performance
indicates that plausible marginal distributions can mask these violations under
static evaluation.

\vspace{-0.28em}

\label{sec:cross-sectional-longitudinal}
\begin{table}[ht]
\centering
\caption{Cross-sectional sub-metric profiles on Rossmann. All scores are computed per time slice and lie in $[0,1]$, with higher values indicating better. Bold and underline mark the best and second-best scores per column.}
\vspace{-0.8em}
\Description{A results table reporting cross-sectional sub-metric scores on the Rossmann dataset under the sparse not-a-number fill variant. Rows compare synthetic data generation models, and columns report numerical marginal similarity, categorical marginal similarity, summary-statistic similarity, category coverage, and contingency similarity. Scores range from zero to one, with higher values indicating better cross-sectional fidelity.}
\label{tab:cs-rossmann}

\small
\setlength{\tabcolsep}{3.0pt}
\renewcommand{\arraystretch}{1.05}

\begin{tabular*}{\linewidth}{@{\extracolsep{\fill}}lccccc@{}}
\toprule
\textbf{Model}
  & {\footnotesize\texttt{CS-KSComp}}
  & {\footnotesize\texttt{CS-TVComp}}
  & {\footnotesize\texttt{CS-StatSim}}
  & {\footnotesize\texttt{CS-CatCov}}
  & {\footnotesize\texttt{CS-ContSim}} \\
\midrule
\textbf{ClavaDDPM} & $\underline{0.951}$ & $0.912$ & $0.775$ & $\underline{0.969}$ & $0.828$ \\
\textbf{RCTGAN}    & $0.874$ & $0.688$ & $0.723$ & $\mathbf{0.972}$ & $0.505$ \\
\textbf{RDBDiff}   & $0.948$ & $\underline{0.913}$ & $0.774$ & $\underline{0.969}$ & $\underline{0.830}$ \\
\textbf{RTF}       & $0.904$ & $0.828$ & $\underline{0.875}$ & $0.910$ & $0.702$ \\
\textbf{RGCLD}     & $0.940$ & $0.892$ & $0.765$ & $0.935$ & $0.806$ \\
\textbf{RelDiff}   & $0.881$ & $0.738$ & $0.681$ & $\mathbf{0.972}$ & $0.563$ \\
\textbf{SDV}       & $0.562$ & $0.496$ & $0.727$ & $0.734$ & $0.263$ \\
\textbf{TabDiT}    & $\mathbf{0.996}$ & $\mathbf{0.947}$ & $\mathbf{0.947}$ & $\mathbf{0.972}$ & $\mathbf{0.896}$ \\
\bottomrule
\end{tabular*}

\vspace{-0.5em}
\end{table}

\noindent\textbf{Cross-sectional dimension.}
Table~\ref{tab:cs-rossmann} shows the cross-sectional sub-metric profiles on
Rossmann. The displayed columns correspond to \texttt{CS-KSComplement},
\texttt{CS-TVComplement}, \texttt{\seqsplit{CS-StatisticSimilarity}},
\texttt{CS-CategoryCoverage}, and \texttt{\seqsplit{CS-ContingencySimilarity}}. 
TabDiT has the strongest cross-sectional profile, ranking best on most sub-metrics.
RCTGAN and RelDiff show asymmetric behavior, with high category coverage but
lower contingency similarity, indicating they recover per-slice
categories without preserving their joint combinations. SDV shows the weakest
profile, particularly on contingency similarity. These results show
cross-sectional fidelity requires preserving feature relationships beyond
marginal distributions and category coverage within each time slice.

\begin{figure}[ht]
    \centering
    \includegraphics[width=0.8\columnwidth]{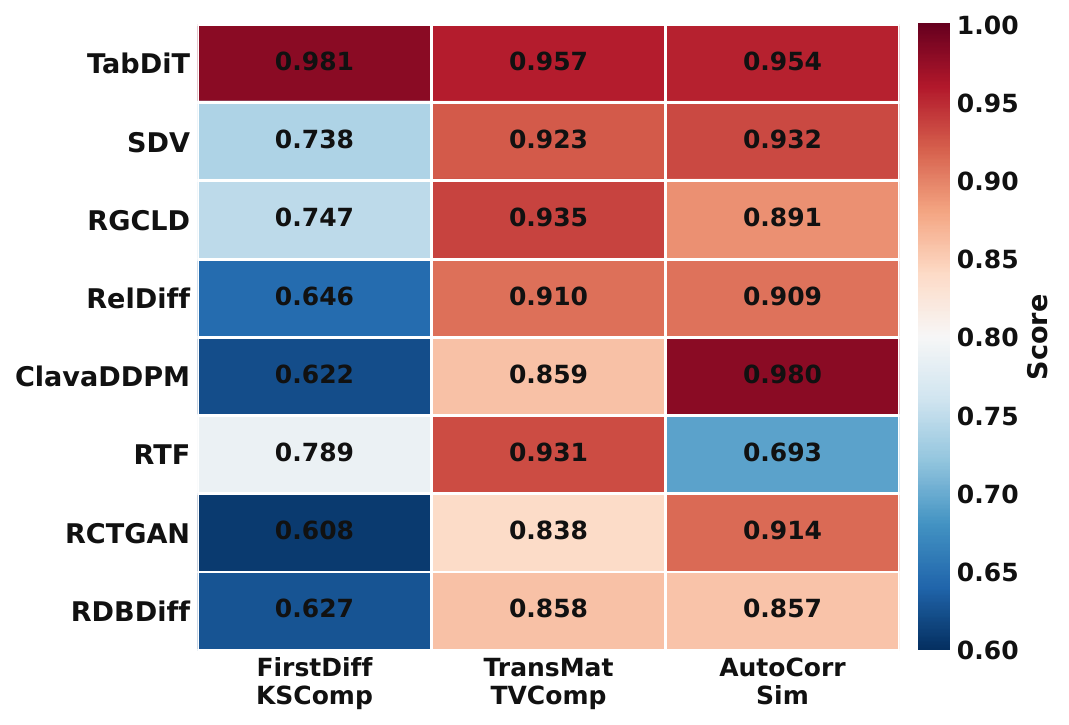}
    \Description{
    A heatmap showing longitudinal fidelity sub-metric profiles on the Rossmann dataset under the interpolation-fill variant.
    Rows correspond to synthetic data generation models, and columns correspond to three longitudinal sub-metrics: first-difference KS-complement, transition-matrix TV-complement, and autocorrelation similarity.
    Cell values range from zero to one, with darker red colors indicating higher scores and darker blue colors indicating lower scores.
    The heatmap shows that TabDiT achieves the highest scores on first-difference and transition-matrix metrics, while ClavaDDPM achieves the highest autocorrelation similarity.
    }
    \vspace{-0.8em}
    \caption{Longitudinal sub-metric profiles on Rossmann under the interpolation-fill variant. All scores lie in $[0,1]$, with higher values indicating better longitudinal fidelity.}
    \label{fig:rossmann_lg_heatmap}
\end{figure}

\noindent\textbf{Longitudinal dimension.}
Figure~\ref{fig:rossmann_lg_heatmap} reports longitudinal sub-metric profiles on
Rossmann using \texttt{\seqsplit{FirstDiffKSComplement}},
\texttt{\seqsplit{TransMatrixTVDComplement}}, and
\texttt{\seqsplit{AutoCorrSimilarity}}. TabDiT shows the strongest
profile, while RTF also performs well on order-sensitive metrics, consistent
with the autoregressive advantage observed in the timestamp analysis.
Differences across sub-metrics show that longitudinal fidelity is not a
single property. ClavaDDPM and SDV preserve autocorrelation well but are weaker
on local changes and transitions, whereas RTF shows the opposite pattern.
This indicates that longitudinal fidelity captures distinct aspects of
within-entity temporal dynamics that should be evaluated separately.

\label{sec:structural}

\begin{figure}[ht]
    \centering
    \includegraphics[width=0.9\columnwidth]{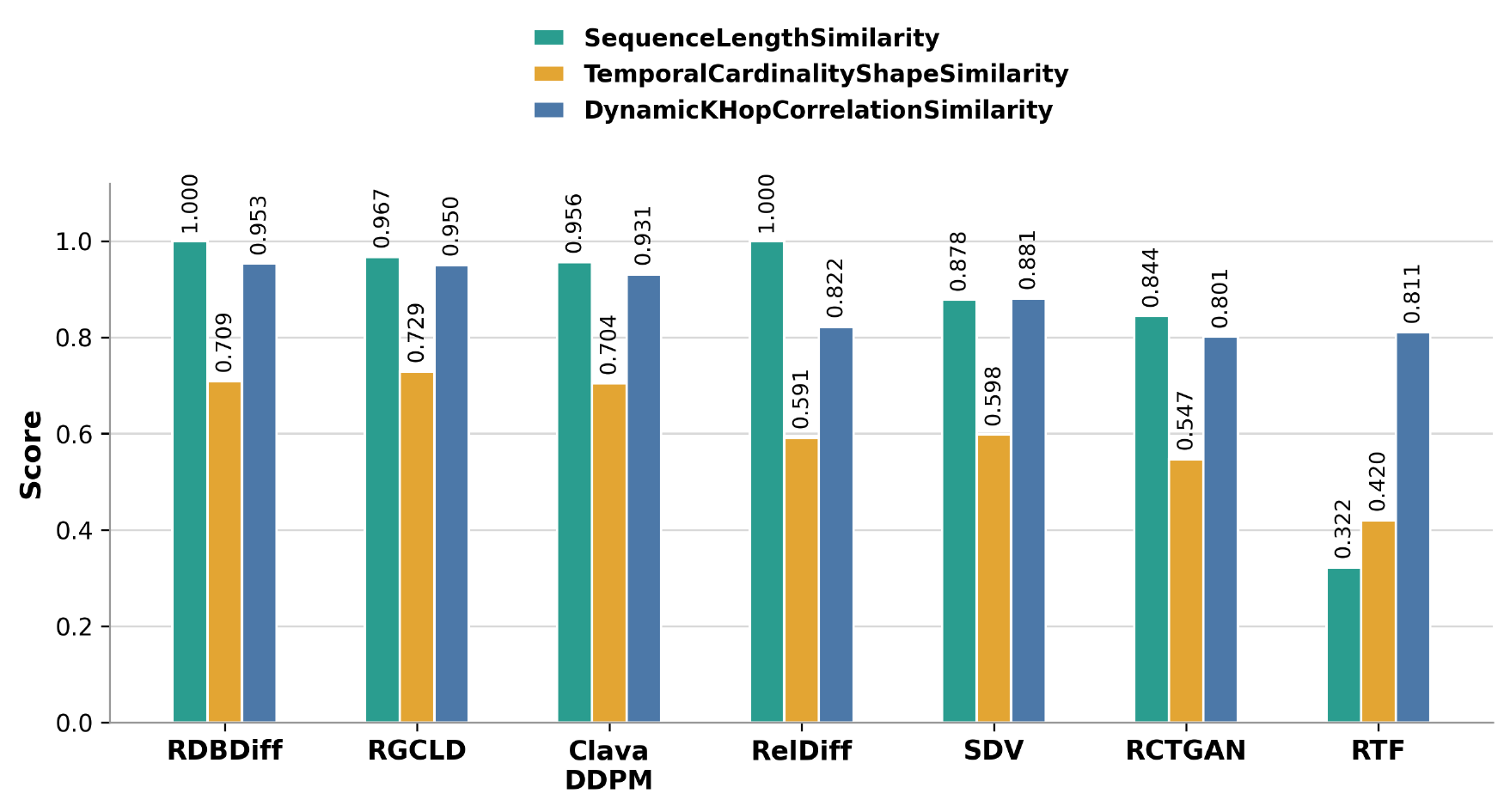}
    \Description{A grouped bar chart showing structural fidelity sub-metric scores on the Walmart dataset. The x-axis lists synthetic data generation models, and the y-axis reports normalized scores from zero to one, with higher values indicating better structural fidelity. For each model, three bars represent sequence-length similarity, temporal-cardinality similarity, and dynamic-hop similarity. The chart shows that RDBDiff, RGCLD, ClavaDDPM, RelDiff, and SDV achieve relatively high sequence-length and dynamic-hop scores, while temporal-cardinality scores are generally lower across models. RTF has the lowest sequence-length and temporal-cardinality scores but maintains a moderate dynamic-hop score.}
    \vspace{-0.8em}
    \caption{Structural fidelity sub-metrics on Walmart. All scores are normalized in $[0,1]$,
with higher values indicating better structural fidelity.}
    \label{fig:st_walmart}
\end{figure}

\noindent\textbf{Temporal relational structure.}
Figure~\ref{fig:st_walmart} describes structural sub-metric profiles on
Walmart, the only core dataset with applicable \textsc{st} sub-metric data.
The 3 sub-metrics are \texttt{Sequence Length Similarity},
\texttt{Temporal Cardinality Shape Similarity}, and
\texttt{Dynamic K-Hop Correlation Similarity}. The first 2
sub-metrics are computed from raw generator output, whereas
\texttt{\seqsplit{Dynamic K-Hop Correlation Similarity}} uses the sparse variant.
RTF reveals a limitation of autoregressive sequence generation in relational
settings. Although it performs strongly on timestamp and longitudinal
dimensions, its low structural fidelity is mainly driven by failures in
\texttt{\seqsplit{Sequence Length Similarity}} and
\texttt{\seqsplit{Temporal Cardinality Shape Similarity}}, indicating that it does not
reliably control the number of child records per parent or their temporal
cardinality patterns. By contrast, relational diffusion models such as RDBDiff
and ClavaDDPM improve substantially under \textsc{st}, despite weaker
\textsc{sd} rankings, because their joint generation under the relational
schema better preserves sequence length and cross-table correlation structure
across parent entities.
Near-saturated \textsc{st} scores are expected for linear schemas, where structural fidelity reduces to sequence-length similarity. In
multi-child schemas, additional structural sub-metrics become applicable and
better distinguish model behavior. Per-dataset results are reported in
Appendix~\ref{app:addition_results} (Table~\ref{tab:per-dataset-all}).

\paragraph{Finding 3: Temporal evaluation provides a complementary lens beyond static privacy and utility metrics.}
\label{sec:privacy}

\begin{table}[ht]
\centering
\caption{Privacy metric profiles on Rossmann. \texttt{DCR} and \texttt{CS-DCR}
report raw median nearest-neighbor distances. \texttt{NNDR},
\texttt{CS-NNDR}, and \texttt{NGP}(n) for $n\in\{1,3\}$ are normalized
scores in $[0,1]$, where \texttt{NGP}(n)$=1-\texttt{NgramExposure}(n)$.
Higher values are safer; bold/underline indicate best/second-best.}
\vspace{-0.4em}
\Description{A results table reporting privacy metric profiles on the Rossmann dataset. Rows compare synthetic data generation models, and columns are grouped into row-level privacy, cross-sectional privacy, and n-gram privacy. Higher values indicate safer behavior and lower privacy risk.}
\label{tab:privacy-rossmann}

\small
\setlength{\tabcolsep}{2.6pt}
\renewcommand{\arraystretch}{1.02}

\begin{tabular*}{\columnwidth}{@{\extracolsep{\fill}}lcccccc@{}}
\toprule
\multirow{2}{*}{\textbf{Model}}
  & \multicolumn{2}{c}{\footnotesize\textbf{Row-level}}
  & \multicolumn{2}{c}{\footnotesize\textbf{Cross-sectional}}
  & \multicolumn{2}{c}{\footnotesize\textbf{$n$-gram Privacy}} \\
\cmidrule(lr){2-3}
\cmidrule(lr){4-5}
\cmidrule(lr){6-7}
  & \texttt{DCR} & \texttt{NNDR}
  & \texttt{CS-DCR} & \texttt{CS-NNDR}
  & \texttt{NGP}(1) & \texttt{NGP}(3) \\
\midrule
\textbf{ClavaDDPM} & $8.06$             & $0.707$          & $49.46$             & $0.653$          & $0.002$ & $0.912$ \\
\textbf{RCTGAN}    & $9.22$             & $0.709$          & $73.48$             & $0.695$          & $0.074$ & $\underline{0.998}$ \\
\textbf{RDBDiff}   & $6.08$             & $0.662$          & $16.24$             & $0.489$          & $0.000$ & $0.833$             \\
\textbf{RTF}       & $9.06$             & $\underline{0.721}$          & $64.80$             & $\underline{0.742}$          & $0.028$ & $0.180$             \\
\textbf{RGCLD}     & $8.06$             & $0.707$          & $54.72$             & $0.697$          & $0.014$ & $0.907$             \\
\textbf{RelDiff}   & $\mathbf{12.04}$   & $0.703$          & $\mathbf{96.72}$    & $0.693$          & $\underline{0.209}$ & $\underline{0.998}$ \\
\textbf{SDV}       & $\underline{9.85}$ & $0.707$          & $\underline{86.76}$ & $0.661$          & $\mathbf{0.221}$ & $\mathbf{0.999}$    \\
\textbf{TabDiT}    & $11.00$            & $\mathbf{0.766}$ & $62.84$             & $\mathbf{0.753}$ & $0.008$ & $0.267$             \\
\bottomrule
\end{tabular*}

\vspace{-1em}
\end{table}

\mbox{}\\[-5pt]

\noindent\textbf{Privacy.} 
Standard \texttt{DCR}, \texttt{NNDR}, and single-row exposure treat records as
exchangeable across time. This can misrepresent temporal privacy in two ways.
First, identical values at unrelated timestamps may reflect incidental overlap
rather than re-identification risk, especially with bounded row vocabularies.
Second, row-level metrics cannot detect reproduced consecutive sub-sequences of
real trajectories. We therefore compare them with \texttt{CS-DCR},
\texttt{CS-NNDR}, and $n$-gram privacy (\(\texttt{NGP}(n)=1-\texttt{NgramExposure}(n)\)) on Rossmann. Table~\ref{tab:privacy-rossmann} shows that row-level metrics can misrepresent temporal privacy risk.
Time-agnostic nearest-neighbor metrics such as \texttt{DCR} and \texttt{NNDR}
ignore whether matched synthetic and real rows belong to the same time slice.
In low-cardinality sequential data, this can
count incidental cross-time matches as privacy risk; the larger
\texttt{CS-DCR} values indicate that row-level nearest neighbors come from
unrelated timestamps. Conversely, single-row exposure can miss sequential
leakage: \texttt{NGP}$^{(1)}$ does not detect copied consecutive sub-sequences.
RTF and TabDiT illustrate this failure, appearing safe under
row-level metrics but becoming the riskiest models under
\texttt{NGP}$^{(3)}$. Thus, the autoregressive mechanism that improves
within-trajectory fidelity can increase trajectory-level
privacy risk. Results on the remaining 6
datasets are reported in Appendix~\ref{app:privacy-online}.

\begin{table}[ht]
\centering
\caption{Correlation results for \texttt{MLE-TS} and \texttt{MLE-Temporal}. 
Bold/underline mark best/second-best per row.}
\Description{Correlation results on the four datasets shared by MLE-TS and MLE-Temporal. The table reports correlations between downstream utility and three fidelity scores under both evaluation protocols. The Average row denotes the mean correlation across the four datasets. Bold and underlined values indicate the best and second-best correlations within each row, respectively.}
\label{tab:rho_corr_mle_ts_temporal}
\footnotesize
\setlength{\tabcolsep}{3pt}
\renewcommand{\arraystretch}{0.95}
\vspace{-0.5em}
\resizebox{\columnwidth}{!}{%
\begin{adjustbox}{max width=\linewidth}
\begin{tabular}{l|*{3}{C{0.05\textwidth}}|*{3}{C{0.05\textwidth}}}
\toprule
\multirow{2}{*}{Dataset}
& \multicolumn{3}{c|}{\texttt{MLE-TS}}
& \multicolumn{3}{c}{\texttt{MLE-Temporal}} \\
\cmidrule(lr){2-4}
\cmidrule(lr){5-7}
& $\rho_{\mathrm{SD}}$
& $\rho_{\mathrm{long}}$
& $\rho_{\mathrm{all}}$
& $\rho_{\mathrm{SD}}$
& $\rho_{\mathrm{long}}$
& $\rho_{\mathrm{all}}$ \\
\midrule
\textbf{Rossmann}
& 0.690 & 0.524 & 0.667
& \textbf{0.929} & 0.690 & \underline{0.786} \\

\textbf{Fannie Mae}
& \underline{0.321} & \underline{0.321} & 0.107
& $-0.143$ & \textbf{0.643} & $-0.214$ \\

\textbf{Walmart}
& $-0.464$ & 0.607 & 0.500
& 0.071 & \underline{0.643} & \textbf{0.750} \\

\textbf{Freddie Mac}
& 0.143 & 0.143 & 0.214
& \underline{0.750} & \underline{0.750} & \textbf{0.821} \\
\midrule
Average
& 0.173 & \textbf{0.399} & 0.372
& 0.402 & \textbf{0.682} & \underline{0.536} \\
\bottomrule
\end{tabular}%
\end{adjustbox}}
\vspace{-0.8em}
\end{table}

\noindent\textbf{Utility.}
Table~\ref{tab:rho_corr_mle_ts_temporal} reports correlation results on the
four datasets shared by \texttt{MLE-TS} and \texttt{MLE-Temporal}. 
These four datasets follow directly from the applicability of
\texttt{MLE-Temporal}, which requires absolute time and a regular observation
grid under a chronological split.
The results show that \textsc{sd}-based evaluation does not fully explain downstream
temporal utility. Under \texttt{MLE-TS}, longitudinal fidelity correlates more
strongly with utility than \textsc{sd}
($\rho_{\mathrm{long}}=0.399$ vs.\ $\rho_{\mathrm{SD}}=0.173$), and this gap
widens under \texttt{MLE-Temporal}
($0.682$ vs.\ $0.402$). This indicates that as downstream tasks use more
historical context, utility depends increasingly on preserving within-trajectory
dynamics rather than static marginal distributions alone. Full results are
available in Appendices~\ref{app:mle-ts-online} and~\ref{app:mle-temporal-online}.
\section{Discussion}
\label{sec:discussion}
\paragraph{Implications for benchmark use.}
The results provide empirical evidence that Seq2Synth should be used as a diagnostic protocol rather than as a single leaderboard score. Static-distribution fidelity remains
useful for checking pooled record-level plausibility, but it is insufficient
for model selection when the deployment depends on temporal behavior. The
taxonomy-to-metric mapping is therefore essential: timestamp and longitudinal
metrics test whether individual trajectories are valid, cross-sectional
metrics test whether population structure evolves correctly over absolute
time, and structural metrics test whether relational cardinalities and
cross-table dependencies are preserved. These dimensions should be interpreted
jointly, because a model can be strong along one temporal axis and fail along
another. The same principle applies to privacy and utility: row-level privacy
and static downstream performance should be complemented by trajectory-aware
views when sequential leakage or temporal prediction is the relevant risk.

\paragraph{Implications for generator design.}
The failure modes identified by Seq2Synth point to three concrete design
requirements for future sequential-tabular generators. First, timestamps should
be modeled as ordered objects rather than as ordinary continuous attributes;
otherwise, a model can match timestamp marginals while producing invalid
chronological structure. Second, autoregressive generators need explicit
absolute-time anchoring, such as calendar embeddings, anchor tokens, or
auxiliary timestamp losses, so that token position is not confused with
calendar time. Third, sequence length and relational cardinality should be
conditioned on directly rather than left to emerge implicitly, especially in
multi-table schemas where parent--child counts evolve over time. These changes
do not require a new model family: they can be incorporated into existing
diffusion, autoregressive, or relational generators through constrained
decoding, auxiliary objectives, or conditioning inputs.
\section{Conclusion}
\label{sec:conclusion}

We presented \textit{Seq2Synth}, a taxonomy-guided benchmark for evaluating
synthetic sequential tabular data. Rather than treating all datasets with a
fixed metric suite, Seq2Synth maps time representation, sampling regularity,
trajectory dependence, and schema structure to applicable temporal fidelity
dimensions: timestamp, cross-sectional, longitudinal, and structural fidelity.
Across a 13-dataset benchmark and eight generators, our experiments show that
strong static-distribution fidelity does not reliably translate to temporal fidelity. Model rankings differ substantially between static and temporal-aware evaluations, demonstrating that temporal fidelity cannot be inferred from static evaluation alone and must be assessed directly.
The trajectory-aware utility and privacy analyses show that
temporal evaluation changes how model quality and leakage risk should be
interpreted. This indicates that synthetic sequential tabular data are best evaluated as temporally ordered trajectories and relational processes, not as pooled rows.

\begin{acks}
This work was supported by the National Research Foundation of Korea (NRF) grant funded by the Korea government (MSIT) (No. RS-2025-24803208, No. RS-2024-00354727) and the Institute of Information \& Communications Technology Planning \& Evaluation (IITP) grant funded by the Korea government (MSIT) (No. RS-2020-II201336, Artificial Intelligence Graduate School Program (UNIST), No.RS-2026-25616928, AI Star Fellowship Support Program, RS-2024-00436936, Development of a user-friendly and efficiency-\seqsplit{optimized} real-time homomorphic statistical analysis processing platform, RS-2026-25528781). This work was also supported by the Hankuk University of Foreign Studies Research Fund (Of 2026).
\end{acks}






\section*{GenAI Usage Disclosure}
Generative AI tools were employed in the preparation of this manuscript to support language editing, manuscript drafting, and limited code refinement. They were not used for dataset generation, experimental execution, statistical analysis, or formulation of scientific claims. The authors independently reviewed and validated all technical content, references, code, results, and conclusions, and take full responsibility for the final manuscript.
\bibliographystyle{ACM-Reference-Format}
\bibliography{references}

\clearpage
\appendix


\section*{Note on the Appendices}
Appendix~\ref{app:metrics} below is the appendix of the CIKM~'26
camera-ready version. Appendices~\ref{app:taxonomy}--\ref{app:addition_results}
are the online appendix that the published version distributes through the
project page; their lettering and subsection numbering are unchanged, so a
pointer such as ``Online Appendix~E.2'' in the published paper corresponds to
Appendix~\ref{app:postprocessing-ablation} here.

\providecommand{\OApp}[2]{Online Appendix #1}

\begin{table*}[t]
\centering
\caption{Applicability Conditions, Feature Scope, and Provenance of the Evaluation Metrics.
\textbf{Scope}: U = Univariate (per feature, averaged), B = Bivariate (per feature pair, averaged), M = Multivariate (joint computation).
The ``Applicable Features'' column specifies which feature types each metric operates on.
\textbf{Type} is marked on the metric name (see the note below the table).}
\Description{A catalog table summarizing the applicability conditions and feature scope of the proposed evaluation metrics. Rows are grouped by evaluation perspective, including timestamp, cross-sectional, longitudinal, structural, and privacy metrics. Columns describe each metric name, whether it is computed over univariate, bivariate, or multivariate feature scope, the applicable feature types, supported time representation, periodicity requirement, trajectory independence condition, and data-structure requirement. The table shows that the proposed metrics cover a broad range of temporal and relational settings, while some metrics require specific conditions such as absolute time, regular periodicity, dependent trajectories, or multi-child relational structures. Each metric name additionally carries a provenance marker: no marker means the metric is newly introduced in this work, a double dagger means it is adapted from an existing measure, and a section symbol means it is adopted as originally defined, with the source cited in the appendix.}
\label{tab:metric_applicability}
\resizebox{\textwidth}{!}{%
\begin{tabular}{llcccccl}
\toprule
\textbf{Perspective} & \textbf{Metric Name} & \textbf{Scope} & \textbf{Applicable Features} & \textbf{Time Repr.} & \textbf{Periodicity} & \textbf{Traj. Indep.} & \textbf{Data Str.} \\
\midrule

\multirow{7}{*}{\rotatebox{90}{Timestamp}}
& \texttt{TimeIntervalDistributionSimilarity}$^\ddagger$ & U & $x^{time}$       & Both & Irregular & Both & All \\
& \texttt{RegularityConsistency}                    & U & $x^{time}$       & Both & Both & Both & All \\
& \texttt{TemporalRangeCompliance}                 & U & $x^{time}$       & Both & Both & Both & All \\
& \texttt{GridCompleteness}                         & U & $x^{time}$       & Both & Regular & Both & All \\
& \texttt{TemporalOrderConsistency}                  & U & $x^{time}$       & Both & Both & Both & All \\
& \texttt{TimestampUniqueness}                      & U & $x^{time}$       & Both & Both & Both & All \\
& \texttt{TrajectoryDurationSimilarity}$^\ddagger$ & U & $x^{time}$       & Both & Both & Both & All \\
\midrule
 
\multirow{7}{*}{\rotatebox{90}{Cross-sectional}}
& \texttt{CS-KSComplement}$^\ddagger$          & U & Each $x^{num}$              & Abs. & Both & Dep. & All$^*$ \\
& \texttt{CS-StatisticSimilarity}$^\ddagger$   & U & Each $x^{num}$              & Abs. & Both & Dep. & All$^*$ \\
& \texttt{CS-RangeCoverage}$^\ddagger$         & U & Each $x^{num}$              & Abs. & Both & Dep. & All$^*$ \\
& \texttt{CS-TVComplement}$^\ddagger$          & U & Each $x^{cat}$              & Abs. & Both & Dep. & All$^*$ \\
& \texttt{CS-CategoryCoverage}$^\ddagger$      & U & Each $x^{cat}$              & Abs. & Both & Dep. & All$^*$ \\
& \texttt{CS-CorrelationSimilarity}$^\ddagger$ & B & Each $(x^{num}_a, x^{num}_b)$ pair & Abs. & Both & Dep. & All$^*$ \\
& \texttt{CS-ContingencySimilarity}$^\ddagger$ & B & Each $(x^{cat}_a, x^{cat}_b)$ pair & Abs. & Both & Dep. & All$^*$ \\
\midrule
 
\multirow{5}{*}{\rotatebox{90}{\small Longitudinal}}
& \texttt{FirstDifferenceKSComplement}$^\ddagger$ & U & Each $x^{num}$   & Both & Regular & Both & All \\
& \texttt{TransitionMatrixTVDComplement}$^\ddagger$ & U & Each $x^{cat}$   & Both & Regular & Both & All \\
& \texttt{AutocorrelationSimilarity}$^\ddagger$ & U & Each $x^{num}$   & Both$^\dagger$ & Regular & Both$^\dagger$ & All \\
& \texttt{TT-WassersteinDistance}$^\ddagger$  & M & All $x^{num}$ jointly & Both & Regular & Both & All \\
& \texttt{MLD-TS}$^\S$            & M & All features at time $t$ & Both & Both & Both & All \\
\midrule
 
\multirow{3}{*}{\rotatebox{90}{Struct.}}
& \texttt{SequenceLengthSimilarity}$^\S$ & U & trajectory length $L_k$         & Both & Both & Both & All \\
& \texttt{TemporalCardinalityShapeSimilarity}$^\ddagger$ & U & Each parent--child relation     & Abs. & Both & Both & M-Child \\
& \texttt{DynamicKHopCorrelationSimilarity}$^\ddagger$ & B & Cross-table $(x_a, x_b)$ pair    & Abs. & Regular & Dep. & M-Child \\
\midrule
 
\multirow{3}{*}{\rotatebox{90}{Privacy}}
& \texttt{CS-DCR}$^\ddagger$                       & M & All features at time $t$ & Abs. & Both & Dep. & All \\
& \texttt{CS-NNDR}$^\ddagger$                     & M & All features at time $t$ & Abs. & Both & Dep. & All \\
& \texttt{n-gramPrivacy}                      & M & All features at time $t$ & Both & Both & Both & All \\
\midrule

\multirow{2}{*}{\rotatebox[origin=c]{90}{Utility}}
& \rule{0pt}{2.9ex}\texttt{MLE-TS}$^\S$
& M & All features at time $t$ & Abs. & Both & Both & All \\
& \rule{0pt}{2.9ex}\texttt{MLE-Temporal}
& M & All features at time $t$ & Abs. & Regular & Both & All \\
\bottomrule

\end{tabular}%
}
\vspace{2pt}
{\footnotesize $^*$For multi-table data only. \;
$^\dagger$Mode depends on trajectory independence. \;
Scope: U is per-feature, B is per-pair, and M is joint over all features; U and B are averaged.\\
\textbf{Type} (on the metric name): unmarked $=$ newly introduced here (7); $^\ddagger$ $=$ adapted (17);
$^\S$ $=$ adopted as defined (3). The source and modification of each adapted metric, and the source of each adopted metric, are given with their definitions in Appendix~\ref{app:metrics}; provenance is provided in \OApp{C}{app:metric_provenance}.
 \texttt{DCR}/\texttt{NNDR} are time-agnostic baselines and are not counted.}
\end{table*}

\section{Evaluation Metrics}
\label{app:metrics}

This appendix gives the operational definition of every metric used in
the main paper. All metrics produce a scalar in $[0,1]$ unless noted otherwise; $1$
denotes perfect fidelity. Notation follows
Section~\ref{subsec:data_model}: $T$ and $\hat{T}$ are the real and
synthetic tables, $x^{time}_{k,j}$ is the timestamp of the $j$-th
record of trajectory $k$, $\Delta t_{k,j}=x^{time}_{k,j}-x^{time}_{k,j-1}$,
$L_k$ is trajectory length, and $K$ is the number of trajectories.
Each definition states the metric's Type as marked in Table~\ref{tab:metric_applicability}.

\subsection{Feature Scope and Aggregation}
\label{app:feature_scope}

Each metric operates at one of three scopes:
\textbf{univariate} (per feature column $f$),
\textbf{bivariate} (per feature pair $(f_1,f_2)$),
or \textbf{multivariate} (joint feature vector / trajectory embedding).
For univariate and bivariate metrics, the per-feature (or per-pair)
score is averaged uniformly across the applicable set $\mathcal{F}$
(or $\mathcal{P}$); the applicable feature types per metric are listed in
Table~\ref{tab:metric_applicability}.

\subsection{Timestamp Metrics}
\label{app:timestamp_metrics}

\begingroup
\setlength{\abovedisplayskip}{3pt}
\setlength{\belowdisplayskip}{3pt}
\setlength{\abovedisplayshortskip}{2pt}
\setlength{\belowdisplayshortskip}{2pt}

All timestamp metrics are computed on the \emph{raw} synthetic output before
sorting, deduplication, boundary trimming, or imputation, so generator-level
timestamp failures are not hidden by post-processing.
The first five metrics are new. \texttt{TimeIntervalDistrSim} and \texttt{TrajDurationSim} are
adapted: they apply the KS complement of SDMetrics~\cite{sdvlibrary} to the inter-arrival and
trajectory-duration distributions instead of to feature values.

\noindent\textbf{TemporalOrderConsistency} (\texttt{TempOrderConsist}; all series) measures the fraction of consecutive timestamp pairs that are non-decreasing:
\begin{equation}
    1 - \frac{\sum_k \sum_{j=2}^{L_k}
    \mathbb{I}(\Delta t_{\mathrm{synth},k,j} < 0)}
    {\sum_k (L_k - 1)}.
    \label{eq:chrono}
\end{equation}

\noindent\textbf{TimestampUniqueness} (\texttt{TimestampUnique}; all series) measures the fraction of consecutive intervals that are strictly positive, detecting duplicate timestamps before deduplication:
\begin{equation}
    1 - \frac{\sum_k \sum_{j=2}^{L_k}
    \mathbb{I}(\Delta t_{\mathrm{synth},k,j} = 0)}
    {\sum_k (L_k - 1)}.
    \label{eq:uniqueness}
\end{equation}

\noindent\textbf{TemporalRangeCompliance} (\texttt{TempRangeComp}; all series) measures the fraction of synthetic records within the valid real temporal domain $\Omega_{\mathrm{real}}$; for relative time, the first timestamp must also be anchored at zero:
\begin{equation}
    1 - \frac{1}{N_{\mathrm{synth}}}\sum_k \sum_j V_{k,j},
    \label{eq:boundary}
\end{equation}
where $V_{k,j}$ is the per-record violation indicator.

\noindent\textbf{RegularityConsistency} (\texttt{RegularityConsist}; regular series only) measures the fraction of synthetic intervals matching the canonical interval $c$ within tolerance $\epsilon=0.01c$:
\begin{equation}
    1 - \frac{\sum_k \sum_{j=2}^{L_k}
    \mathbb{I}(|\Delta t_{\mathrm{synth},k,j} - c| > \epsilon)}
    {\sum_k (L_k - 1)}.
    \label{eq:regularity}
\end{equation}

\noindent\textbf{GridCompleteness} (\texttt{GridCompleteness}; regular series only) compares real and synthetic canonical grid-point multiplicities, capping synthetic credit at the real multiplicity:
\begin{equation}
    \frac{\sum_g \min(M_{\mathrm{real}}(g), M_{\mathrm{synth}}(g))}
    {\sum_g M_{\mathrm{real}}(g)}.
\end{equation}

\noindent\textbf{TimeIntervalDistributionSimilarity} (\texttt{TimeIntervalDistrSim}; irregular series only) compares pooled real and synthetic inter-arrival distributions using the KS complement:
\begin{equation}
    1 - \sup_{\Delta t}
    \bigl|F_{\mathrm{real}}(\Delta t) - F_{\mathrm{synth}}(\Delta t)\bigr|.
    \label{eq:timedist}
\end{equation}

\noindent\textbf{TrajectoryDurationSimilarity} (\texttt{TrajDurationSim}; all series) compares real and synthetic trajectory-duration distributions using the KS complement, with $D_k=x^{time}_{k,L_k}-x^{time}_{k,1}$:
\begin{equation}
    1 - \sup_D
    \bigl|F_{\mathrm{real}}(D) - F_{\mathrm{synth}}(D)\bigr|.
    \label{eq:duration}
\end{equation}
\endgroup

\subsection{Cross-Sectional Metrics}
\label{app:cross_sectional_metrics}

Cross-sectional metrics compare real and synthetic feature distributions within
aligned temporal partitions. They apply to datasets with \emph{Absolute Time}
and \emph{Dependent Trajectories}. Let
$\Omega=\bigsqcup_{l=1}^{B}\Omega_l$ be a partition of the temporal domain,
and let $T\vert_l$ and $\hat{T}\vert_l$ be records whose timestamps fall in
$\Omega_l$. Any base measure $\mu$ is lifted to a cross-sectional metric as
\begin{equation}
\label{eq:cs_general}
    \mu_{CS}(T,\hat{T}) =
    \frac{1}{B}\sum_{l=1}^{B} \mu(T\vert_l,\hat{T}\vert_l).
\end{equation}
All seven metrics here are \emph{adapted}: each $\mu$ is the corresponding SDMetrics
measure~\cite{sdvlibrary} (KSComplement, StatisticSimilarity, BoundaryAdherence, TVComplement,
CategoryCoverage, CorrelationSimilarity, ContingencySimilarity, respectively). 
For feature $f$, $T_l^{(f)}$ and $\hat{T}_l^{(f)}$ denote values in partition
$\Omega_l$; $F_{T_l}^{(f)}$ is the numerical CDF and $P_{T_l}^{(f)}$ the categorical distribution.

\noindent\textbf{CS-KSComplement} compares numerical distribution shapes within each
time partition:
\begin{equation}
    \mu^{KS}(T\vert_l,\hat{T}\vert_l)
    =
    1 - \sup_x
    \left|F_{T_l}^{(f)}(x) - F_{\hat{T}_l}^{(f)}(x)\right|.
\end{equation}

\textbf{CS-StatisticSimilarity} compares time-local summary statistics
$S\in\{\mathrm{mean},\mathrm{median},\mathrm{std}\}$:
\begin{equation}
    \mu^{Stat}_{S}(T\vert_l,\hat{T}\vert_l)
    =
    1 -
    \frac{
    \left|S(T_l^{(f)}) - S(\hat{T}_l^{(f)})\right|
    }{
    \max_{l'} S(T_{l'}^{(f)}) - \min_{l'} S(T_{l'}^{(f)}) + \epsilon
    },
\end{equation}

\noindent\textbf{CS-RangeCoverage} measures whether synthetic numerical values remain
inside the real range at each time partition:
\begin{equation}
    \mu^{Range}(T\vert_l,\hat{T}\vert_l)
    =
    \frac{1}{|\hat{T}_l^{(f)}|}
    \sum_{x\in\hat{T}_l^{(f)}}
    \mathbb{I}\!\left[
    x \in
    \left[\min T_l^{(f)}, \max T_l^{(f)}\right]
    \right].
\end{equation}

\noindent\textbf{CS-TVComplement} compares categorical distributions using the total
variation complement:
\begin{equation}
    \mu^{TV}(T\vert_l,\hat{T}\vert_l)
    =
    1 - \frac{1}{2}
    \sum_{c}
    \left|P_{T_l}^{(f)}(c) - P_{\hat{T}_l}^{(f)}(c)\right|.
\end{equation}

\textbf{CS-CategoryCoverage} measures whether real categories at a time
partition also appear in synthetic data:
\begin{equation}
    \mu^{CatCov}(T\vert_l,\hat{T}\vert_l)
    =
    \frac{
    |C_{T_l}^{(f)} \cap C_{\hat{T}_l}^{(f)}|
    }{
    |C_{T_l}^{(f)}|
    },
\end{equation}
where empty real partitions are excluded from the average.

\textbf{CS-CorrelationSimilarity} compares numerical pairwise correlations
within each time partition:
\begin{equation}
    \mu^{Corr}(T\vert_l,\hat{T}\vert_l)
    =
    1 -
    \frac{
    |\rho_{T_l}(f_1,f_2) - \rho_{\hat{T}_l}(f_1,f_2)|
    }{2}.
\end{equation}

\textbf{CS-ContingencySimilarity} compares categorical co-occurrence patterns
using the TV complement over joint category frequencies:
\begin{equation}
    \mu^{Cont}(T\vert_l,\hat{T}\vert_l)
    =
    1 - \frac{1}{2}
    \sum_{a}\sum_{b}
    \left|
    P_{T_l}^{(f_1,f_2)}(a,b)
    -
    P_{\hat{T}_l}^{(f_1,f_2)}(a,b)
    \right|.
\end{equation}

\subsection{Longitudinal Metrics}
\label{app:longitudinal_metrics}

Longitudinal metrics treat each entity as an ordered trajectory. Step-level
metrics compare consecutive changes; trajectory-level metrics compare longer
range temporal structure. 

\noindent\textbf{FirstDifferenceKSComplement}
(\texttt{\seqsplit{FirstDiffKSComplement}}; numerical; \emph{adapted}). Applies the KS
complement~\cite{sdvlibrary} to pooled first differences
$\Delta x_{k,j}=x_{k,j}-x_{k,j-1}$ rather than to raw values:
\begin{equation}
    1 -
    \sup_{\delta}
    \left|
    F^{\Delta}_{\mathrm{real}}(\delta)
    -
    F^{\Delta}_{\mathrm{synth}}(\delta)
    \right|.
    \label{eq:fd_ks}
\end{equation}

\medskip
\noindent\textbf{TransitionMatrixTVDComplement}
(\texttt{TransMatrixTVDComplement}; categorical; \emph{adapted}). Applies the total-variation
complement~\cite{sdvlibrary} to per-timestep \emph{conditional} transition matrices. 
For category set $C$ and valid transition times $\mathcal{T}$,
\begin{equation}
1-
\frac{1}{2|C||\mathcal{T}|}
\sum_{t\in\mathcal{T}}\sum_{i,j\in C}
\left|P^r_t(j\mid i)-P^s_t(j\mid i)\right|.
\label{eq:transition}
\end{equation}
Here $P^r_t$ and $P^s_t$ denote real and synthetic transition probabilities;
scores are averaged over categorical features.

\noindent\textbf{AutocorrelationSimilarity}
(\texttt{\seqsplit{AutoCorrSimilarity}}; numerical; \emph{adapted}). Adapts the classical
autocorrelation function used to compare synthetic and real series~\cite{fints}, with a
dependence-aware computation mode, over lags up to $H=\min(L/4,50)$:
\begin{equation}
    1 - \frac{1}{2H}\sum_{h=1}^{H}
    \left|
    \rho^{\mathrm{real}}_h
    -
    \rho^{\mathrm{synth}}_h
    \right|.
    \label{eq:acf}
\end{equation}
For independent trajectories, autocorrelation is computed per trajectory and
averaged; for dependent trajectories with shared absolute time, it is computed
on the cross-sectional mean trajectory.

\medskip
\noindent\textbf{MLD-TS} (\texttt{MLD-TS}; multivariate; \emph{adopted}  from TabDiT~\cite{tabdit}). MLD-TS evaluates
whether real and synthetic trajectories are distinguishable by a supervised
classifier. Each trajectory is embedded as $z_k=\psi(T(k))$, labeled by source,
and classified by $h$; raw MLD-TS is detection accuracy.
Values near chance indicate low distinguishability, while higher values
indicate more detectable temporal mismatch. Classifier settings are provided
in \OApp{E.6}{app:details_tab}.

\medskip

\noindent\textbf{TT-WassersteinDistance}
(\texttt{TT-WassersteinDistance}; multivariate; \emph{adapted} from~\citet{ttwasserstein}).
TT-Wasserstein compares the global distribution of trajectory patterns via
spectral embeddings. Each trajectory $X\in\mathbb{R}^{L_k\times C}$ is
z-score normalized and first-differenced, with
$\phi^{(c)}(X)=\bigl|\mathrm{DFT}(z^{(c)}_t-z^{(c)}_{t-1})_{1:F}\bigr|$
and $\phi(X)=\mathrm{vec}([\phi^{(1)},\ldots,\phi^{(C)}])$.
With cost
$M_{ij}=\|\phi(X_i^{\mathrm{real}})-\phi(X_j^{\mathrm{synth}})\|^2$,
\begin{equation}
    D_{\mathrm{TTW}}
    =
    \left(
    \min_{\Gamma\in\Pi(a,b)}
    \sum_{i,j}\Gamma_{ij}M_{ij}
    \right)^{1/2}.
    \label{eq:tt_wasserstein}
\end{equation}
Smaller $D_{\mathrm{TTW}}$ indicates better preservation of the trajectory
distribution; it is reported as a distance rather than a $[0,1]$ similarity.

\subsection{Structural Metrics}
\label{app:structural_metrics}

Structural metrics evaluate relational properties of sequential data; larger
values indicate better structural fidelity.

\noindent\textbf{SequenceLengthSimilarity}
(\emph{adopted} from SDMetrics~\cite{sdvlibrary}; all \seqsplit{schemas)} compares the real and
synthetic trajectory-length distributions:
\begin{equation}
    1 - \sup_{L}
    \left|
    F_{\mathrm{real}}(L) - F_{\mathrm{synth}}(L)
    \right|.
    \label{eq:seq_len}
\end{equation}

\noindent\textbf{TemporalCardinalityShapeSimilarity}
(\emph{adapted} from CardinalityShapeSimilarity~\cite{syntherela}; multi-child schemas) replaces the global child-count distribution with counts inside aligned temporal windows,
per parent--child relation:
\begin{equation}
    1 - \sup_{s}
    \left|
    F_{\mathrm{real}}(s) - F_{\mathrm{synth}}(s)
    \right|,
    \label{eq:cardinality}
\end{equation}
where $s$ denotes child-record counts within aligned temporal windows.

\noindent\textbf{DynamicKHopCorrelationSimilarity}
(\emph{adapted} from SyntheRela's $k$-hop correlation similarity~\cite{syntherela}; multi-child
schemas) replaces the static cross-table correlation with the time average of $\rho(t)$:
\begin{equation}
    1 -
    \frac{1}{|\mathbb{T}_{\mathrm{eval}}|}
    \sum_{t\in\mathbb{T}_{\mathrm{eval}}}
    \frac{|
    \rho_{\mathrm{real}}(t) - \rho_{\mathrm{synth}}(t)
    |}{2}.
    \label{eq:khop}
\end{equation}
The score is averaged over applicable $k$-hop variable pairs, with $k=1$ by default.

\subsection{Privacy and Utility Metrics}
\label{app:privacy_metrics}

Privacy metrics evaluate three leakage granularities: row-level proximity
(\texttt{DCR}, \texttt{NNDR}), same-time-slice proximity (\texttt{CS-DCR},
\texttt{CS-NNDR}), and sequential-pattern privacy (\texttt{NGP}(n)). Higher
values are safer, except that \texttt{DCR} and \texttt{CS-DCR} are raw distances.
Let $X^{(i)}_{\rm real}$ be the numerical feature vector of real record $i$.

\noindent\textbf{DCR} and \textbf{NNDR} are standard time-agnostic nearest-neighbor
measures~\cite{syntheval,synmeter}, reported as baselines. \texttt{DCR} is the median
nearest-neighbor distance from each real record to the synthetic table:
\[
\operatorname{Median}_{i}
\left(
\min_j
\left\|
X^{(i)}_{\mathrm{real}} -
X^{(j)}_{\mathrm{synth}}
\right\|_2
\right).
\]

\noindent\textbf{NNDR} is the median ratio between the first-
and second-nearest synthetic neighbors of each real record; higher ratios imply
fewer isolated near-matches.

\noindent\textbf{CS-DCR} and \textbf{CS-NNDR} (\emph{adapted} from
\texttt{DCR}/\texttt{NNDR}~\cite{syntheval,synmeter}) restrict the same nearest-neighbor
computations to records in the same timestamp or temporal partition:
\[
\operatorname{Median}_{t\in\mathbb{T}_{\mathrm{eval}}}
\operatorname{Median}_{i:x^{time}_i=t}
\left(
\min_{j:\hat{x}^{time}_j=t}
\left\|
X^{(i)}_{\mathrm{real}} -
X^{(j)}_{\mathrm{synth}}
\right\|_2
\right),
\]
with \texttt{CS-NNDR} defined analogously using same-time-slice first- and
second-nearest neighbors.

\noindent\textbf{$n$-gramPrivacy} \texttt{NGP(n)} is new: it targets sequential rather than
row-level leakage, measuring whether discretized synthetic trajectory fragments reproduce real
length-$n$ token subsequences:
\begin{equation}
    \mathrm{NgramExposure}(n)
    =
    \frac{1}{|\hat{\mathcal{K}}|}
    \sum_{k\in\hat{\mathcal{K}}}
    \frac{1}{|G_k^{(n)}|}
    \sum_{g \in G_k^{(n)}}
    \mathbb{I}\!\left(g \in \mathcal{G}_{\mathrm{real}}^{(n)}\right),
    \label{eq:ngram_exposure}
\end{equation}
where $G_k^{(n)}$ is the set of synthetic length-$n$ token subsequences and
$\mathcal{G}_{\mathrm{real}}^{(n)}$ is the real subsequence set. We define
\begin{equation}
    \mathrm{NGP}(n) = 1 - \mathrm{NgramExposure}(n),
    \label{eq:ngp}
\end{equation}
with \texttt{NGP}(1) and \texttt{NGP}(3) distinguishing single-row from longer
within-trajectory sequential-pattern reproduction.

\medskip
\noindent\textbf{MLE-TS} \texttt{(MLE-TS)}\quad\label{app:utility_metric}
MLE-TS is \emph{adopted} from TabDiT~\cite{tabdit}. For each trajectory
$T^a(k)$, $a\in\{\mathrm{real,synth}\}$, we compute $z_k^a=\psi(T^a(k))$ and pair it with label $y_k$. For a generator $g$, a predictor $f_g$ trained on synthetic pairs is evaluated on held-out real trajectories by
\[
\mathrm{MLE\text{-}TS}(g)
=
\mathrm{AUC}\bigl(\{f_g(z_k^{\mathrm{real}})\},\{y_k^{\mathrm{real}}\}\bigr).
\]

\medskip
\noindent\textbf{MLE-Temporal} \texttt{(MLE-Temporal)}\quad
MLE-Temporal is new: it moves the TSTR~\cite{tabdit} from
trajectory-level to record-level prediction under a chronological split. For each trajectory
$T^a(k)$, $a\in\{\mathrm{real,synth}\}$, and prediction time $t$, we compute
$z_{k,t}^a=\psi(T^a(k)_{\le t})$ from the history observed up to $t$ and pair
it with label $y_{k,t}$. A predictor $f_g$ trained on synthetic temporal pairs
is evaluated on held-out real pairs from later timestamps by
\[
\mathrm{MLE\text{-}Temporal}(g)
=
\mathcal{M}\!\left(
\{f_g(z_{k,t}^{\mathrm{real}})\}_{(k,t)\in\mathcal{D}_{\mathrm{test}}^{\mathrm{real}}},
\{y_{k,t}^{\mathrm{real}}\}_{(k,t)\in\mathcal{D}_{\mathrm{test}}^{\mathrm{real}}}
\right),
\]
where $\mathcal{M}$ is AUC for classification and $R^2$ for regression.
It measures whether synthetic data preserve task-relevant historical dynamics
for time-ordered record-level prediction.
\section{Taxonomy: Formal Definitions}
\label{app:taxonomy}

\noindent\textbf{Temporal Characteristics.} Observations with \textbf{Absolute Time} operate on a globally shared calendar domain $\mathbb{T}_{global}$, so simultaneous events across trajectories share the same temporal reference: $x^{time}_{k,j} \in \mathbb{T}_{global}$. In \textbf{Relative Time}, each trajectory operates on its own lifecycle timeline, with the first observation normalized to zero: $x^{time}_{k,1} = 0\ \forall k$. A \textbf{Regular} series has a fixed inter-arrival interval $c$: $\Delta t_{k,j} = c\ \forall k, j$. An \textbf{Irregular} series has stochastic inter-arrival times drawn from a distribution $P(\Delta t)$: $\Delta t_{k,j} \sim P(\Delta t)$.

\noindent\textbf{Trajectory Dynamics.} \textbf{Independent} trajectories represent entirely isolated entities; the trajectory of one carries no information about another: $P(T(k), T(k')) = P(T(k))\,P(T(k'))$ for any $k \neq k'$. \textbf{Dependent} trajectories share macro-level dynamics, so their trajectories are statistically correlated: $\exists\, k \neq k'$ s.t.\ $P(T(k), T(k')) \neq P(T(k))\,P(T(k'))$. Evaluating trajectory dependencies requires Absolute Time, as a globally shared axis is a prerequisite for aligning observations across trajectories.

\noindent\textbf{Structural Complexity.} We characterize relational schemas by the number of 1:N foreign-key relationships each row participates in. Given parent table $T_p$, child table $T_c$, and FK map $f_{FK}: T_c \to T_p$, the child set of a parent row is $\mathcal{C}(\mathbf{r}_p) = \{\mathbf{r}_c \in T_c \mid f_{FK}(\mathbf{r}_c) = \mathbf{r}_p\}$. We distinguish four schema types \textit{Single} (no external FK; trajectory in one table), \textit{Linear} ($\forall\,\mathbf{r}_p,\,\exists!\,T_c$: $\mathcal{C}(\mathbf{r}_p) \subseteq T_c$), \textit{Multi-Child} ($\mathcal{C}^{(m)}(\mathbf{r}_p) \subseteq T_{c_m}$ for $m\in\{1,\ldots,M\}$, $M{\geq}2$), and \textit{Multi-Parent} ($f_{FK}^{(m)}(\mathbf{r}_c)=\mathbf{r}_{p_m}\in T_{p_m}$ for $m\in\{1,\ldots,M\}$, $M{\geq}2$). This schema axis introduces relational consistency as an additional evaluation requirement beyond temporal fidelity and determines which structural sub-metrics apply to each dataset.

\section{Metric Provenance}
\label{app:metric_provenance}

The metric-applicability table of Appendix~\ref{app:metrics}
(Table~\ref{tab:metric_applicability}) marks each metric as newly introduced
(unmarked), \emph{adapted} ($^\ddagger$), or \emph{adopted} ($^\S$). Table~\ref{tab:metric_provenance}
below expands those markers: for every adapted or adopted metric it names the measure it is based on,
cites the source, and states the exact modification. Formal definitions of all metrics are given in
Appendix~\ref{app:metrics}.

\begin{table*}[t]
\centering
\caption{Provenance of every metric in Seq2Synth. Type follows the markers used in the
metric-applicability table of Appendix~\ref{app:metrics} (Table~\ref{tab:metric_applicability}):
New $=$ introduced in this work, Adapted $=$ an existing measure
modified for the temporal setting, Adopted $=$ used exactly as originally defined.
\texttt{DCR} and \texttt{NNDR} are reported as time-agnostic baselines and are listed for completeness.}
\Description{A table listing every evaluation metric together with its provenance. Rows are grouped by
evaluation perspective: timestamp, cross-sectional, longitudinal, structural, privacy, and utility. For
each metric the table gives its type, which is new, adapted, or adopted; the existing measure it is
based on together with a citation where applicable; and a description of the modification made in this
work. Seven metrics are newly introduced, seventeen are adapted from existing measures, and three are
adopted without modification.}
\label{tab:metric_provenance}
\footnotesize
\setlength{\tabcolsep}{4pt}
\renewcommand{\arraystretch}{1.05}
\begin{tabular}{@{}l l p{0.22\textwidth} p{0.42\textwidth}@{}}
\toprule
\textbf{Metric} & \textbf{Type} & \textbf{Basis} & \textbf{Novelty / modification} \\
\midrule
\multicolumn{4}{l}{\textit{Timestamp}}\\
\texttt{TemporalOrderConsistency} & New & --- & Fraction of consecutive timestamp pairs that are non-decreasing \\
\texttt{TimestampUniqueness} & New & --- & Fraction of zero-length consecutive intervals, computed before deduplication \\
\texttt{TemporalRangeCompliance} & New & --- & Violation rate w.r.t.\ the real temporal support $\Omega$; under relative time the first timestamp must additionally be anchored at zero \\
\texttt{RegularityConsistency} & New & --- & Deviation from the canonical interval $c$ within tolerance $\epsilon=0.01c$ \\
\texttt{GridCompleteness} & New & --- & Compares canonical grid-point multiplicities, capping synthetic credit at the real multiplicity \\
\texttt{TimeIntervalDistributionSimilarity} & Adapted & KS complement~\cite{sdvlibrary} & Applied to the pooled inter-arrival distribution rather than to feature values \\
\texttt{TrajectoryDurationSimilarity} & Adapted & KS complement~\cite{sdvlibrary} & Applied to the distribution of trajectory durations $D_k$ \\
\midrule
\multicolumn{4}{l}{\textit{Cross-sectional}}\\
\texttt{CS-KSComplement} & Adapted & KSComplement~\cite{sdvlibrary} & Evaluated within each temporal partition and averaged \\
\texttt{CS-StatisticSimilarity} & Adapted & StatisticSimilarity~\cite{sdvlibrary} & Time-local statistics, normalized by the range of the statistic across partitions \\
\texttt{CS-RangeCoverage} & Adapted & BoundaryAdherence~\cite{sdvlibrary} & Adherence to the per-partition real value range \\
\texttt{CS-TVComplement} & Adapted & TVComplement~\cite{sdvlibrary} & Temporal lifting \\
\texttt{CS-CategoryCoverage} & Adapted & CategoryCoverage~\cite{sdvlibrary} & Per-partition category sets; partitions empty in the real data are excluded from the average \\
\texttt{CS-CorrelationSimilarity} & Adapted & CorrelationSimilarity~\cite{sdvlibrary} & Correlations compared within each partition \\
\texttt{CS-ContingencySimilarity} & Adapted & ContingencySimilarity~\cite{sdvlibrary} & Joint category frequencies compared within each partition \\
\midrule
\multicolumn{4}{l}{\textit{Longitudinal}}\\
\texttt{FirstDifferenceKSComplement} & Adapted & Two-sample KS complement~\cite{sdvlibrary} & Applied to pooled first differences $\Delta x$ rather than raw values \\
\texttt{TransitionMatrixTVDComplement} & Adapted & Total variation complement~\cite{sdvlibrary} & Per-timestep conditional transition matrices, averaged over valid transition times \\
\texttt{AutoCorrelationSimilarity} & Adapted & Autocorrelation function~\cite{fints} & Computation mode branches on trajectory dependence: per-trajectory and averaged for independent trajectories, on the cross-sectional mean trajectory for dependent ones \\
\texttt{TT-WassersteinDistance} & Adapted & Li et al.~\cite{ttwasserstein} & Trajectory embedding and optimal-transport cost used as defined; reported here as a raw distance and converted to $1-\mathrm{TT\text{-}W}$ when aggregated into the longitudinal dimension \\
\texttt{MLD-TS} & Adopted & Garuti et al.~\cite{tabdit} & Used as defined \\
\midrule
\multicolumn{4}{l}{\textit{Structural}}\\
\texttt{SequenceLengthSimilarity} & Adopted & SDMetrics~\cite{sdvlibrary} & Used as defined \\
\texttt{TemporalCardinalityShapeSimilarity} & Adapted & CardinalityShapeSimilarity~\cite{syntherela} & Global child-count distribution extended to child counts within aligned temporal windows \\
\texttt{DynamicKHopCorrelationSimilarity} & Adapted & $k$-hop correlation similarity~\cite{syntherela} & Static cross-table correlation extended to the time average of $\rho(t)$ \\
\midrule
\multicolumn{4}{l}{\textit{Privacy}}\\
\texttt{CS-DCR} & Adapted & \texttt{DCR}~\cite{syntheval,synmeter} & Nearest-neighbor search restricted to the same time slice; median over slices of within-slice medians \\
\texttt{CS-NNDR} & Adapted & \texttt{NNDR}~\cite{syntheval,synmeter} & First- and second-nearest neighbors taken within the same time slice \\
\texttt{NGP}$(n)$ & New & --- & Exposure defined as the reproduction rate of real length-$n$ token subsequences over discretized trajectories; $\mathrm{NGP}(n)=1-\mathrm{NgramExposure}(n)$ \\
\midrule
\multicolumn{4}{l}{\textit{Utility}}\\
\texttt{MLE-TS} & Adopted & Garuti et al.~\cite{tabdit} & Used as defined \\
\texttt{MLE-Temporal} & New & MLE protocol~\cite{tabdit} & Record-level rather than trajectory-level prediction, chronological split, embedding built only from history observed up to $t$ \\
\bottomrule
\end{tabular}
\end{table*}

\section{Temporal Post-processing Pipeline Overview}
\label{app:online_postprocessing}

\begin{figure*}[h]
    \centering
    \includegraphics[width=\textwidth]{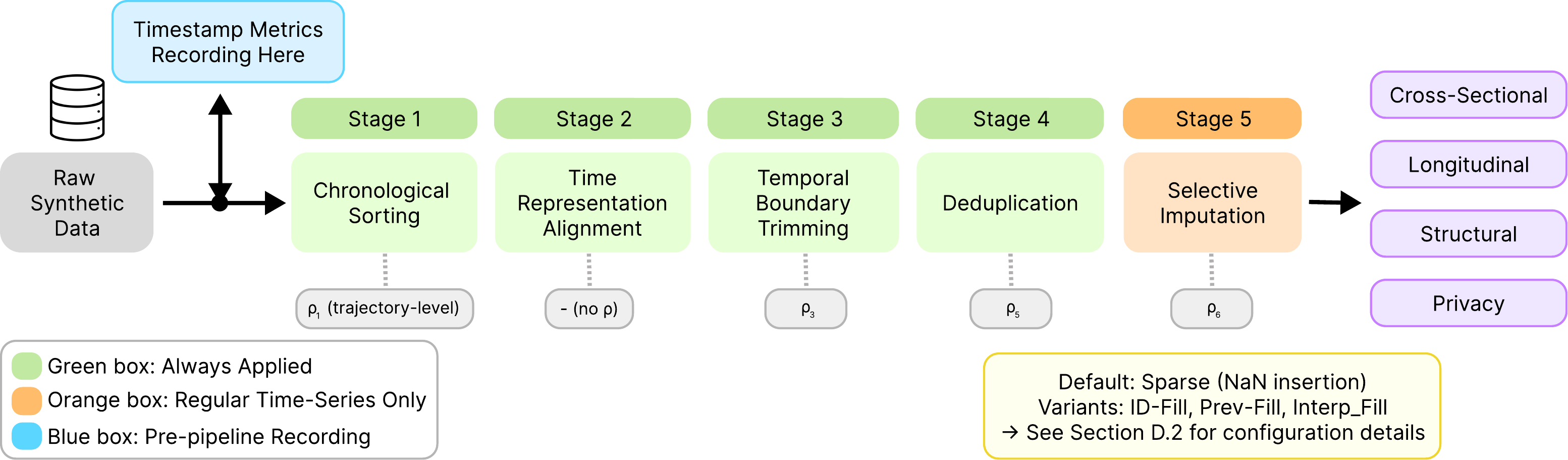}
    \caption{Detailed overview of the temporal post-processing pipeline. This deterministic five-stage process is applied to generative model outputs before non-timestamp fidelity evaluation.}
    \Description{A pipeline diagram showing the temporal post-processing process applied to raw synthetic data before non-timestamp fidelity evaluation. The pipeline first records timestamp metrics before processing, then applies five stages in order: chronological sorting, time representation alignment, temporal boundary trimming, deduplication, and selective imputation. Green stages are always applied, the orange stage which is only for stage 5 is applied only to regular time-series data, and the blue box indicates pre-pipeline timestamp metric recording. The processed data are then used for cross-sectional, longitudinal, structural, and privacy evaluation.}
    \label{fig:postprocessing_online}
\end{figure*}

We apply a deterministic five-stage pipeline to repair temporal violations such as chronological ordering and calendar alignment before computing non-timestamp metrics. Specifically, the pipeline sorts records chronologically, aligns timestamps to the prescribed grid when applicable, trims out-of-range records, resolves duplicate entity--timestamp pairs, and optionally imputes missing grid points in regular series. Figure~\ref{fig:postprocessing_online} provides an overview of this sequence, illustrating how raw synthetic outputs are progressively sorted, aligned, trimmed, deduplicated, and optionally imputed.

\subsection{Detailed Pipeline Stages}
\label{app:pipeline_stages}
Generative models that treat timestamps as ordinary features frequently emit outputs that violate the preconditions assumed by non-timestamp metrics: chronological ordering, calendar alignment, and gap-free trajectories.
We apply a deterministic five-stage pipeline to repair these violations before computing non-timestamp metrics; timestamp metrics, by contrast, are computed on the raw synthetic output so that violations are not masked by the corrections themselves.
At each stage we log the intervention ratio $\rho_s = (N^{\text{mod}}_s + N^{\text{drop}}_s)/N^{\text{in}}_s$, the fraction of records modified or dropped, as an auxiliary signal of temporal-coherence failure.

\paragraph{Stages (in dependency order).}
\textbf{(1) Sort}: records sorted ascending by $(x^{key}, x^{time})$.
\textbf{(2) Time Alignment}: synthetic timestamps snapped to the canonical grid at interval $c$; non-calendar outputs are anchored at the earliest real period and assigned $c$-spaced increments.
\textbf{(3) Temporal Boundary Trimming}: records outside the observed support $[T_{\min}, T_{\max}]$ are removed.
\textbf{(4) Deduplication}: duplicate $(x^{key}, x^{time})$ pairs are resolved by retaining the first occurrence.
\textbf{(5) Selective Imputation} (regular series only): unoccupied grid points within each trajectory's observed range are filled according to the chosen variant; points outside this range are not imputed.

\paragraph{Method variants.}
Stages (1)--(4) are deterministic; the four variants differ only in Stage~(5)'s imputation: \textsc{sparse} (default; insert as \textsc{NaN}, exposing the model's coverage pattern as a measurable property); \textsc{id-fill} (within-trajectory mean/mode); \textsc{prev-fill} (forward-fill, backward-fill for leading gaps); \textsc{interp-fill} (linear interpolation for numerical, forward/backward fill for categorical).

\subsection{Recommended Configurations}
\label{app:recommended_settings}
Among the pipeline steps, only Stages~(2) and (5) require configuration.
For Stage~2, calendar-aligned synthetic timestamps are anchored to canonical period boundaries (e.g., first day of the month) rather than nearest-neighbor rounding, guaranteeing membership in the grid required by \texttt{GridCompleteness}.
For Stage~5, we strongly recommend the \textsc{sparse} variant as the default.
Standard imputations (\textsc{id-fill}, \textsc{prev-fill}, \textsc{interp-fill}) inject in-distribution values, mechanically inflating fidelity scores and concealing a model's failure to maintain temporal coverage.
While these fill variants serve as useful ablation baselines for evaluating feature-level fidelity \emph{conditional} on complete coverage (detailed in Appendix~\ref{app:postprocessing-ablation}), the intervention ratio $\rho_5$ must always be reported alongside metric scores to ensure transparency.

\section{Experimental Details}
\label{app:experimental_details_online}

\subsection{Baseline Training Setup \& Hardware}
\label{app:baseline_preprocessing}
To ensure fair reproducibility, all generative models were trained using default hyperparameters and configurations from their official repositories or the SyntheRela benchmark~\cite{syntherela}.
For baseline dataset preprocessing, we applied a uniform pipeline: numerical missing values were mean-imputed (rounded for integers), categorical gaps were filled with a placeholder (`?'), and hash-based IDs were globally integerized to maintain relational integrity.
Excessively large datasets were subsampled at the trajectory (primary key) level for computational feasibility.
All experiments were executed on a server equipped with NVIDIA RTX 6000 Ada GPUs.

\subsection{Imputation Sensitivity Analysis}
\label{app:postprocessing-ablation}

We conduct an ablation study to empirically validate that the choice of
imputation strategy (Stage~5) does not materially alter the comparative
conclusions drawn from Seq2Synth. Specifically, we compare four variants ---
\textsc{sparse} (default, missing grid points left as \textsc{NaN}),
\textsc{id-fill} (within-trajectory mean/mode), \textsc{prev-fill}
(forward-fill with backward-fill for leading gaps), and
\textsc{interp-fill} (linear interpolation for numerical; forward/backward
fill for categorical) --- across representative regular datasets, Rossmann and Walmart.
Imputation variants are applicable only to regular-series datasets; for
irregular-series datasets (Airbnb, Berka) and the PTB-XL ECG dataset, which
involve independent trajectories or sparse event structures, the evaluation
reduces to the \textsc{sparse} baseline and a single ``default'' point is
shown.

\begin{figure}[htbp]
    \centering
    \includegraphics[width=\linewidth]{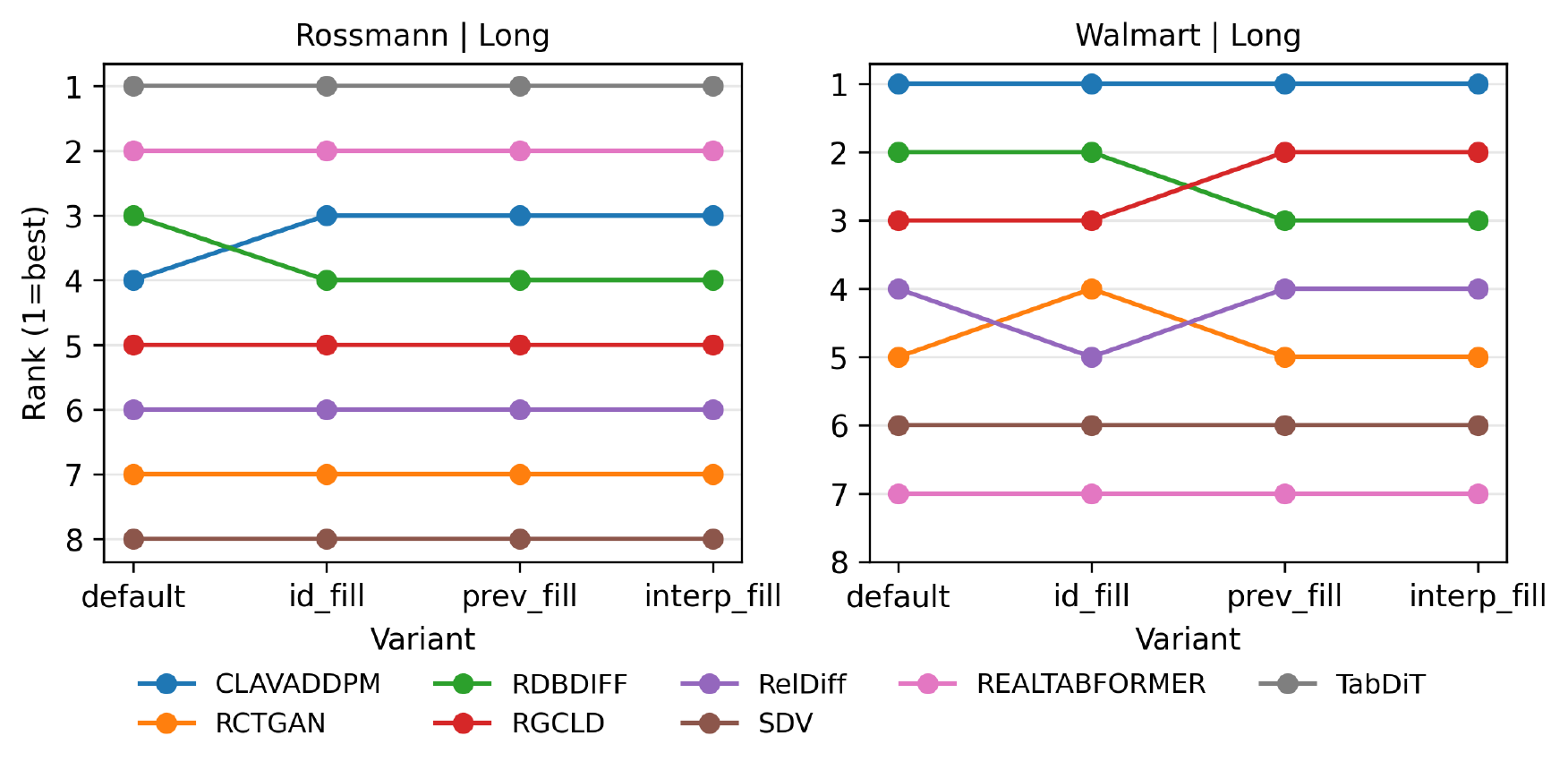}
    \caption{
    Model ranks under different post-processing imputation variants on Rossmann and Walmart.
    Each colored line traces one model across imputation variants.
    The rank ordering is largely stable across default sparse insertion, ID-fill, previous-fill, and interpolation-fill variants, suggesting that the choice of imputation strategy has limited impact on comparative model evaluation.
    }
    \Description{
    A line-plot figure showing model ranks across post-processing imputation variants.
    The x-axis lists imputation variants, including default, ID-fill, previous-fill, and interpolation-fill, and the y-axis reports model rank with 1 as the best rank.
    Each colored line represents a synthetic data generation model.
    The visible panels show Rossmann and Walmart under the longitudinal metric group.
    Most lines remain flat or change only slightly across variants, indicating that model rankings are largely stable under different imputation strategies.
    }
    \label{fig:imputation_sensitivity}
\end{figure}

\paragraph{Protocol.} For each dataset $d$, metric group $m \in
\{\text{SD}, \text{CS}, \text{Long}, \text{Struct}\}$, and imputation
variant $v$, we compute each model's \emph{rank} (1 = best) within the
set of evaluated models. Model ranks are then tracked across imputation
variants. A stable rank profile indicates that the relative ordering of
models --- and thus the benchmark's comparative conclusions --- is insensitive
to the imputation choice. Figure~\ref{fig:imputation_sensitivity} reports the
longitudinal group, which is the group most directly exposed to Stage~5
because it is the only dimension evaluated on interpolation-filled inputs
(Section~\ref{sec:benchmark_protocol}).

\paragraph{Results.}
Figure~\ref{fig:imputation_sensitivity} shows model ranks on Rossmann and
Walmart under all four imputation variants. The key observation
is that rank orderings remain largely stable across variants. Rank crossings, when they occur,
are almost exclusively between models that are already closely ranked under
the default setting, reflecting genuine measurement noise rather than a
systematic artefact of imputation. No imputation variant consistently
elevates or demotes a specific model across datasets, and models that
rank near the top (or bottom) under \textsc{sparse} tend to maintain their
relative position under all three fill-based alternatives.

On Rossmann, only ClavaDDPM and RDBDiff exchange the third and fourth
positions when moving away from \textsc{sparse}; every other model keeps its
rank across all four variants. On Walmart, the crossings are confined to two
adjacent pairs (RDBDiff/RGCLD and RelDiff/RCTGAN), while the top and bottom of
the ordering is unchanged. This indicates that the effect of filling missing
grid points with in-distribution values does not substantially inflate or
suppress any model's score relative to its competitors. The result is
consistent with our theoretical expectation: because all models are subject to
the same imputation procedure, the filling introduces a uniform bias that
largely cancels in comparative analysis.

The remaining datasets do not admit this ablation at all. Stage~5 operates on
canonical grid points, so it is defined only for regular series; the
irregular-series datasets (Airbnb, Berka) and the independent-trajectory
dataset PTB-XL are therefore always evaluated under \textsc{sparse}, and
their ranks are determined solely by the model's ability to generate the
correct temporal support. Imputation is consequently not a confound for these
data types.

\paragraph{Implications.}
These results jointly support two claims. First, the
\textsc{sparse} default is empirically justified: adopting any of the
three fill-based alternatives leads to the same comparative conclusions
while introducing unnecessary feature-level bias as discussed in
Appendix~\ref{app:recommended_settings}. Second, the metrics
themselves are robust: they quantify genuine model-level differences in
temporal fidelity rather than artefacts of the post-processing pipeline.
Together, these findings validate both the imputation design choice and the
metric computations reported in Section~\ref{sec:results}.

\subsection{Detailed Dataset Statistics}
\label{app:dataset_stats_online}

To provide a comprehensive operational overview of our evaluation testbed, Table~\ref{tab:dataset_statistics} summarizes the structural dimensions of the 13 sequential tabular datasets, including the total number of relational tables, absolute rows, schema columns, unique entity trajectories, and the resulting average trajectory length.

\begin{table}[htbp]
\centering
\caption{Detailed Statistics of Sequential Tabular Datasets}
\Description{A statistics table summarizing the sequential tabular datasets used in the benchmark. Each row corresponds to a dataset, and columns report the number of tables, rows, columns, trajectories, and average trajectory length. The table shows substantial variation in dataset scale and temporal structure, ranging from small datasets with a few dozen trajectories to large datasets with hundreds of thousands of rows or long average trajectories.}
\label{tab:dataset_statistics}
\begin{small}
\resizebox{\linewidth}{!}{%
\begin{tabular}{@{}lrrrrr@{}}
\toprule
\textbf{Dataset} & \textbf{\# tables} & \textbf{\# rows} & \textbf{\# columns} & \textbf{\# trajectories} & \textbf{avg traj. length} \\ \midrule
\textbf{Walmart}          & 3 & 15,317     & 20  & 45      & 334.38  \\
\textbf{PTB-XL}           & 2 & 101,000    & 35  & 1,000   & 100     \\
\textbf{FreddieMAC}       & 2 & 536,578    & 39  & 43,966  & 11.20   \\
\textbf{FannieMAE}        & 2 & 2,992,782  & 47  & 183,777 & 15.28   \\
\textbf{H\&M}             & 3 & 297,879    & 36  & 10,000  & 24.03   \\
\textbf{Coupon}           & 5 & 259,833    & 38  & 22,873  & 5.41    \\
\textbf{Google}           & 5 & 756,167    & 47  & 498     & 1,019.20 \\
\textbf{Home Credit}      & 7 & 1,263,300  & 218 & 10,000  & 36.35   \\
\textbf{Rossmann}         & 2 & 13,826     & 20  & 223     & 61.00   \\
\textbf{AirBnB}           & 2 & 40,281     & 32  & 1,942   & 19.74   \\
\textbf{Berka}            & 2 & 45,666     & 19  & 900     & 49.74   \\
\textbf{CMAPSS}           & 1 & 160,359    & 26  & 709     & 226.18  \\
\textbf{Citi Bike}        & 1 & 559,644    & 15  & 5,793   & 96.61   \\ \bottomrule
\end{tabular}%
}
\end{small}
\end{table}

\subsection{Per-Dataset Description}
\label{app:dataset_description_online}

We evaluate our benchmark on a diverse set of real-world sequential tabular
datasets, selected to cover a wide range of temporal characteristics,
relational structures, and application domains. Their taxonomy is given in
Table~\ref{tab:dataset_taxonomy} of the main text and their row-level
statistics in Table~\ref{tab:dataset_statistics}
(Appendix~\ref{app:dataset_stats_online}).

\textbf{Walmart}
We use the subsampled Walmart dataset from SyntheRela~\cite{syntherela}. It exhibit regular, absolute timestamps with fixed-length trajectories.

\textbf{H\&M Personalized Fashion Recommendations.}
H\&M~\cite{hm_kaggle} contains customer purchase records over time, along with auxiliary information on users and items. It features a relational schema with two parent tables and one child table, posing challenges for models that cannot handle multi-parent dependencies (e.g., RTF).

\textbf{Coupon Purchase Prediction.}
The Coupon dataset~\cite{coupon} includes user profiles, coupon metadata, browsing logs, and purchase events. It has a multi-parent, multi-child relational structure with 2-hop dependencies, making it suitable for evaluating relational consistency across multiple tables.

\textbf{PTB-XL.}
PTB-XL~\cite{ptbxl} consists of electrocardiography (ECG) signals collected at high temporal resolution. It is suitable for evaluating synthetic data generation in the healthcare domain, where privacy preservation and realistic temporal modeling are critical. Each record forms a fixed-length time series, allowing evaluation of models on dense and high-frequency temporal data.

\textbf{Freddie Mac and Fannie MAE.}
The Freddie Mac dataset~\cite{freddiemac} provides loan-level mortgage performance data spanning multiple decades while Fannie MAE~\cite{fanniemae} contains detailed records of single-family mortgage loans. 

\textbf{Home Credit Default Risk.}
The Home Credit dataset~\cite{homecredit} provides transactional and application data for credit risk prediction. Notably, the application table contains over 120 features, allowing evaluation of high-dimensional tabular generation.

\textbf{Google Cluster Traces v2.}
This dataset~\cite{googlecluster} records task execution traces from a large-scale cluster system. Time is represented in relative units (600-second intervals), and each task forms a temporal sequence, enabling evaluation under system-level workload dynamics.

\textbf{CMAPSS.}
CMAPSS~\cite{cmapss} is a simulated dataset of aircraft engine degradation trajectories. Each sequence evolves over relative time cycles, making it suitable for evaluating temporal progression and failure patterns.

\textbf{Citi Bike NYC.}
The Citi Bike dataset~\cite{citibike_nyc} contains trip-level records with absolute timestamps. Each trajectory corresponds to a bike usage session, allowing evaluation of real-world mobility patterns with irregular temporal spacing.

\textbf{AirBnB, Berka and Rossmann}
The AirBnB dataset~\cite{tabdit} includes temporally ordered reservation records together with static listing attributes, enabling evaluation of models on relational data with heterogeneous temporal patterns and sparse event occurrences. The Berka dataset~\cite{tabdit} is a financial relational dataset consisting of customer accounts, transactions, loans, and demographic information. Transaction histories evolve over absolute time and exhibit long-term sequential dependencies, making the dataset suitable for evaluating temporal consistency and relational coherence in synthetic financial records. The Rossmann dataset~\cite{tabdit} contains daily sales records from retail stores along with static store metadata and promotion-related information. Each store forms a fixed-frequency temporal trajectory with regular daily intervals, allowing evaluation of seasonal patterns, trend preservation, and long-range temporal dependencies in retail forecasting scenarios.

\subsection{Preprocessing Details per Dataset}
\label{app:preprocessing_online}
Baseline missing-value preprocessing is described in Appendix~\ref{app:baseline_preprocessing}; temporal post-processing is described in Appendix~\ref{app:online_postprocessing}.
\paragraph{Retail Domain.}
We utilize the subsampled version Walmart provided by SyntheRela~\cite{syntherela}. For Rossmann, AirBnB, and Berka, we consider test-set splits that authors of TabDiT~\cite{tabdit} provided are used as the real datasets for evaluating all seven baseline models, while TabDiT-generated samples are directly used. For H\&M, we randomly subsample 10,000 customers from the original customer IDs. Image-related information is excluded, and only the corresponding subsampled article metadata is retained in the articles table.
\paragraph{Medical Domain.}
PTB-XL is preprocessed using the official pipeline provided by the source dataset and then decomposed into a parent-child relational structure, where metadata is stored in the parent table and ECG signals are stored in the child table. We use the 100Hz recordings, for which each \texttt{ecg\_id} originally contains 1,000 temporal observations. Since generated data may not reliably preserve microsecond-level temporal precision, we introduce a relative timestep column (\texttt{step}) to represent temporal ordering explicitly. We further apply temporal downsampling to reduce each trajectory to 100 timesteps for more stable and computationally efficient benchmarking. A total of 1,000 \texttt{ecg\_id}s are randomly subsampled. Columns with high missing-value ratios are removed to improve robustness.
\paragraph{Finance Domain.}
For Freddie Mac and Fannie Mae, we use loan records reported after 2024. In the case of Fannie Mae, the original single-table structure is transformed into a relational schema by separating static loan attributes into a parent table and time-varying records into a child table. For Home Credit Default Risk, 10,000 IDs are randomly subsampled from the full dataset.
\paragraph{Other Domains.}
For Google Cluster Traces v2, we use the first five shards and randomly subsample 500 machine IDs. To establish a relational schema with a designated primary entity, static job-level attributes are separated from event logs to construct a root table named \texttt{jobs}. This restructuring is not applicable to machine events because they contain only static information; therefore, the machine events and machine attribute tables are excluded. For CMAPSS, the four simulation subsets are merged into a single unified table.

\subsection{MLD-TS and MLE-TS Evaluation Details}
\label{app:details_tab}

\begin{table}[h]
\centering
\caption{XGBoost hyperparameter grid, with one configuration sampled per seed.}
\Description{A table listing the XGBoost hyperparameter search grid used in the utility evaluation. The table includes four hyperparameters: maximum tree depth, learning rate, minimum child weight, and number of estimators. One configuration is sampled from the listed candidate values for each experimental seed.}
\label{tab:xgb_hparams}
\begin{tabular}{lc}
\toprule
\textbf{Hyperparameter} & \textbf{Value} \\
\midrule
\texttt{max\_depth} & $\{3, 5, 7, 9\}$ \\
\texttt{learning\_rate} & $\{0.01, 0.03, 0.05, 0.10\}$ \\
\texttt{min\_child\_weight} & $\{1, 3\}$ \\
\texttt{n\_estimators} & $\{50\}$ \\
\bottomrule
\end{tabular}
\end{table}

\texttt{MLD-TS} and \texttt{MLE-TS} are adopted without modification from Garuti et
al.~\cite{tabdit}, which introduced them for tabular time-series generation; we reuse their
trajectory embedding, detector/predictor protocol, and reporting convention, and only fix the
backbone and hyperparameter grid described here.
We used XGBoost~\cite{xgboost} for both MLD-TS and MLE-TS evaluations. 
In MLD-TS, real and synthetic samples were combined and then split into training and test sets with a ratio of 8:2. The model was trained as a binary classifier to distinguish real samples from synthetic samples, and accuracy was used as the evaluation metric. In MLE-TS, the training set was constructed using only synthetic data, while the test set consisted only of real data, following the same 8:2 split ratio. For each random seed, hyperparameters were randomly sampled from a predefined search space summarized in Table~\ref{tab:xgb_hparams}.


\paragraph{Downstream Tasks for MLE-TS.}
Since MLE-TS evaluates whether synthetic data preserves the predictive structure of real data, it requires a target label for each sample. For datasets where the original target label is not directly available, we define binary prediction tasks using attributes from the corresponding parent, store, customer, or entity-level tables. Specifically:

\begin{itemize}
    \item In Airbnb, we define the task \texttt{n\_sessions\_ge\_17}. The target label is set to 1 if \texttt{n\_sessions} is greater than or equal to 17, and 0 otherwise.

    \item In Berka, we define the task \texttt{region\_is\_moravia}. The target label is set to 1 if the customer's region is either South Moravia or North Moravia, and 0 otherwise.

    \item In Citi Bike, we define the task \texttt{mode\_gender\_is\_1}. For each bike, we compute the mode of \texttt{gender} across its records and set the target label to 1 if the modal value is 1, and 0 otherwise.

    \item In CMAPSS, we define the task \texttt{final\_s11\_high\_regime}. For each unit, we use the value of \texttt{s\_11} at the final time step and predict whether it is greater than or equal to the median value computed from the real data.

    \item In Coupon, we define the task \texttt{sex\_is\_m}. The target label is set to 1 if \texttt{sex} is male, and 0 otherwise.

    \item In Fannie Mae, and Freddie Mac, we define the task \texttt{\seqsplit{property\_type\_is\_4}}. The target label is set to 1 if the corresponding property-type field is equal to 4, and 0 otherwise.

    \item In Google Cluster, we define the task \texttt{\seqsplit{scheduling\_class\_ge\_2}}. The target label is set to 1 if \texttt{scheduling\_class} is greater than or equal to 2, and 0 otherwise.

    \item In H\&M, we use the binary attribute \texttt{\seqsplit{fashion\_news\_regularly}} as the target label.

    \item In Home Credit, we use the binary attribute \texttt{TARGET} as the target label.

    \item In PTB-XL, we use the binary attribute \texttt{sex} as the target label.

    \item In Rossmann, we use the binary attribute \texttt{Promo2} from the store table. The target label is set to 1 if \texttt{Promo2} is equal to 1, and 0 otherwise.

    \item In Walmart, we define the task \texttt{\seqsplit{store\_type\_is\_A}}. The target label is set to 1 if the store type \texttt{Type} is ``A'', and 0 otherwise.
\end{itemize}

\section{Additional Results}
\label{app:addition_results}
\subsection{Results on Additional Dataset}
\label{app:addition-dataset-result}

\noindent\textbf{Additional results on single-table and complex datasets.}
Tables~\ref{tab:single_table_results} and~\ref{tab:complex_data_results}
report the remaining dataset-level results beyond the seven core datasets, and
Table~\ref{tab:per-dataset-all} consolidates the dimension-level scores of all
13 benchmark datasets in a single view. Trajectory-level results for the
fixed-length datasets are reported separately in
Appendix~\ref{app:tt_wasserstein}.

\begin{table}[htbp]
\vspace{-0.1em}
\caption{Results for single table datasets}
\Description{A results table reporting fidelity scores for single-table datasets, Citi Bike and CMAPSS. Rows compare synthetic data generation models, and columns report five evaluation dimensions: statistical distribution, timestamp fidelity, cross-sectional fidelity, longitudinal fidelity, and structural fidelity. Scores range from zero to one, with higher values indicating better fidelity. The table shows that model performance differs across the two single-table datasets and across temporal fidelity dimensions.}
\label{tab:single_table_results}
\begin{adjustbox}{width=\columnwidth}
\begin{tabular}{l|ccccc|ccccc}
\toprule
\multirow{2}{*}{\textbf{Model}}
  & \multicolumn{5}{c|}{\textbf{Citi Bike}}
  & \multicolumn{5}{c}{\textbf{CMAPSS}} \\
\cmidrule(lr){2-6} \cmidrule(lr){7-11}
  & \textsc{sd} & \textsc{ts} & \textsc{cs} & \textsc{lg} & \textsc{st}
  & \textsc{sd} & \textsc{ts} & \textsc{cs} & \textsc{lg} & \textsc{st} \\
\midrule
\textbf{ClavaDDPM}
  & $\mathbf{0.990}$ & $\underline{0.769}$ & $\mathbf{0.941}$ & $\underline{0.965}$ & 0.908
  & 0.963 & 0.575 & -- & 0.471 & 0.709 \\
\textbf{CPAR}
  & 0.706 & 0.721 & 0.614 & 0.811 & 0.271
  & 0.811 & $\mathbf{0.969}$ & -- & $\underline{0.573}$ & 0.836 \\
\textbf{RCTGAN}
  & $\underline{0.943}$ & 0.765 & $\underline{0.900}$ & 0.960 & $\mathbf{0.979}$
  & 0.813 & 0.369 & -- & 0.459 & 0.330 \\
\textbf{RTF}
  & 0.870 & 0.678 & 0.885 & $\mathbf{0.967}$ & 0.676
  & $\underline{0.972}$ & 0.322 & -- & $\mathbf{0.618}$ & 0.000 \\
\textbf{SDV}
  & 0.883 & $\mathbf{0.782}$ & 0.848 & 0.950 & $\mathbf{0.979}$
  & 0.916 & 0.545 & -- & 0.425 & $\mathbf{0.961}$ \\
\textbf{TabARGN}
  & 0.890 & 0.201 & 0.757 & 0.921 & 0.154
  & $\mathbf{0.983}$ & $\underline{0.578}$ & -- & 0.485 & $\underline{0.917}$ \\
\bottomrule
\end{tabular}
\end{adjustbox}
\end{table}
\vspace{-0.5em}

The single-table datasets confirm that temporal fidelity remains
multi-dimensional even without explicit inter-table relationships. On Citi
Bike, ClavaDDPM leads static and cross-sectional fidelity, RTF leads
longitudinal fidelity, and RCTGAN/SDV achieve the strongest structural
scores. On CMAPSS, TabularARGN obtains the best static-distribution score,
whereas CPAR, RTF, and SDV lead timestamp, longitudinal, and structural
fidelity, respectively. Thus, even in single-table settings, no single
dimension summarizes overall temporal quality.

\begin{table}[htbp]
\caption{Results for complex datasets}
\Description{A results table reporting fidelity scores for four complex datasets: Coupon, Google Cluster, H\&M, and Home Credit. Rows compare synthetic data generation models, and columns report statistical distribution, timestamp fidelity, cross-sectional fidelity, longitudinal fidelity, and structural fidelity. Scores range from zero to one, with higher values indicating better fidelity. The table shows that model rankings vary substantially across complex relational and temporal datasets, indicating that performance depends on both data structure and fidelity dimension.}
\label{tab:complex_data_results}

\begin{adjustbox}{width=\columnwidth, max totalheight=0.3\textheight}
\begin{tabular}{l|ccccc|ccccc}
\toprule
\multirow{2}{*}{\textbf{Model}}
  & \multicolumn{5}{c|}{\textbf{Coupon}}
  & \multicolumn{5}{c}{\textbf{Google Cluster}} \\
\cmidrule(lr){2-6} \cmidrule(lr){7-11}
  & \textsc{sd} & \textsc{ts} & \textsc{cs} & \textsc{lg} & \textsc{st}
  & \textsc{sd} & \textsc{ts} & \textsc{cs} & \textsc{lg} & \textsc{st} \\
\midrule
\textbf{ClavaDDPM}
  & $\underline{0.895}$ & 0.828 & $\mathbf{0.801}$ & 0.636 & 0.974
  & $\underline{0.863}$ & 0.604 & -- & 0.481 & 0.977 \\
\textbf{RCTGAN}
  & 0.894 & 0.817 & $\underline{0.798}$ & 0.626 & 0.930
  & 0.715 & 0.574 & -- & 0.457 & 0.708 \\
\textbf{RDBDiff}
  & $\mathbf{0.976}$ & $\mathbf{0.896}$ & 0.789 & 0.668 & $\mathbf{0.987}$
  & $\mathbf{0.935}$ & 0.627 & -- & $\underline{0.507}$ & $\mathbf{1.000}$ \\
\textbf{RGCLD}
  & 0.863 & 0.606 & 0.737 & 0.674 & 0.949
  & 0.706 & $\mathbf{0.650}$ & -- & $\mathbf{0.516}$ & 0.986 \\
\textbf{RelDiff}
  & 0.881 & $\underline{0.848}$ & 0.779 & $\mathbf{0.908}$ & $\underline{0.979}$
  & 0.781 & $\underline{0.645}$ & -- & 0.455 & $\mathbf{1.000}$ \\
\textbf{SDV}
  & 0.827 & 0.805 & 0.474 & $\underline{0.837}$ & 0.895
  & 0.599 & 0.630 & -- & 0.486 & 0.567 \\
\bottomrule
\end{tabular}
\end{adjustbox}

\vspace{0.1em}

\begin{adjustbox}{width=\columnwidth, max totalheight=0.3\textheight}
\begin{tabular}{l|ccccc|ccccc}
\toprule
\multirow{2}{*}{\textbf{Model}}
  & \multicolumn{5}{c|}{\textbf{H\&M}}
  & \multicolumn{5}{c}{\textbf{Home Credit}} \\
\cmidrule(lr){2-6} \cmidrule(lr){7-11}
  & \textsc{sd} & \textsc{ts} & \textsc{cs} & \textsc{lg} & \textsc{st}
  & \textsc{sd} & \textsc{ts} & \textsc{cs} & \textsc{lg} & \textsc{st} \\
\midrule
\textbf{ClavaDDPM}
  & $\mathbf{0.933}$ & $0.680$ & $0.820$ & $\underline{0.873}$ & $\underline{0.640}$
  & $0.671$ & $0.628$ & -- & $0.916$ & $0.953$ \\
\textbf{RCTGAN}
  & 0.626 & $\mathbf{0.709}$ & 0.521 & 0.715 & 0.555
  & 0.850 & 0.611 & -- & 0.888 & 0.870 \\
\textbf{RDBDiff}
  & $\underline{0.891}$ & 0.682 & 0.789 & 0.756 & 0.638
  & $\mathbf{0.896}$ & \underline{0.690} & -- & $\underline{0.935}$ & $\underline{0.996}$ \\
\textbf{RGCLD}
  & 0.871 & $\underline{0.704}$ & $\mathbf{0.874}$ & $\mathbf{0.876}$ & 0.640
  & $\underline{0.860}$ & $\mathbf{0.719}$ & -- & $\mathbf{0.937}$ & 0.994 \\
\textbf{RelDiff}
  & 0.871 & 0.688 & $\underline{0.842}$ & 0.772 & 0.640
  & 0.826 & 0.671 & -- & 0.910 & $\mathbf{1.000}$ \\
\textbf{SDV}
  & 0.641 & 0.581 & 0.456 & 0.680 & $\mathbf{0.648}$
  & -- & -- & -- & -- & -- \\
\bottomrule
\end{tabular}
\end{adjustbox}
\vspace{-0.1em}

\end{table}

The complex-schema datasets further amplify this pattern. On Coupon, RDBDiff
dominates static, timestamp, and structural fidelity, while RelDiff achieves
the best longitudinal score. On Google Cluster, diffusion-based relational
models preserve structural consistency well, but their longitudinal scores
remain moderate. On H\&M and Home Credit, graph- and diffusion-based methods
such as RGCLD, RDBDiff, and RelDiff generally perform strongly, but the best
model still changes across dimensions. These results show that complex
schemas introduce additional failure modes: preserving relational structure
does not guarantee faithful timestamp generation or trajectory evolution, and
static-distribution fidelity remains insufficient as a proxy for temporal
quality.

\begin{table*}[thbp]
\centering
\caption{Per-dataset fidelity scores across all benchmark datasets and five
evaluation dimensions. All scores lie in $[0,1]$, with higher values indicating
better fidelity. ``--'' denotes inapplicable cases. Bold and underline mark the
best and second-best scores per dataset--metric row.}
\label{tab:per-dataset-all}

\footnotesize
\setlength{\tabcolsep}{2.0pt}
\renewcommand{\arraystretch}{1.06}

\begin{tabular*}{0.98\textwidth}{@{\extracolsep{\fill}}llcccccccccc@{}}
\toprule
\textbf{Dataset}
& \textbf{Metric}
& \textbf{ClavaDDPM}
& \textbf{RCTGAN}
& \textbf{RDBDiff}
& \textbf{RelDiff}
& \textbf{RGCLD}
& \textbf{RTF}
& \textbf{SDV}
& \textbf{TabDiT}
& \textbf{CPAR}
& \textbf{TabARGN} \\
\midrule

\multirow[c]{5}{*}{\textbf{Rossmann}}
& \textsc{sd} & 0.850 & 0.810 & $\underline{0.858}$ & 0.799 & 0.821 & 0.809 & 0.734 & $\mathbf{0.932}$ & -- & -- \\
& \textsc{ts} & 0.607 & 0.562 & 0.608 & 0.602 & 0.598 & $\underline{0.823}$ & 0.565 & $\mathbf{0.824}$ & -- & -- \\
& \textsc{cs} & $\underline{0.912}$ & 0.808 & $\underline{0.912}$ & 0.817 & 0.897 & 0.882 & 0.635 & $\mathbf{0.965}$ & -- & -- \\
& \textsc{lg} & 0.917 & 0.872 & 0.915 & 0.882 & 0.902 & $\underline{0.948}$ & 0.789 & $\mathbf{0.975}$ & -- & -- \\
& \textsc{st} & $\mathbf{1.000}$ & $\mathbf{1.000}$ & $\mathbf{1.000}$ & $\mathbf{1.000}$ & $\mathbf{1.000}$ & $\mathbf{1.000}$ & $\mathbf{1.000}$ & $\mathbf{1.000}$ & -- & -- \\

\midrule

\multirow[c]{5}{*}{\textbf{Berka}}
& \textsc{sd} & 0.948 & 0.865 & $\mathbf{0.992}$ & 0.898 & 0.922 & $\underline{0.951}$ & 0.695 & 0.909 & -- & -- \\
& \textsc{ts} & 0.594 & 0.620 & 0.759 & 0.603 & 0.731 & $\mathbf{0.907}$ & 0.535 & $\underline{0.777}$ & -- & -- \\
& \textsc{cs} & $\underline{0.691}$ & 0.632 & $\mathbf{0.829}$ & 0.626 & 0.671 & 0.665 & 0.448 & 0.651 & -- & -- \\
& \textsc{lg} & 0.877 & 0.858 & $\mathbf{0.949}$ & 0.868 & 0.885 & $\underline{0.896}$ & 0.686 & 0.890 & -- & -- \\
& \textsc{st} & 0.993 & 0.952 & $\mathbf{1.000}$ & $\mathbf{1.000}$ & 0.993 & 0.952 & 0.814 & 0.958 & -- & -- \\

\midrule

\multirow[c]{5}{*}{\textbf{Fannie Mae}}
& \textsc{sd} & $\underline{0.900}$ & 0.860 & 0.876 & 0.899 & $\mathbf{0.906}$ & 0.854 & 0.814 & -- & -- & -- \\
& \textsc{ts} & 0.567 & 0.550 & $\underline{0.571}$ & $\mathbf{0.598}$ & 0.503 & 0.474 & 0.481 & -- & -- & -- \\
& \textsc{cs} & $\mathbf{0.965}$ & 0.852 & 0.870 & 0.849 & $\underline{0.876}$ & 0.777 & 0.772 & -- & -- & -- \\
& \textsc{lg} & 0.820 & 0.787 & 0.781 & 0.822 & $\underline{0.858}$ & 0.804 & $\mathbf{0.864}$ & -- & -- & -- \\
& \textsc{st} & $\mathbf{0.998}$ & 0.953 & 0.822 & 0.750 & $\underline{0.990}$ & 0.817 & 0.848 & -- & -- & -- \\

\midrule

\multirow[c]{5}{*}{\textbf{Walmart}}
& \textsc{sd} & 0.796 & 0.861 & 0.815 & $\underline{0.898}$ & $\mathbf{0.950}$ & 0.719 & 0.857 & -- & -- & -- \\
& \textsc{ts} & $\mathbf{0.610}$ & 0.526 & $\underline{0.608}$ & 0.598 & 0.600 & 0.455 & 0.522 & -- & -- & -- \\
& \textsc{cs} & 0.867 & 0.785 & $\mathbf{0.896}$ & 0.828 & $\underline{0.891}$ & 0.605 & 0.746 & -- & -- & -- \\
& \textsc{lg} & $\mathbf{0.887}$ & 0.844 & 0.868 & 0.852 & $\underline{0.872}$ & 0.690 & 0.814 & -- & -- & -- \\
& \textsc{st} & 0.864 & 0.731 & $\mathbf{0.887}$ & 0.804 & $\underline{0.882}$ & 0.518 & 0.786 & -- & -- & -- \\

\midrule

\multirow[c]{5}{*}{\textbf{Airbnb}}
& \textsc{sd} & $\mathbf{0.967}$ & 0.909 & $\underline{0.963}$ & 0.937 & 0.925 & 0.879 & 0.711 & 0.923 & -- & -- \\
& \textsc{ts} & $\mathbf{0.875}$ & 0.840 & 0.824 & $\underline{0.860}$ & 0.843 & 0.777 & 0.811 & $\underline{0.860}$ & -- & -- \\
& \textsc{cs} & -- & -- & -- & -- & -- & -- & -- & -- & -- & -- \\
& \textsc{lg} & 0.964 & $\mathbf{0.970}$ & 0.965 & $\underline{0.969}$ & $\underline{0.969}$ & 0.967 & 0.909 & 0.965 & -- & -- \\
& \textsc{st} & 0.987 & 0.924 & $\mathbf{1.000}$ & $\mathbf{1.000}$ & 0.981 & 0.631 & 0.938 & 0.912 & -- & -- \\

\midrule

\multirow[c]{5}{*}{\textbf{PTB-XL}}
& \textsc{sd} & $\underline{0.886}$ & 0.576 & $\mathbf{0.896}$ & 0.669 & 0.692 & 0.625 & 0.758 & -- & -- & -- \\
& \textsc{ts} & 0.594 & 0.330 & 0.577 & 0.591 & 0.534 & $\mathbf{1.000}$ & $\underline{0.675}$ & -- & -- & -- \\
& \textsc{cs} & -- & -- & -- & -- & -- & -- & -- & -- & -- & -- \\
& \textsc{lg} & $\mathbf{0.927}$ & 0.841 & 0.909 & $\underline{0.922}$ & 0.906 & 0.915 & 0.920 & -- & -- & -- \\
& \textsc{st} & $\mathbf{1.000}$ & $\mathbf{1.000}$ & $\mathbf{1.000}$ & $\mathbf{1.000}$ & $\mathbf{1.000}$ & 0.999 & $\mathbf{1.000}$ & -- & -- & -- \\

\midrule

\multirow[c]{5}{*}{\textbf{Freddie Mac}}
& \textsc{sd} & 0.944 & 0.882 & $\mathbf{0.986}$ & 0.920 & $\underline{0.975}$ & 0.853 & 0.810 & -- & -- & -- \\
& \textsc{ts} & 0.713 & 0.688 & 0.660 & $\mathbf{0.752}$ & 0.736 & 0.581 & $\underline{0.747}$ & -- & -- & -- \\
& \textsc{cs} & 0.924 & 0.843 & $\mathbf{0.982}$ & 0.847 & $\underline{0.930}$ & 0.777 & 0.740 & -- & -- & -- \\
& \textsc{lg} & 0.842 & 0.797 & $\mathbf{0.915}$ & 0.841 & $\underline{0.875}$ & 0.784 & 0.780 & -- & -- & -- \\
& \textsc{st} & 0.995 & 0.972 & $\mathbf{1.000}$ & $\mathbf{1.000}$ & 0.993 & 0.695 & 0.965 & -- & -- & -- \\

\midrule

\multirow[c]{5}{*}{\textbf{Coupon}}
& \textsc{sd} & $\underline{0.895}$ & 0.894 & $\mathbf{0.976}$ & 0.881 & 0.863 & -- & 0.827 & -- & -- & -- \\
& \textsc{ts} & 0.828 & 0.817 & $\mathbf{0.896}$ & $\underline{0.848}$ & 0.606 & -- & 0.805 & -- & -- & -- \\
& \textsc{cs} & $\mathbf{0.801}$ & $\underline{0.798}$ & 0.789 & 0.779 & 0.737 & -- & 0.474 & -- & -- & -- \\
& \textsc{lg} & 0.636 & 0.626 & 0.668 & $\mathbf{0.908}$ & 0.674 & -- & $\underline{0.837}$ & -- & -- & -- \\
& \textsc{st} & 0.974 & 0.930 & $\mathbf{0.987}$ & $\underline{0.979}$ & 0.949 & -- & 0.895 & -- & -- & -- \\

\midrule

\multirow[c]{5}{*}{\textbf{Google Cluster}}
& \textsc{sd} & $\underline{0.863}$ & 0.715 & $\mathbf{0.935}$ & 0.781 & 0.706 & -- & 0.599 & -- & -- & -- \\
& \textsc{ts} & 0.604 & 0.574 & 0.627 & $\underline{0.645}$ & $\mathbf{0.650}$ & -- & 0.630 & -- & -- & -- \\
& \textsc{cs} & -- & -- & -- & -- & -- & -- & -- & -- & -- & -- \\
& \textsc{lg} & 0.481 & 0.457 & $\underline{0.507}$ & 0.455 & $\mathbf{0.516}$ & -- & 0.486 & -- & -- & -- \\
& \textsc{st} & 0.977 & 0.708 & $\mathbf{1.000}$ & $\mathbf{1.000}$ & 0.986 & -- & 0.567 & -- & -- & -- \\

\midrule

\multirow[c]{5}{*}{\textbf{H\&M}}
& \textsc{sd} & $\mathbf{0.933}$ & 0.626 & $\underline{0.891}$ & 0.871 & 0.871 & -- & 0.641 & -- & -- & -- \\
& \textsc{ts} & 0.680 & $\mathbf{0.709}$ & 0.682 & 0.688 & $\underline{0.704}$ & -- & 0.581 & -- & -- & -- \\
& \textsc{cs} & 0.820 & 0.521 & 0.789 & $\underline{0.842}$ & $\mathbf{0.874}$ & -- & 0.456 & -- & -- & -- \\
& \textsc{lg} & $\underline{0.873}$ & 0.715 & 0.756 & 0.772 & $\mathbf{0.876}$ & -- & 0.680 & -- & -- & -- \\
& \textsc{st} & $\underline{0.640}$ & 0.555 & 0.638 & $\underline{0.640}$ & $\underline{0.640}$ & -- & $\mathbf{0.648}$ & -- & -- & -- \\

\midrule

\multirow[c]{5}{*}{\textbf{Home Credit}}
& \textsc{sd} & 0.671 & 0.850 & $\mathbf{0.896}$ & 0.826 & $\underline{0.860}$ & -- & -- & -- & -- & -- \\
& \textsc{ts} & 0.628 & 0.611 & $\underline{0.690}$ & 0.671 & $\mathbf{0.719}$ & -- & -- & -- & -- & -- \\
& \textsc{cs} & -- & -- & -- & -- & -- & -- & -- & -- & -- & -- \\
& \textsc{lg} & 0.916 & 0.888 & $\underline{0.935}$ & 0.910 & $\mathbf{0.937}$ & -- & -- & -- & -- & -- \\
& \textsc{st} & 0.953 & 0.870 & $\underline{0.996}$ & $\mathbf{1.000}$ & 0.994 & -- & -- & -- & -- & -- \\

\midrule 

\multirow[c]{5}{*}{\textbf{Citi Bike}}
& \textsc{sd} & $\mathbf{0.990}$ & $\underline{0.943}$ & -- & -- & -- & 0.870 & 0.883 & -- & 0.706 & 0.890 \\
& \textsc{ts} & $\underline{0.769}$ & 0.765 & -- & -- & -- & 0.678 & $\mathbf{0.782}$ & -- & 0.721 & 0.201 \\
& \textsc{cs} & $\mathbf{0.941}$ & $\underline{0.900}$ & -- & -- & -- & 0.885 & 0.848 & -- & 0.614 & 0.757 \\
& \textsc{lg} & $\underline{0.965}$ & 0.960 & -- & -- & -- & $\mathbf{0.967}$ & 0.950 & -- & 0.811 & 0.921 \\
& \textsc{st} & 0.908 & $\mathbf{0.979}$ & -- & -- & -- & 0.676 & $\mathbf{0.979}$ & -- & 0.271 & 0.154 \\

\midrule

\multirow[c]{5}{*}{\textbf{CMAPSS}}
& \textsc{sd} & 0.963 & 0.813 & -- & -- & -- & $\underline{0.972}$ & 0.916 & -- & 0.811 & $\mathbf{0.983}$ \\
& \textsc{ts} & 0.575 & 0.369 & -- & -- & -- & 0.322 & 0.545 & -- & $\mathbf{0.969}$ & $\underline{0.578}$ \\
& \textsc{cs} & -- & -- & -- & -- & -- & -- & -- & -- & -- & -- \\
& \textsc{lg} & 0.471 & 0.459 & -- & -- & -- & $\mathbf{0.618}$ & 0.425 & -- & $\underline{0.573}$ & 0.485 \\
& \textsc{st} & 0.709 & 0.330 & -- & -- & -- & 0.000 & $\mathbf{0.961}$ & -- & 0.836 & $\underline{0.917}$ \\

\bottomrule
\end{tabular*}

\vspace{-1.5mm}
\end{table*}

\subsection{MLE-TS}
\label{app:mle-ts-online}

\begin{table*}[t]
\centering
\caption{AUC for MLE-TS. Original denotes the real-data baseline trained and evaluated on real data. Higher values indicate better downstream temporal utility, and the best synthetic result for each dataset is highlighted in bold.}
\Description{A downstream utility table reporting AUC scores for the MLE-TS task across all datasets. Rows correspond to datasets, and columns compare the real-data baseline with synthetic data generation models. Higher AUC values indicate better downstream temporal utility, and the best synthetic result for each dataset is highlighted. The table shows that temporal utility varies substantially by dataset and model, with some synthetic models approaching or matching the real-data baseline on certain datasets while performing much worse on others.}
\label{tab:mle_ts_whole}
\small
\setlength{\tabcolsep}{3pt}
\renewcommand{\arraystretch}{0.9}
\resizebox{\linewidth}{!}{%
\begin{tabular}{@{}lccccccccccc@{}}
\toprule
\textbf{Dataset} & \textbf{Original} & \textbf{ClavaDDPM} & \textbf{CPAR} & \textbf{RCTGAN} & \textbf{RTF} & \textbf{SDV} & \textbf{RDBDIFF} & \textbf{RGCLD} & \textbf{RelDiff} & \textbf{TABARGN} & \textbf{TabDiT} \\
\midrule
\textbf{Rossmann}        & $0.806$ & $0.771$ & --        & $0.314$ & $0.500$ & $0.533$ & $0.749$ & $0.778$ & $0.708$ & -- & $\mathbf{0.782}$ \\
\textbf{Berka}           & $0.502$ & $0.445$ & --        & $0.480$ & $0.511$ & $0.467$ & $0.492$ & $0.550$ & $\mathbf{0.557}$ & --        & $0.438$ \\
\textbf{Fannie Mae}      & $0.685$ & $0.601$ & --        & $0.533$ & $0.628$ & $0.602$ & $0.622$ & $0.654$ & $\mathbf{0.658}$ & -- & -- \\
\textbf{Walmart}         & $0.717$ & $\mathbf{0.867}$ & -- & $0.550$ & $0.767$ & $0.733$ & $0.833$ & $0.792$ & $0.500$ & -- & -- \\
\textbf{Airbnb}          & $1.000$ & $0.806$ & --        & $0.881$ & $0.624$ & $0.978$ & $\mathbf{1.000}$ & $1.000$ & $0.930$ & --        & $\mathbf{1.000}$ \\
\textbf{PTB-XL}          & $0.712$ & $0.473$ & --        & $0.476$ & $\mathbf{0.672}$ & $0.500$ & $0.515$ & $0.524$ & $0.446$ & -- & -- \\
\textbf{Freddie Mac}     & $0.691$ & $0.486$ & --        & $0.519$ & $0.572$ & $0.500$ & $0.504$ & $\mathbf{0.664}$ & $0.596$ & -- & -- \\
\textbf{Citi Bike}       & $0.994$ & $0.995$ & $\mathbf{0.996}$ & $0.839$ & $0.995$ & $0.836$ & -- & -- & -- & $0.669$ & -- \\
\textbf{CMAPSS}          & $0.848$ & $0.757$ & $0.500$ & $0.475$ & $0.757$ & $\mathbf{0.820}$ & -- & -- & -- & $0.771$ & -- \\
\textbf{Coupon}          & $0.679$ & $0.577$ & --        & $0.507$ & --      & $0.527$ & $0.542$ & $\mathbf{0.645}$ & $0.610$ & -- & -- \\
\textbf{Google Cluster}  & $0.950$ & $0.828$ & --        & $0.616$ & --      & $0.437$ & $\mathbf{0.885}$ & $0.790$ & $0.725$ & -- & -- \\
\textbf{H\&M}            & $0.602$ & $\mathbf{0.585}$ & -- & $0.479$ & --      & $0.536$ & $0.579$ & $0.577$ & $0.574$ & -- & -- \\
\textbf{Home Credit }    & $0.590$ & $0.467$ & --        & $0.460$ & --      & --      & $0.517$ & $\mathbf{0.573}$ & $0.530$ & -- & -- \\
\bottomrule
\end{tabular}%
}
\end{table*}

Table~\ref{tab:mle_ts_whole} summarizes downstream temporal utility on the
MLE-TS benchmark. The best synthetic generators often approach, and sometimes
surpass, the real-data baseline, showing that synthetic data can support
downstream temporal prediction when it preserves task-relevant patterns rather
than only marginal distributions. At the same time, utility remains highly
dataset-dependent: no method consistently dominates, and larger gaps on
datasets with complex temporal dynamics and heterogeneous feature interactions
suggest that current generators still struggle to reproduce fine-grained
predictive structure. This highlights the need to evaluate synthetic data not
only by distributional fidelity, but also by whether it transfers useful
temporal signals to downstream tasks.

\subsection{MLE-Temporal}
\label{app:mle-temporal-online}

\begin{table}[htbp]
\centering
\caption{MLE-Temporal utility results across datasets. Bold and underlined scores indicate the best and second-best results within each dataset, respectively.}
\Description{MLE-Temporal utility results across four datasets. AUC is reported for Fannie Mae and Freddie Mac, while $R^2$ is reported for Rossmann and Walmart. Bold and underlined scores indicate the best and second-best results within each dataset, respectively.}
\label{tab:mle_temporal_results}
\normalsize
\setlength{\tabcolsep}{5pt}
\renewcommand{\arraystretch}{0.95}
\begin{tabular}{lllc}
\toprule
Dataset & Metric & Model & Performance \\
\midrule
\multirow{8}{*}{\textbf{Rossmann}}
& $R^2$ & \textbf{ClavaDDPM}     & \underline{0.577} \\
& $R^2$ & \textbf{RCTGAN}        & 0.422 \\
& $R^2$ & \textbf{RDBDIFF}       & 0.530 \\
& $R^2$ & \textbf{RTF}           & 0.285 \\
& $R^2$ & \textbf{RelDiff}       & 0.187 \\
& $R^2$ & \textbf{RGCLD}         & 0.571 \\
& $R^2$ & \textbf{SDV}           & $-64.945$ \\
& $R^2$ & \textbf{TabDiT}        & \textbf{0.578} \\
\midrule
\multirow{7}{*}{\textbf{Fannie Mae}}
& AUC & \textbf{ClavaDDPM}     & 0.502 \\
& AUC & \textbf{RCTGAN}        & 0.702 \\
& AUC & \textbf{RDBDIFF}       & 0.781 \\
& AUC & \textbf{RTF}           & 0.852 \\
& AUC & \textbf{RelDiff}       & \textbf{0.924} \\
& AUC & \textbf{RGCLD}         & 0.890 \\
& AUC & \textbf{SDV}           & \underline{0.913} \\
\midrule
\multirow{7}{*}{\textbf{Walmart}}
& $R^2$ & \textbf{ClavaDDPM}     & 0.355 \\
& $R^2$ & \textbf{RCTGAN}        & 0.247 \\
& $R^2$ & \textbf{RDBDIFF}       & \underline{0.475} \\
& $R^2$ & \textbf{RTF}           & 0.104 \\
& $R^2$ & \textbf{RelDiff}       & $-0.014$ \\
& $R^2$ & \textbf{RGCLD}         & \textbf{0.516} \\
& $R^2$ & \textbf{SDV}           & 0.012 \\
\midrule
\multirow{7}{*}{\textbf{Freddie Mac}}
& AUC & \textbf{ClavaDDPM}     & 0.596 \\
& AUC & \textbf{RCTGAN}        & 0.643 \\
& AUC & \textbf{RDBDIFF}       & 0.686 \\
& AUC & \textbf{RTF}           & 0.527 \\
& AUC & \textbf{RelDiff}       & \underline{0.848} \\
& AUC & \textbf{RGCLD}         & \textbf{0.869} \\
& AUC & \textbf{SDV}           & 0.500 \\
\bottomrule
\end{tabular}
\end{table}

Table~\ref{tab:mle_temporal_results} reports MLE-Temporal results under a
chronological train/test split, where models are trained on earlier observations
and evaluated on later ones. This setting provides a stricter test than the
random-split MLE-TS protocol because synthetic data must support forward-looking
temporal prediction. On the binary classification tasks, RelDiff and RGCLD obtain
the best AUC scores on Fannie Mae and Freddie Mac, respectively, with SDV also
performing strongly on Fannie Mae. In contrast, the regression tasks show weaker
utility: the best $R^2$ values on Rossmann and Walmart remain moderate, and some
generators produce near-zero or negative scores. This contrast suggests that
current generators can preserve coarse temporal signals useful for classification
more reliably than the fine-grained continuous dynamics required for temporal
forecasting.

\subsection{MLD-TS}
\label{app:mld-ts-online}

\begin{table*}[th]
\centering
\caption{Test accuracy for MLD-TS. child denotes the setting where only the child table is used, while merged denotes the setting where the child table is merged with the parent table. Bold and underline indicate the lowest and second-lowest detection results among synthetic generators, respectively.}
\Description{A detection-accuracy table for the MLD-TS task across all benchmark datasets. Each dataset has two settings: child, where only the child table is used, and merged, where the child table is merged with the parent table. Columns compare the original real-data baseline with synthetic data generation models. Lower detection accuracy for synthetic generators indicates that synthetic data are harder to distinguish from real data and therefore more realistic under this detection task. The table shows that most synthetic generators are still detected with high accuracy, while the lowest and second-lowest synthetic detection results vary across datasets and settings.}
\label{tab:mld_ts_whole}
\normalsize
\setlength{\tabcolsep}{2pt}
\renewcommand{\arraystretch}{1.08}
\resizebox{\linewidth}{!}{%
\begin{tabular}{@{}llccccccccccc@{}}
\toprule
\textbf{Dataset} & \textbf{Setting} & \textbf{Original} & \textbf{ClavaDDPM} & \textbf{CPAR} & \textbf{RCTGAN} & \textbf{RTF} & \textbf{SDV} & \textbf{RDBDIFF} & \textbf{RGCLD} & \textbf{RelDiff} & \textbf{TABARGN} & \textbf{TabDiT} \\
\midrule
\multirow{2}{*}{\textbf{Airbnb}}
& child  & $0.503$ & $0.964$ & -- & $0.987$ & $0.897$ & $1.000$ & $0.988$ & \underline{$0.879$} & $0.889$ & -- & $\mathbf{0.852}$ \\
& merged & $0.499$ & $0.965$ & -- & $0.987$ & $0.929$ & $1.000$ & $0.987$ & $\mathbf{0.879}$ & \underline{$0.912$} & -- & $0.947$ \\
\midrule
\multirow{2}{*}{\textbf{Berka}}
& child  & $0.498$ & $0.964$ & -- & $0.990$ & \underline{$0.672$} & $0.999$ & $\mathbf{0.532}$ & $0.738$ & $0.988$ & -- & $0.862$ \\
& merged & $0.493$ & $0.965$ & -- & $0.990$ & \underline{$0.669$} & $0.999$ & $\mathbf{0.523}$ & $0.741$ & $0.988$ & -- & $0.861$ \\
\midrule
\multirow{2}{*}{\textbf{Citi Bike}}
& child  & $0.499$ & $\mathbf{0.804}$ & $1.000$ & $0.970$ & \underline{$0.956$} & $0.997$ & -- & -- & -- & $0.992$ & -- \\
& merged & $0.499$ & $\mathbf{0.804}$ & $1.000$ & $0.970$ & \underline{$0.956$} & $0.997$ & -- & -- & -- & $0.992$ & -- \\
\midrule
\multirow{2}{*}{\textbf{CMAPSS}}
& child  & $0.502$ & $0.997$ & $1.000$ & $0.997$ & $\mathbf{0.976}$ & $0.998$ & -- & -- & -- & \underline{$0.981$} & -- \\
& merged & $0.502$ & $0.997$ & $1.000$ & $0.997$ & $\mathbf{0.976}$ & $0.998$ & -- & -- & -- & \underline{$0.981$} & -- \\
\midrule
\multirow{2}{*}{\textbf{Coupon}}
& child  & $0.499$ & $0.842$ & -- & $0.878$ & -- & $0.965$ & $\mathbf{0.758}$ & \underline{$0.812$} & $0.881$ & -- & -- \\
& merged & $0.498$ & $0.841$ & -- & $0.879$ & -- & $0.965$ & $\mathbf{0.759}$ & \underline{$0.808$} & $0.881$ & -- & -- \\
\midrule
\multirow{2}{*}{\textbf{Fannie Mae}}
& child  & $0.499$ & $1.000$ & -- & $1.000$ & $\mathbf{0.838}$ & \underline{$1.000$} & $1.000$ & $1.000$ & $1.000$ & -- & -- \\
& merged & $0.500$ & $1.000$ & -- & $1.000$ & $\mathbf{0.840}$ & $1.000$ & $1.000$ & \underline{$1.000$} & $1.000$ & -- & -- \\
\midrule
\multirow{2}{*}{\textbf{Freddie Mac}}
& child  & $0.502$ & $1.000$ & -- & $1.000$ & $0.999$ & $1.000$ & $1.000$ & $\mathbf{0.966}$ & \underline{$0.996$} & -- & -- \\
& merged & $0.502$ & $1.000$ & -- & $1.000$ & $0.999$ & $1.000$ & $1.000$ & $\mathbf{0.966}$ & \underline{$0.996$} & -- & -- \\
\midrule
\multirow{2}{*}{\textbf{Google Cluster}}
& child  & $0.503$ & $0.964$ & -- & $0.996$ & -- & $0.999$ & \underline{$0.915$} & $\mathbf{0.798}$ & $0.976$ & -- & -- \\
& merged & $0.500$ & $0.965$ & -- & $0.996$ & -- & $0.999$ & \underline{$0.915$} & $\mathbf{0.802}$ & $0.976$ & -- & -- \\
\midrule
\multirow{2}{*}{\textbf{H\&M}}
& child  & $0.495$ & \underline{$0.734$} & -- & $0.986$ & -- & $1.000$ & $0.903$ & $\mathbf{0.712}$ & $0.736$ & -- & -- \\
& merged & $0.496$ & $\mathbf{0.735}$ & -- & $0.986$ & -- & $1.000$ & $0.903$ & \underline{$0.902$} & $0.907$ & -- & -- \\
\midrule
\multirow{2}{*}{\textbf{Home Credit}}
& child  & $0.496$ & $0.928$ & -- & $0.919$ & -- & -- & $0.999$ & $\mathbf{0.691}$ & \underline{$0.812$} & -- & -- \\
& merged & $0.497$ & $1.000$ & -- & $0.981$ & -- & -- & $0.999$ & $\mathbf{0.732}$ & \underline{$0.849$} & -- & -- \\
\midrule
\multirow{2}{*}{\textbf{PTB-XL}}
& child  & $0.501$ & $1.000$ & -- & $1.000$ & $1.000$ & $1.000$ & $\mathbf{0.858}$ & \underline{$0.988$} & $0.994$ & -- & -- \\
& merged & $0.491$ & $1.000$ & -- & $1.000$ & $1.000$ & $1.000$ & $\mathbf{0.861}$ & \underline{$0.994$} & $0.994$ & -- & -- \\
\midrule
\multirow{2}{*}{\textbf{Rossmann}}
& child  & $0.502$ & $0.997$ & -- & $0.985$ & \underline{$0.984$} & $0.997$ & $0.986$ & \underline{$0.984$} & $0.998$ & -- & $\mathbf{0.825}$ \\
& merged & $0.504$ & $0.997$ & -- & $0.990$ & \underline{$0.984$} & $0.997$ & $0.985$ & $0.986$ & $0.998$ & -- & $\mathbf{0.817}$ \\
\midrule
\multirow{2}{*}{\textbf{Walmart}}
& child  & $0.548$ & $0.989$ & -- & $0.989$ & $0.982$ & $0.989$ & \underline{$0.978$} & $\mathbf{0.970}$ & $0.989$ & -- & -- \\
& merged & $0.533$ & $0.989$ & -- & $0.989$ & $0.982$ & $0.989$ & \underline{$0.978$} & $\mathbf{0.970}$ & $0.989$ & -- & -- \\
\bottomrule
\end{tabular}%
}
\end{table*}

Table~\ref{tab:mld_ts_whole} reports the detection accuracy of the MLD-TS classifier across all datasets, for both the child setting (child table only) and the merged setting (child joined with parent attributes). Lower accuracy indicates that real and synthetic records are harder to distinguish, and therefore that the synthetic data better reproduce the joint structure of the real records. The real-data baseline lies near $0.5$ across all datasets, confirming that the classifier is not exploiting trivial artifacts. In contrast, most synthetic generators are detected with accuracy well above $0.95$, indicating that current methods still leave clear distributional fingerprints under a learned discriminator. A few exceptions show measurable progress: RDBDIFF on Berka achieves the lowest detection accuracy ($0.523$ in the merged setting), and TabDiT on Rossmann ($0.817$) and Airbnb ($0.852$ in the child setting) likewise approach the lower end of the detectability range. RGCLD performs best on several relational datasets (H\&M, Home Credit, Walmart), consistent with its competitive temporal fidelity scores reported earlier. Differences between the child and merged settings are generally small, indicating that the detector primarily relies on child-table content rather than on parent--child interactions, and that synthetic-vs-real distinguishability is concentrated in the temporal records themselves. Overall, while MLD-TS confirms that recent relational and temporal generators have narrowed the realism gap on certain datasets, the gap to the real-data baseline remains large, suggesting that distinguishability under learned discriminators is still an open challenge.

\subsection{Privacy: Per-Dataset Results}
\label{app:privacy-online}

\begin{table}[ht]
\centering
\caption{Privacy metric profiles on the six additional core
datasets. Convention: higher values indicate safer behavior.
\texttt{DCR} and \texttt{CS-DCR} are raw median nearest-neighbor
distances (not normalized);
$\texttt{NGP}(n)=1-\texttt{NgramExposure}(n)\in[0,1]$.
Bold and underline mark the best and second-best per column
\emph{within each dataset block}.}
\Description{A single combined results table reporting privacy
metric profiles across six datasets (Berka, Airbnb, PTB-XL,
Fannie Mae, Freddie Mac, Walmart). Each dataset block lists
synthetic data generation models as rows. Columns report
row-level privacy (DCR, NNDR), cross-sectional privacy (CS-DCR,
CS-NNDR), and n-gram privacy (NGP(1), NGP(3)). Bold and underline
mark the best and second-best per column within each dataset
block. The table shows that privacy behavior differs across
models and datasets, with models varying in their row-level,
time-slice-level, and temporal-pattern privacy risks.}
\label{tab:privacy-online}
\scriptsize
\setlength{\tabcolsep}{3pt}
\renewcommand{\arraystretch}{0.95}
\begin{adjustbox}{width=\columnwidth,center}
\begin{tabular}{l l c c c c c c}
\toprule
\textbf{Dataset} & \textbf{Model}
 & \texttt{DCR} & \texttt{CS-DCR}
 & \texttt{NNDR} & \texttt{CS-NNDR}
 & \texttt{NGP}(1) & \texttt{NGP}(3) \\
\midrule
\multirow{8}{*}{\textbf{Berka}}
 & ClavaDDPM & $47.06$ & $5761$ & $0.685$ & $0.600$ & $0.001$ & $0.888$ \\
 & RCTGAN    & $51.50$ & $\underline{9919}$ & $0.718$ & $0.656$ & $0.089$ & $0.954$ \\
 & RDBDiff   & $43.92$ & $2919$ & $0.677$ & $0.461$ & $0.001$ & $0.531$ \\
 & RTF       & $49.38$ & $9106$ & $0.690$ & $\underline{0.679}$ & $0.006$ & $0.285$ \\
 & RGCLD     & $48.36$ & $7820$ & $0.686$ & $0.669$ & $0.664$ & $0.991$ \\
 & RelDiff   & $\underline{54.50}$ & $8669$ & $0.692$ & $0.653$ & $\underline{0.688}$ & $\underline{0.999}$ \\
 & SDV       & $\mathbf{32768}$ & $\mathbf{43831}$ & $\mathbf{1.000}$ & $\mathbf{0.905}$ & $\mathbf{0.996}$ & $\mathbf{1.000}$ \\
 & TabDiT    & $49.40$ & $8094$ & $\underline{0.729}$ & $0.673$ & $0.644$ & $0.988$ \\
\midrule
\multirow{8}{*}{\textbf{Airbnb}}
 & ClavaDDPM & $0.000$ & $10.47$ & $0.500$ & $0.495$ & $0.078$ & $0.959$ \\
 & RCTGAN    & $0.000$ & $15.97$ & $0.500$ & $\mathbf{0.512}$ & $0.202$ & $\underline{0.994}$ \\
 & RDBDiff   & $0.000$ & $11.46$ & $0.500$ & $0.494$ & $0.232$ & $0.773$ \\
 & RTF       & $0.000$ & $12.02$ & $0.500$ & $0.509$ & $0.078$ & $0.801$ \\
 & RGCLD     & $0.000$ & $10.78$ & $0.500$ & $0.495$ & $0.125$ & $0.767$ \\
 & RelDiff   & $0.000$ & $17.10$ & $0.500$ & $\underline{0.510}$ & $\underline{0.285}$ & $0.939$ \\
 & SDV       & $\mathbf{1.000}$ & $\mathbf{501.6}$ & $\underline{0.545}$ & $0.488$ & $\mathbf{1.000}$ & $\mathbf{1.000}$ \\
 & TabDiT    & $0.000$ & $\underline{43.07}$ & $\mathbf{0.635}$ & $0.503$ & $0.133$ & $0.714$ \\
\midrule
\multirow{7}{*}{\textbf{PTB-XL}}
 & ClavaDDPM & $0.059$ & $0.110$ & $0.904$ & $0.901$ & $0.026$ & $0.778$ \\
 & RCTGAN    & $\mathbf{0.317}$ & $\mathbf{0.722}$ & $\underline{0.927}$ & $0.839$ & $\mathbf{0.986}$ & $\mathbf{1.000}$ \\
 & RDBDiff   & $0.051$ & $0.101$ & $0.878$ & $0.883$ & $0.018$ & $0.229$ \\
 & RTF       & $0.064$ & $0.115$ & $0.921$ & $\underline{0.915}$ & $0.070$ & $0.327$ \\
 & RGCLD     & $0.064$ & $0.117$ & $0.920$ & $0.913$ & $0.036$ & $0.238$ \\
 & RelDiff   & $\underline{0.125}$ & $\underline{0.195}$ & $\mathbf{0.944}$ & $\mathbf{0.935}$ & $\underline{0.354}$ & $\underline{0.885}$ \\
 & SDV       & $0.095$ & $0.169$ & $\underline{0.927}$ & $0.909$ & $0.144$ & $0.760$ \\
\midrule
\multirow{7}{*}{\textbf{Fannie Mae}}
 & ClavaDDPM & $45.93$ & $101.7$ & $\underline{0.808}$ & $0.775$ & $0.288$ & $\mathbf{1.000}$ \\
 & RCTGAN    & $68.37$ & $353.7$ & $0.778$ & $0.717$ & $\mathbf{0.979}$ & $\mathbf{1.000}$ \\
 & RDBDiff   & $152.5$ & $782.0$ & $0.781$ & $0.734$ & $0.704$ & $\mathbf{1.000}$ \\
 & RTF       & $26.51$ & $68.86$ & $\mathbf{0.857}$ & $\underline{0.789}$ & $0.421$ & $\mathbf{1.000}$ \\
 & RGCLD     & $\underline{179.7}$ & $\underline{802.4}$ & $0.765$ & $0.725$ & $0.427$ & $0.966$ \\
 & RelDiff   & $82.75$ & $300.9$ & $0.793$ & $0.733$ & $0.813$ & $\mathbf{1.000}$ \\
 & SDV       & $\mathbf{456.1}$ & $\mathbf{2099}$ & $\underline{0.808}$ & $\mathbf{0.796}$ & $\underline{0.969}$ & $\mathbf{1.000}$ \\
\midrule
\multirow{7}{*}{\textbf{Freddie Mac}}
 & ClavaDDPM & $14.61$ & $45.04$ & $\underline{0.797}$ & $0.697$ & $0.005$ & $\mathbf{0.999}$ \\
 & RCTGAN    & $\mathbf{198.6}$ & $\mathbf{1364}$ & $0.738$ & $\underline{0.725}$ & $0.302$ & $\underline{0.997}$ \\
 & RDBDiff   & $16.61$ & $55.16$ & $0.790$ & $0.704$ & $0.004$ & $0.573$ \\
 & RTF       & $31.08$ & $71.97$ & $0.780$ & $0.715$ & $\underline{0.517}$ & $\underline{0.997}$ \\
 & RGCLD     & $42.06$ & $185.2$ & $0.760$ & $0.723$ & $0.016$ & $0.640$ \\
 & RelDiff   & $135.8$ & $\underline{641.2}$ & $0.741$ & $0.719$ & $0.161$ & $0.892$ \\
 & SDV       & $\underline{138.7}$ & $421.5$ & $\mathbf{0.814}$ & $\mathbf{0.788}$ & $\mathbf{0.616}$ & $0.928$ \\
\midrule
\multirow{7}{*}{\textbf{Walmart}}
 & ClavaDDPM & $42.50$ & $38.41$ & $0.535$ & $0.546$ & $\mathbf{1.000}$ & $\mathbf{1.000}$ \\
 & RCTGAN    & $822.5$ & $\underline{1495}$ & $\underline{0.669}$ & $\underline{0.639}$ & $0.512$ & $\mathbf{1.000}$ \\
 & RDBDiff   & $99.43$ & $158.8$ & $0.551$ & $0.529$ & $\mathbf{1.000}$ & $\mathbf{1.000}$ \\
 & RTF       & $779.4$ & $718.7$ & $0.651$ & $\mathbf{0.673}$ & $0.510$ & $0.964$ \\
 & RGCLD     & $541.1$ & $703.1$ & $0.645$ & $0.636$ & $0.461$ & $\mathbf{1.000}$ \\
 & RelDiff   & $\underline{860.9}$ & $1415$ & $\mathbf{0.674}$ & $\underline{0.639}$ & $0.545$ & $\mathbf{1.000}$ \\
 & SDV       & $\mathbf{1420}$ & $\mathbf{2990}$ & $0.559$ & $0.603$ & $0.516$ & $\mathbf{1.000}$ \\
\bottomrule
\end{tabular}
\end{adjustbox}
\end{table}

Table~\ref{tab:privacy-online} reports the privacy metrics on the remaining
six core datasets, complementing the Rossmann analysis of
Section~\ref{sec:privacy}. All six metrics follow the
\emph{higher-is-safer} convention; \texttt{DCR} and
\texttt{CS-DCR} are raw median nearest-neighbor distances and are
not normalized. For the $n$-gram metric we report
$\texttt{NGP}(n)=1-\texttt{NgramExposure}(n)$, so values close to
$1$ indicate that the synthetic data reproduce few real
length-$n$ subsequences.

\paragraph{Cross-sectional restriction.}
Across the six datasets, $\texttt{CS-DCR}$ exceeds $\texttt{DCR}$
on roughly $90\%$ of model--dataset pairs, confirming that the
row-level closeness signal is dominated by neighbors at unrelated
time slices. The ratio varies sharply with dataset
characteristics: on Berka (long, irregular financial
trajectories) the median inflation is $\sim\!160\times$, on
Fannie Mae and Freddie Mac (loan histories)
$\sim\!3$--$5\times$, and on PTB-XL (short fixed-length ECG
sequences) only $\sim\!1.8\times$. Walmart is the partial
exception: ClavaDDPM has
$\texttt{CS-DCR}\!=\!38.4 < \texttt{DCR}\!=\!42.5$ and RTF has
$\texttt{CS-DCR}\!=\!718.7 < \texttt{DCR}\!=\!779.4$, meaning
that same-week nearest neighbors are closer on average than
time-agnostic ones---an instance where the row-level metric
understates rather than overstates time-aligned identifiability.

\paragraph{$n$-gram Privacy.}
The $n{=}1$ vs.\ $n{=}3$ disagreement seen on Rossmann is
dataset-dependent. On datasets with high row vocabulary
(Berka, Fannie Mae, Freddie Mac, PTB-XL), \texttt{NGP}(1)
already shows a wide spread among models, so \texttt{NGP}(3)
refines rather than overturns the single-token privacy
conclusion. On datasets with bounded row vocabulary (Airbnb,
Walmart), \texttt{NGP}(1) saturates near the extremes and
\texttt{NGP}(3) is the informative quantity. Across all seven
datasets, RTF retains the lowest \texttt{NGP}(3) on
five datasets and is in the bottom two on the remaining two,
confirming that autoregressive sequence-level reproduction is an
architecture-level rather than a dataset-specific effect.

\subsection{Trajectory fidelity.}
\label{app:tt_wasserstein}



\begin{table}[ht]
\centering
\caption{TT-Wasserstein distance after applying \textsc{Interpolation-fill}. Lower values indicate better preservation of real-data trajectory patterns. The best synthetic result in each dataset is highlighted in bold.}
\Description{TT-Wasserstein distance results for Rossmann, Walmart, and PTB-XL after applying Interpolation-fill. Rows correspond to the Original baseline and synthetic data generators, and columns correspond to datasets. Lower values indicate better preservation of real-data trajectory patterns. The best synthetic result in each dataset is highlighted in bold.}
\label{tab:tt_wasserstein}
\normalsize
\setlength{\tabcolsep}{5pt}
\renewcommand{\arraystretch}{0.95}
\begin{tabular*}{\columnwidth}{@{\extracolsep{\fill}}lccc@{}}
\toprule
\textbf{Method} 
& \textbf{Rossmann} 
& \textbf{Walmart} 
& \textbf{PTB-XL} \\
\midrule
\textbf{Original}   & 0.032 & 0.136 & 0.014 \\
\textbf{ClavaDDPM}  & 0.111 & 0.203 & 0.016 \\
\textbf{RCTGAN}     & 0.136 & 0.233 & 0.017 \\
\textbf{RTF}        & 0.050 & 0.240 & 0.018 \\
\textbf{SDV}        & 0.136 & \textbf{0.182} & 0.016 \\
\textbf{RDBDiff}    & 0.136 & 0.197 & \textbf{0.015} \\
\textbf{RGCLD}      & 0.096 & 0.200 & 0.016 \\
\textbf{RelDiff}    & 0.138 & 0.264 & 0.016 \\
\textbf{TabDiT}     & \textbf{0.045} & --    & -- \\
\bottomrule
\end{tabular*}
\vspace{-0.5em}
\end{table}

\texttt{TT-WassersteinDistance} follows Li et al.~\cite{ttwasserstein}: the spectral trajectory
embedding and the optimal-transport cost are used exactly as defined there. The score is reported
below as a distance rather than as a $[0,1]$ similarity; only when it enters the aggregated
longitudinal dimension is it reoriented as $1-\mathrm{TT\text{-}W}$, which is why it is marked as
\emph{adapted} in Table~\ref{tab:metric_applicability}.
Table~\ref{tab:tt_wasserstein} reports TT-Wasserstein distances across
datasets and generators. The Original column measures the distance between two real-data splits and serves as a baseline for natural trajectory
variation. Synthetic datasets consistently exceed this baseline, indicating
additional trajectory-level discrepancies beyond inherent entity heterogeneity.
The best model varies by dataset, with TabDiT achieving the lowest distance on
Rossmann, suggesting that explicit temporal modeling can help preserve
trajectory-level patterns.


\begin{figure}[h]
    \centering
    \includegraphics[width=0.45\textwidth]{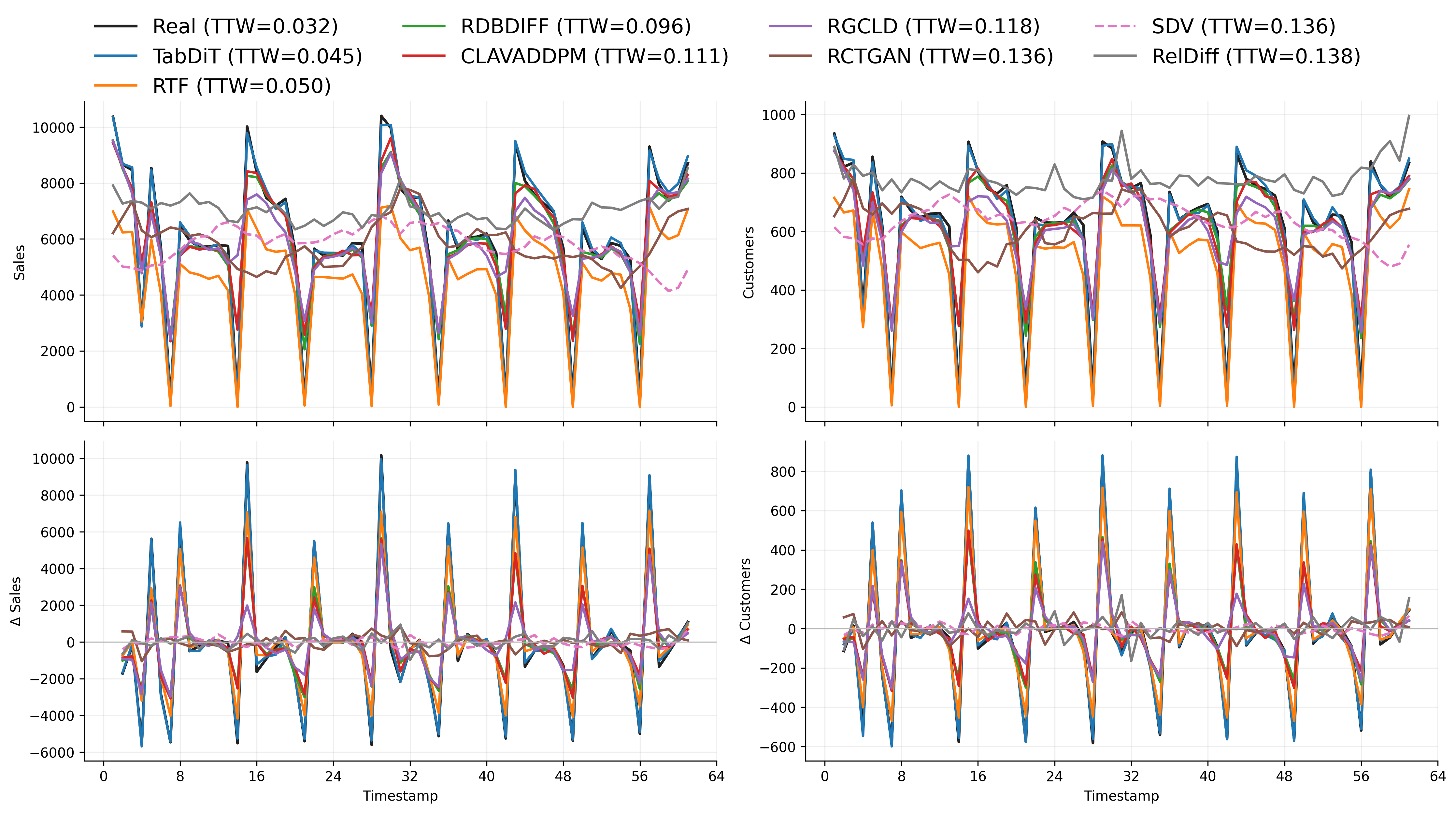}
    \caption{TT-Wasserstein evaluation on the Rossmann dataset. The upper panel shows raw trajectories, while the lower panel shows first-differenced trajectories.}
    \Description{A two-row panel figure comparing real and synthetic trajectories on the Rossmann dataset. The upper row plots raw value trajectories over time for the real data and for each synthetic data generation model, and the lower row plots the corresponding first-differenced trajectories. Models with lower TT-Wasserstein distance produce curves that track the real trajectory more closely in both rows, while models with higher distance produce trajectories that are either overly smoothed or shifted relative to the real data.}
    \label{fig:rossmann_tabdit_ttw}
\end{figure}

Figure~\ref{fig:rossmann_tabdit_ttw} shows that TT-Wasserstein Distance is well aligned with visually observed trajectory fidelity. Models with lower TTW, such as TabDiT and RTF, generate trajectories that more closely follow the real baseline in both raw values and first-order differences. In particular, TabDiT preserves not only the overall level and periodic trajectory shape, but also the timing and magnitude of abrupt changes, which explains its lowest distance. By contrast, models with intermediate distances, such as RDBDIFF and ClavaDDPM, partially recover the timing of recurring fluctuations but underestimate or distort their transition magnitudes. Models with larger distances produce trajectories that are either overly smoothed or structurally misaligned, indicating failure to preserve both global trajectory morphology and local temporal dynamics.

Full per-metric results across all datasets are provided in
Tables~\ref{tab:submetric-ts}--\ref{tab:submetric-struct}, while the
SD metric results are reported in Table~\ref{tab:submetric-sd}.

\subsection{Qualitative Results}
\label{app:qualitative-online}
In this section, we provide qualitative analyses to complement the quantitative evaluation. Although the proposed metrics summarize different aspects of fidelity, aggregate scores alone do not always reveal how these results manifest in individual examples. We therefore present representative case studies to illustrate the qualitative behavior of synthetic data, including temporal patterns, visually apparent artifacts, and case-specific failure modes. These analyses help clarify where synthetic data successfully preserves the characteristics of real data and where it exhibits unrealistic or inconsistent behaviors that are difficult to interpret from numerical scores alone.

\subsubsection{Synthetic data generation failure mode}

\begin{figure}[htbp]
    \centering
    \includegraphics[width=\columnwidth]{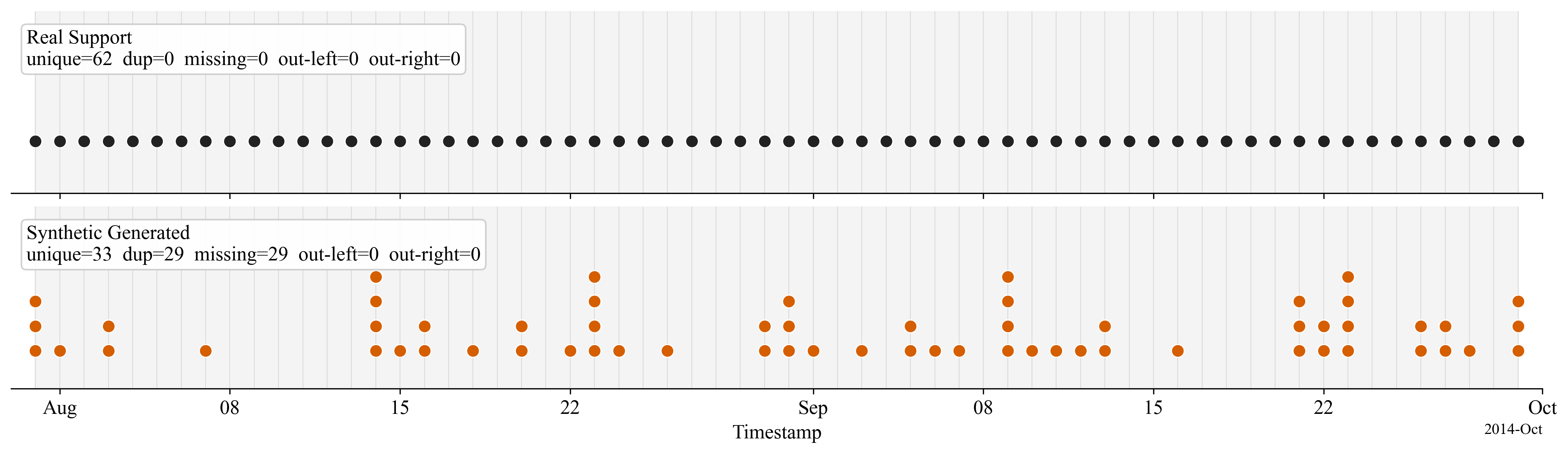}
    \caption{Case study of duplicate timestamps and missing grid points. Each point corresponds to one record, so the number of points at each timestamp represents the record count at that timestamp.}
    \Description{A scatter plot comparing one real trajectory with one synthetic trajectory along the time axis. Each point is a single record, so several points stacked at the same timestamp indicate duplicate records at that timestamp. The real trajectory has exactly one point at every grid position, whereas the synthetic trajectory stacks multiple points at some timestamps and leaves other grid positions empty.}
    \label{fig:case_duplicate_missing}
\end{figure}

\paragraph{Case 1: Duplicate timestamps and missing grid points.}
Figure~\ref{fig:case_duplicate_missing} shows a case in which the real trajectory fully covers the temporal grid, whereas the synthetic trajectory contains both duplicate timestamps and missing grid points. Although the synthetic data are generated within the valid temporal range, their records are unevenly distributed across the temporal grid. Several records are assigned to the same timestamp, while some grid positions observed in the real trajectory are not generated. This indicates that the model captures the temporal support of the data but does not fully preserve the grid coverage structure. The case illustrates that timestamp validity should be evaluated together with timestamp uniqueness and temporal grid coverage.

\begin{figure}[htbp]
    \centering
    \includegraphics[width=\columnwidth]{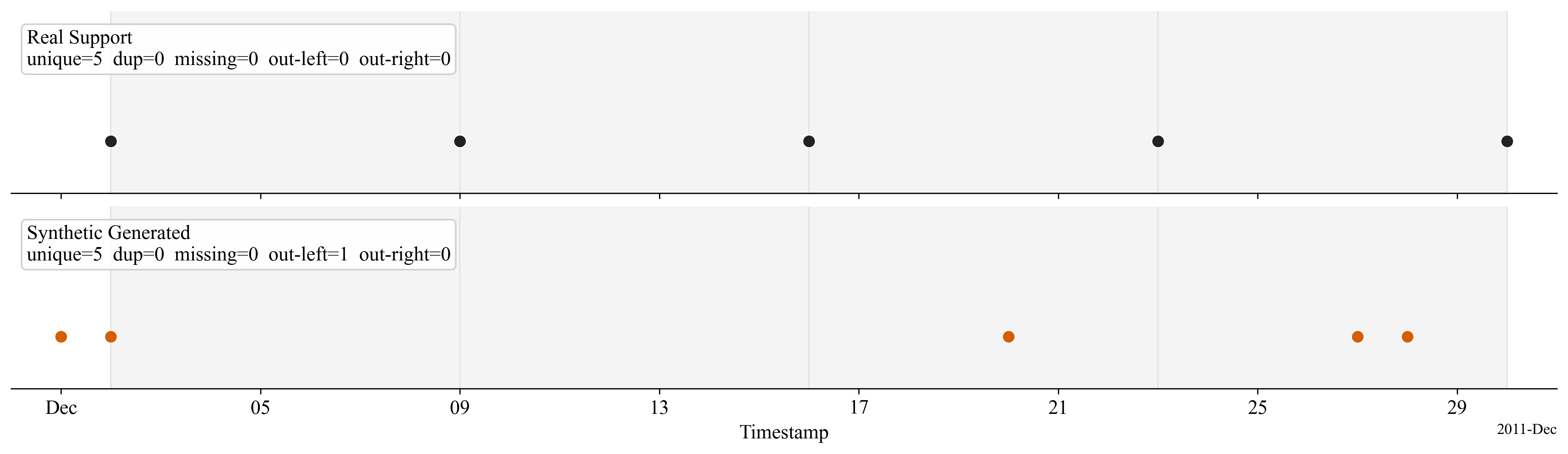}
    \caption{Case study of out-of-range timestamp generation. Each point corresponds to one record, so the number of points at each timestamp represents the record count at that timestamp.}
    \Description{A scatter plot comparing one real trajectory with one synthetic trajectory along the time axis, where each point is a single record. The real records all fall inside the observed temporal support, while the synthetic trajectory places at least one record beyond the right-hand boundary of that support, outside the range covered by the real data.}
    \label{fig:case_out_of_range}
\end{figure}

\paragraph{Case 2: Out-of-range timestamp generation.}
Figure~\ref{fig:case_out_of_range} presents a case in which the synthetic trajectory contains a timestamp outside the temporal support observed in the real data. Although the synthetic trajectory has a comparable number of records to the real trajectory, one generated timestamp appears before the valid observation window. This suggests that the model captures the overall event frequency to some extent, but does not fully preserve the temporal boundary of the data. The case illustrates a discrepancy between sequence level plausibility and timestamp level validity, showing that a realistic trajectory length alone does not guarantee temporally valid synthetic records.

\begin{figure}[htbp]
    \centering
    \includegraphics[width=\columnwidth]{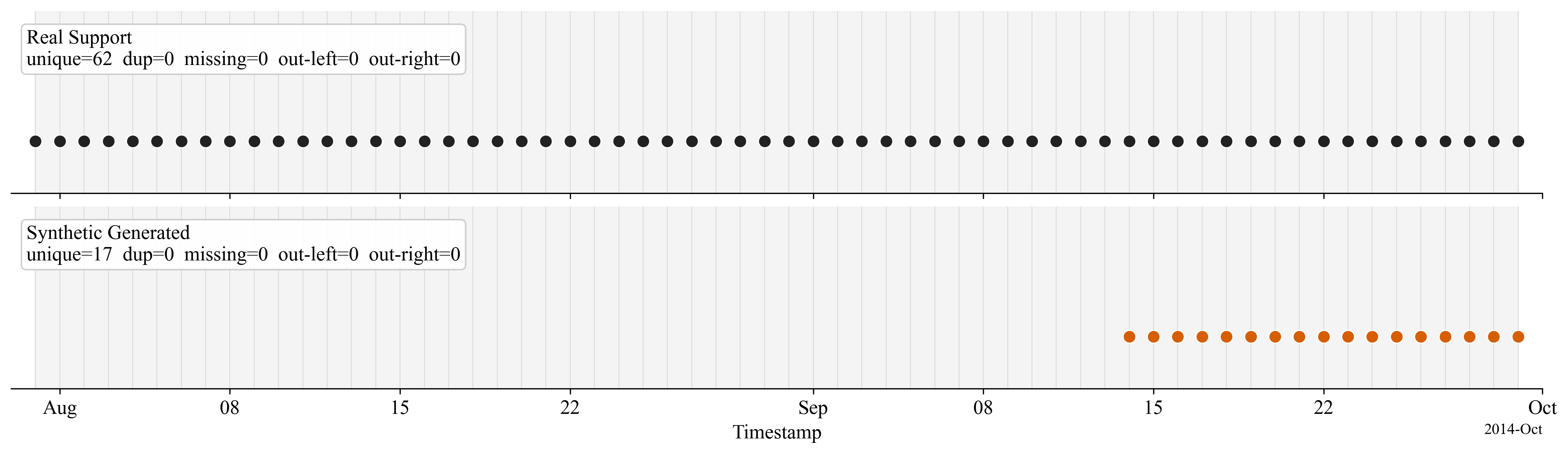}
    \caption{Case study of temporal span truncation. Each point represents a generated record, and the number of points at each timestamp indicates the number of records assigned to that timestamp.}
    \Description{A scatter plot comparing one real trajectory with one synthetic trajectory along the time axis, where each point is a single record. The real records span the full observed temporal range, while the synthetic records are concentrated within a short subinterval, so the synthetic trajectory covers only a fraction of the real temporal span even though its timestamps are unique and in range.}
    \label{fig:case_span_truncation}
\end{figure}

\paragraph{Case 3: Temporal span truncation.}
Figure~\ref{fig:case_span_truncation} shows a case where the synthetic trajectory covers only a short subinterval of the real temporal support. 
Although the generated timestamps are unique and remain within the valid range, the synthetic trajectory fails to reproduce the full temporal span of the real trajectory. This failure mode is especially important because it may not appear as a boundary or duplication error, but still indicates poor preservation of trajectory duration and temporal coverage.

%


\begin{table*}[p]
\centering
\caption{Statistical-distribution sub-metric results across datasets and models. Bold denotes the best score for each dataset--metric column, while underline denotes the second-best score only when the best score is unique.}
\Description{Statistical-distribution sub-metric table comparing synthetic data generators across datasets. Rows correspond to dataset and model pairs, and columns report marginal distribution, correlation, support coverage, and contingency-structure similarities. Higher values indicate better fidelity.}
\label{tab:submetric-sd}
\scriptsize
\setlength{\tabcolsep}{3.0pt}
\renewcommand{\arraystretch}{0.9}
\begin{adjustbox}{max width=\linewidth}
\begin{tabular}{@{}llcccccccc@{}}
\toprule
\textbf{Dataset} & \textbf{Model} & \makecell{\texttt{KS}\\\texttt{Complement}} & \makecell{\texttt{TV}\\\texttt{Complement}} & \makecell{\texttt{Statistic}\\\texttt{Similarity}} & \makecell{\texttt{Correlation}\\\texttt{Similarity}} & \makecell{\texttt{Range}\\\texttt{Coverage}} & \makecell{\texttt{Category}\\\texttt{Coverage}} & \makecell{\texttt{Contingency}\\\texttt{Similarity}} & \makecell{\texttt{Statistic}\\\texttt{MSAS}} \\
\midrule
\multirow{8}{*}{\textbf{Rossmann}} & ClavaDDPM & \underline{0.812} & 0.901 & 0.824 & 0.946 & \underline{0.998} & $\mathbf{0.972}$ & \underline{0.662} & 0.686 \\
 & RCTGAN & 0.755 & 0.838 & 0.826 & \underline{0.956} & 0.996 & $\mathbf{0.972}$ & 0.605 & 0.533 \\
 & RDBDiff & 0.810 & 0.905 & 0.823 & 0.948 & \underline{0.998} & $\mathbf{0.972}$ & 0.660 & 0.744 \\
 & RGCLD & 0.794 & \underline{0.914} & \underline{0.840} & 0.943 & 0.995 & 0.824 & 0.497 & \underline{0.762} \\
 & RelDiff & 0.770 & 0.908 & 0.769 & 0.950 & 0.987 & 0.861 & 0.492 & 0.657 \\
 & SDV & 0.771 & 0.620 & 0.828 & 0.945 & \underline{0.998} & 0.676 & 0.352 & 0.685 \\
 & REaLTabFormer & 0.769 & 0.829 & 0.799 & \underline{0.956} & \underline{0.998} & 0.861 & 0.591 & 0.669 \\
 & TabDiT & $\mathbf{0.847}$ & $\mathbf{0.948}$ & $\mathbf{0.847}$ & $\mathbf{0.964}$ & $\mathbf{1.000}$ & $\mathbf{0.972}$ & $\mathbf{0.881}$ & $\mathbf{1.000}$ \\
\midrule
\multirow{8}{*}{\textbf{Berka}} & ClavaDDPM & \underline{0.972} & \underline{0.978} & \underline{0.993} & 0.982 & $\mathbf{1.000}$ & $\mathbf{1.000}$ & \underline{0.960} & 0.701 \\
 & RCTGAN & 0.871 & 0.808 & 0.925 & 0.963 & $\mathbf{1.000}$ & $\mathbf{1.000}$ & 0.727 & 0.628 \\
 & RDBDiff & $\mathbf{0.980}$ & $\mathbf{0.998}$ & $\mathbf{0.999}$ & $\mathbf{0.998}$ & $\mathbf{1.000}$ & $\mathbf{1.000}$ & $\mathbf{0.992}$ & $\mathbf{0.967}$ \\
 & RGCLD & 0.944 & 0.864 & 0.977 & \underline{0.990} & $\mathbf{1.000}$ & 0.952 & 0.725 & \underline{0.924} \\
 & RelDiff & 0.933 & 0.864 & 0.957 & 0.982 & $\mathbf{1.000}$ & 0.954 & 0.724 & 0.771 \\
 & SDV & 0.631 & 0.694 & 0.650 & 0.958 & $\mathbf{1.000}$ & 0.887 & 0.423 & 0.322 \\
 & REaLTabFormer & 0.941 & 0.959 & 0.955 & 0.975 & $\mathbf{1.000}$ & $\mathbf{1.000}$ & 0.934 & 0.846 \\
 & TabDiT & 0.957 & 0.816 & 0.970 & 0.981 & $\mathbf{1.000}$ & 0.954 & 0.674 & 0.917 \\
\midrule
\multirow{8}{*}{\textbf{Airbnb}} & ClavaDDPM & \underline{0.962} & $\mathbf{0.977}$ & \underline{0.968} & \underline{0.990} & $\mathbf{1.000}$ & 0.979 & $\mathbf{0.946}$ & $\mathbf{0.913}$ \\
 & RCTGAN & 0.878 & 0.862 & 0.894 & 0.958 & $\mathbf{1.000}$ & $\mathbf{0.990}$ & 0.798 & 0.892 \\
 & RDBDiff & $\mathbf{0.995}$ & 0.939 & $\mathbf{0.994}$ & $\mathbf{0.999}$ & $\mathbf{1.000}$ & \underline{0.989} & \underline{0.939} & 0.851 \\
 & RGCLD & 0.945 & \underline{0.957} & 0.958 & 0.988 & $\mathbf{1.000}$ & 0.825 & 0.931 & 0.795 \\
 & RelDiff & 0.935 & 0.929 & 0.934 & 0.973 & $\mathbf{1.000}$ & 0.948 & 0.871 & \underline{0.904} \\
 & SDV & 0.835 & 0.389 & 0.929 & 0.983 & $\mathbf{1.000}$ & 0.649 & 0.200 & 0.702 \\
 & REaLTabFormer & 0.896 & 0.897 & 0.871 & 0.988 & $\mathbf{1.000}$ & 0.951 & 0.866 & 0.559 \\
 & TabDiT & 0.888 & 0.943 & 0.949 & 0.979 & $\mathbf{1.000}$ & 0.809 & 0.914 & \underline{0.904} \\
\midrule
\multirow{7}{*}{\textbf{Fannie Mae}} & ClavaDDPM & $\mathbf{0.961}$ & $\mathbf{0.985}$ & $\mathbf{0.845}$ & \underline{0.983} & 0.973 & 0.967 & \underline{0.833} & 0.653 \\
 & RCTGAN & 0.919 & 0.882 & 0.761 & 0.974 & 0.971 & $\mathbf{0.994}$ & 0.762 & 0.617 \\
 & RDBDiff & 0.916 & 0.949 & \underline{0.804} & $\mathbf{0.984}$ & \underline{0.974} & 0.939 & 0.810 & 0.630 \\
 & RGCLD & \underline{0.935} & 0.946 & 0.779 & 0.975 & 0.971 & 0.920 & 0.810 & $\mathbf{0.910}$ \\
 & RelDiff & 0.888 & \underline{0.968} & 0.753 & 0.977 & 0.964 & \underline{0.984} & 0.812 & \underline{0.849} \\
 & SDV & 0.663 & 0.893 & 0.745 & 0.978 & $\mathbf{0.976}$ & 0.897 & 0.729 & 0.630 \\
 & REaLTabFormer & 0.734 & 0.940 & 0.721 & 0.965 & 0.849 & 0.959 & $\mathbf{0.894}$ & 0.767 \\
\midrule
\multirow{7}{*}{\textbf{Walmart}} & ClavaDDPM & \underline{0.936} & 0.929 & 0.973 & 0.984 & 0.996 & 0.583 & 0.138 & 0.828 \\
 & RCTGAN & 0.808 & 0.921 & 0.878 & 0.928 & $\mathbf{1.000}$ & $\mathbf{1.000}$ & 0.797 & 0.560 \\
 & RDBDiff & $\mathbf{0.963}$ & \underline{0.981} & $\mathbf{0.989}$ & $\mathbf{0.990}$ & 0.996 & 0.583 & 0.154 & $\mathbf{0.864}$ \\
 & RGCLD & 0.929 & $\mathbf{0.992}$ & \underline{0.976} & \underline{0.985} & 0.997 & 0.996 & $\mathbf{0.885}$ & \underline{0.843} \\
 & RelDiff & 0.911 & 0.943 & 0.947 & 0.932 & 0.999 & $\mathbf{1.000}$ & \underline{0.812} & 0.643 \\
 & SDV & 0.796 & 0.927 & 0.812 & 0.954 & $\mathbf{1.000}$ & 0.988 & 0.807 & 0.572 \\
 & REaLTabFormer & 0.739 & 0.586 & 0.862 & 0.912 & 0.982 & 0.792 & 0.304 & 0.574 \\
\midrule
\multirow{7}{*}{\textbf{Freddie Mac}} & ClavaDDPM & 0.965 & 0.965 & $\mathbf{0.984}$ & 0.966 & $\mathbf{1.000}$ & $\mathbf{1.000}$ & 0.924 & 0.743 \\
 & RCTGAN & 0.871 & 0.915 & 0.911 & 0.960 & $\mathbf{1.000}$ & 0.986 & 0.855 & 0.556 \\
 & RDBDiff & $\mathbf{0.989}$ & $\mathbf{0.999}$ & \underline{0.978} & $\mathbf{0.999}$ & $\mathbf{1.000}$ & $\mathbf{1.000}$ & $\mathbf{0.968}$ & $\mathbf{0.956}$ \\
 & RGCLD & \underline{0.980} & \underline{0.971} & 0.977 & \underline{0.996} & $\mathbf{1.000}$ & 0.979 & 0.940 & $\mathbf{0.956}$ \\
 & RelDiff & 0.877 & 0.903 & 0.867 & 0.981 & $\mathbf{1.000}$ & 0.998 & 0.848 & 0.882 \\
 & SDV & 0.715 & 0.807 & 0.870 & 0.983 & $\mathbf{1.000}$ & 0.854 & 0.609 & 0.638 \\
 & REaLTabFormer & 0.745 & 0.946 & 0.775 & 0.955 & 0.898 & 0.946 & \underline{0.953} & 0.605 \\
\midrule
\multirow{7}{*}{\textbf{PTB-XL}} & ClavaDDPM & $\mathbf{0.988}$ & \underline{0.946} & $\mathbf{0.557}$ & \underline{0.994} & $\mathbf{1.000}$ & \underline{0.923} & \underline{0.905} & \underline{0.779} \\
 & RCTGAN & 0.599 & 0.460 & 0.298 & 0.898 & $\mathbf{1.000}$ & 0.902 & 0.226 & 0.221 \\
 & RDBDiff & $\mathbf{0.988}$ & $\mathbf{0.999}$ & \underline{0.443} & $\mathbf{0.997}$ & $\mathbf{1.000}$ & $\mathbf{1.000}$ & $\mathbf{0.943}$ & $\mathbf{0.800}$ \\
 & RGCLD & 0.947 & 0.617 & 0.200 & 0.974 & $\mathbf{1.000}$ & 0.697 & 0.334 & 0.767 \\
 & RelDiff & 0.968 & 0.584 & 0.179 & 0.911 & $\mathbf{1.000}$ & 0.724 & 0.326 & 0.663 \\
 & SDV & 0.847 & 0.780 & 0.319 & 0.989 & $\mathbf{1.000}$ & 0.829 & 0.588 & 0.710 \\
 & REaLTabFormer & 0.842 & 0.658 & 0.193 & 0.964 & $\mathbf{1.000}$ & 0.548 & 0.034 & 0.757 \\
\midrule
\multirow{6}{*}{\textbf{Citi Bike}} & ClavaDDPM & $\mathbf{0.993}$ & \underline{0.971} & 0.973 & $\mathbf{0.999}$ & 0.997 & $\mathbf{1.000}$ & $\mathbf{0.999}$ & -- \\
 & RCTGAN & \underline{0.947} & 0.869 & \underline{0.976} & 0.976 & $\mathbf{1.000}$ & $\mathbf{1.000}$ & 0.833 & -- \\
 & SDV & 0.896 & 0.870 & $\mathbf{0.999}$ & \underline{0.993} & $\mathbf{1.000}$ & $\mathbf{1.000}$ & 0.424 & -- \\
 & CPAR & 0.512 & $\mathbf{0.977}$ & 0.485 & 0.954 & 0.500 & $\mathbf{1.000}$ & 0.512 & -- \\
 & REaLTabFormer & 0.734 & 0.957 & 0.695 & \underline{0.993} & 0.750 & 0.999 & 0.964 & -- \\
 & TabularARGN & 0.811 & 0.888 & 0.758 & 0.968 & 0.875 & 0.956 & \underline{0.971} & -- \\
\midrule
\multirow{6}{*}{\textbf{C-MAPSS}} & ClavaDDPM & 0.898 & -- & $\mathbf{0.987}$ & $\mathbf{0.999}$ & 0.969 & -- & -- & -- \\
 & RCTGAN & 0.709 & -- & 0.905 & 0.636 & $\mathbf{1.000}$ & -- & -- & -- \\
 & SDV & 0.758 & -- & 0.964 & 0.942 & $\mathbf{1.000}$ & -- & -- & -- \\
 & CPAR & 0.622 & -- & $\mathbf{0.987}$ & 0.633 & $\mathbf{1.000}$ & -- & -- & -- \\
 & REaLTabFormer & \underline{0.918} & -- & 0.974 & \underline{0.995} & $\mathbf{1.000}$ & -- & -- & -- \\
 & TabularARGN & $\mathbf{0.962}$ & -- & 0.978 & 0.992 & $\mathbf{1.000}$ & -- & -- & -- \\
\midrule
\multirow{6}{*}{\textbf{Coupon}} & ClavaDDPM & \underline{0.892} & 0.854 & 0.899 & 0.922 & $\mathbf{1.000}$ & 0.995 & 0.723 & 0.877 \\
 & RCTGAN & 0.882 & 0.864 & 0.912 & 0.908 & 0.996 & $\mathbf{1.000}$ & 0.764 & 0.826 \\
 & RDBDiff & $\mathbf{0.984}$ & $\mathbf{0.981}$ & $\mathbf{0.987}$ & $\mathbf{0.979}$ & $\mathbf{1.000}$ & $\mathbf{1.000}$ & $\mathbf{0.957}$ & $\mathbf{0.919}$ \\
 & RGCLD & 0.863 & \underline{0.866} & 0.867 & 0.921 & 0.988 & $\mathbf{1.000}$ & \underline{0.806} & 0.597 \\
 & RelDiff & 0.876 & 0.745 & 0.908 & \underline{0.945} & 0.999 & 0.995 & 0.667 & \underline{0.916} \\
 & SDV & 0.873 & 0.715 & \underline{0.920} & 0.919 & $\mathbf{1.000}$ & 0.964 & 0.544 & 0.683 \\
\midrule
\multirow{6}{*}{\textbf{Google Cluster}} & ClavaDDPM & \underline{0.928} & \underline{0.732} & \underline{0.862} & \underline{0.984} & $\mathbf{1.000}$ & \underline{0.848} & $\mathbf{0.879}$ & 0.673 \\
 & RCTGAN & 0.711 & 0.553 & 0.725 & 0.955 & $\mathbf{1.000}$ & 0.824 & 0.453 & 0.500 \\
 & RDBDiff & $\mathbf{0.946}$ & $\mathbf{0.938}$ & $\mathbf{0.974}$ & $\mathbf{0.996}$ & $\mathbf{1.000}$ & $\mathbf{0.952}$ & \underline{0.857} & 0.814 \\
 & RGCLD & 0.755 & 0.520 & 0.620 & 0.765 & 0.996 & 0.740 & 0.435 & \underline{0.816} \\
 & RelDiff & 0.865 & 0.614 & 0.698 & 0.939 & $\mathbf{1.000}$ & 0.708 & 0.604 & $\mathbf{0.823}$ \\
 & SDV & 0.504 & 0.497 & 0.483 & 0.941 & 0.995 & 0.613 & 0.331 & 0.428 \\
\midrule
\multirow{6}{*}{\textbf{H\&M}} & ClavaDDPM & $\mathbf{0.936}$ & $\mathbf{0.865}$ & $\mathbf{0.996}$ & $\mathbf{0.990}$ & $\mathbf{1.000}$ & \underline{0.893} & $\mathbf{0.920}$ & 0.866 \\
 & RCTGAN & 0.531 & 0.430 & 0.772 & 0.982 & $\mathbf{1.000}$ & 0.875 & 0.335 & 0.082 \\
 & RDBDiff & 0.882 & 0.814 & 0.913 & 0.975 & $\mathbf{1.000}$ & $\mathbf{0.896}$ & \underline{0.764} & \underline{0.886} \\
 & RGCLD & \underline{0.904} & \underline{0.823} & \underline{0.958} & 0.972 & $\mathbf{1.000}$ & 0.759 & 0.725 & 0.828 \\
 & RelDiff & 0.827 & 0.800 & 0.951 & $\mathbf{0.990}$ & $\mathbf{1.000}$ & 0.803 & 0.670 & $\mathbf{0.927}$ \\
 & SDV & 0.503 & 0.639 & 0.670 & 0.915 & $\mathbf{1.000}$ & 0.865 & 0.434 & 0.103 \\
\midrule
\multirow{5}{*}{\textbf{Home Credit}} & ClavaDDPM & 0.457 & 0.625 & 0.379 & 0.897 & 0.931 & 0.884 & 0.404 & 0.791 \\
 & RCTGAN & \underline{0.733} & \underline{0.882} & 0.725 & 0.969 & 0.995 & \underline{0.981} & \underline{0.792} & 0.724 \\
 & RDBDiff & $\mathbf{0.740}$ & $\mathbf{0.909}$ & $\mathbf{0.925}$ & $\mathbf{0.995}$ & 0.996 & $\mathbf{0.985}$ & $\mathbf{0.823}$ & \underline{0.795} \\
 & RGCLD & 0.729 & 0.835 & \underline{0.801} & \underline{0.985} & $\mathbf{0.997}$ & 0.924 & 0.695 & $\mathbf{0.914}$ \\
 & RelDiff & 0.728 & 0.819 & 0.682 & 0.975 & $\mathbf{0.997}$ & 0.942 & 0.681 & 0.787 \\
\bottomrule
\end{tabular}
\end{adjustbox}
\end{table*}


\begin{table*}[p]
\centering
\caption{Timestamp sub-metric results across datasets and models. Bold denotes the best score for each dataset--metric column, while underline denotes the second-best score only when the best score is unique.}
\Description{Timestamp sub-metric table comparing synthetic data generators across datasets. Rows correspond to dataset and model pairs, and columns report timestamp validity and temporal-grid quality sub-metrics. Higher values indicate better fidelity.}
\label{tab:submetric-ts}
\scriptsize
\setlength{\tabcolsep}{3.0pt}
\renewcommand{\arraystretch}{0.9}
\begin{adjustbox}{max width=\linewidth}
\begin{tabular}{@{}llccccccc@{}}
\toprule
\textbf{Dataset} & \textbf{Model} & \makecell{\texttt{Temporal Order}\\\texttt{Consistency}} & \makecell{\texttt{Timestamp}\\\texttt{Uniqueness}} & \makecell{\texttt{Trajectory Duration}\\\texttt{Similarity}} & \makecell{\texttt{Temporal Range}\\\texttt{Compliance}} & \makecell{\texttt{Regularity}\\\texttt{Consistency}} & \makecell{\texttt{Grid}\\\texttt{Completeness}} & \makecell{\texttt{Time Interval}\\\texttt{Distribution Similarity}} \\
\midrule
\multirow{8}{*}{\textbf{Rossmann}} & ClavaDDPM & 0.507 & 0.629 & 0.462 & $\mathbf{1.000}$ & 0.402 & 0.644 & -- \\
 & RCTGAN & $\mathbf{0.512}$ & 0.594 & 0.287 & 0.997 & 0.372 & 0.611 & -- \\
 & RDBDiff & \underline{0.510} & 0.628 & 0.466 & $\mathbf{1.000}$ & 0.402 & 0.644 & -- \\
 & RGCLD & 0.507 & 0.619 & 0.435 & $\mathbf{1.000}$ & 0.392 & 0.636 & -- \\
 & RelDiff & 0.509 & 0.616 & 0.507 & 0.990 & 0.369 & 0.618 & -- \\
 & SDV & 0.508 & 0.594 & 0.350 & 0.978 & 0.364 & 0.596 & -- \\
 & REaLTabFormer & 0.001 & $\mathbf{0.999}$ & \underline{0.942} & $\mathbf{1.000}$ & $\mathbf{0.997}$ & $\mathbf{0.999}$ & -- \\
 & TabDiT & 0.023 & \underline{0.978} & $\mathbf{1.000}$ & $\mathbf{1.000}$ & \underline{0.959} & \underline{0.983} & -- \\
\midrule
\multirow{8}{*}{\textbf{Berka}} & ClavaDDPM & 0.500 & $\mathbf{0.914}$ & 0.011 & $\mathbf{1.000}$ & -- & -- & 0.547 \\
 & RCTGAN & 0.503 & \underline{0.894} & 0.116 & 0.950 & -- & -- & 0.637 \\
 & RDBDiff & 0.513 & 0.553 & $\mathbf{0.944}$ & $\mathbf{1.000}$ & -- & -- & 0.785 \\
 & RGCLD & 0.511 & 0.686 & 0.563 & 0.988 & -- & -- & 0.909 \\
 & RelDiff & 0.508 & 0.825 & 0.012 & 0.915 & -- & -- & 0.753 \\
 & SDV & 0.553 & 0.392 & 0.274 & 0.989 & -- & -- & 0.469 \\
 & REaLTabFormer & $\mathbf{0.960}$ & 0.716 & \underline{0.913} & 0.999 & -- & -- & \underline{0.947} \\
 & TabDiT & \underline{0.865} & 0.741 & 0.333 & 0.999 & -- & -- & $\mathbf{0.949}$ \\
\midrule
\multirow{8}{*}{\textbf{Airbnb}} & ClavaDDPM & 0.502 & 0.962 & $\mathbf{0.947}$ & 0.981 & -- & -- & $\mathbf{0.982}$ \\
 & RCTGAN & \underline{0.510} & 0.900 & 0.880 & 0.991 & -- & -- & 0.921 \\
 & RDBDiff & 0.506 & $\mathbf{0.978}$ & 0.811 & 0.951 & -- & -- & 0.876 \\
 & RGCLD & 0.506 & 0.960 & 0.796 & 0.986 & -- & -- & \underline{0.966} \\
 & RelDiff & 0.506 & 0.924 & \underline{0.932} & \underline{0.996} & -- & -- & 0.944 \\
 & SDV & $\mathbf{0.541}$ & 0.826 & 0.905 & 0.949 & -- & -- & 0.834 \\
 & REaLTabFormer & 0.497 & \underline{0.972} & 0.561 & \underline{0.996} & -- & -- & 0.861 \\
 & TabDiT & 0.502 & 0.953 & 0.884 & $\mathbf{0.998}$ & -- & -- & 0.963 \\
\midrule
\multirow{7}{*}{\textbf{Fannie Mae}} & ClavaDDPM & 0.531 & 0.636 & $\mathbf{0.773}$ & $\mathbf{1.000}$ & $\mathbf{0.235}$ & 0.225 & -- \\
 & RCTGAN & 0.509 & 0.891 & 0.635 & 0.950 & 0.011 & 0.302 & -- \\
 & RDBDiff & \underline{0.533} & 0.608 & 0.560 & $\mathbf{1.000}$ & \underline{0.233} & $\mathbf{0.492}$ & -- \\
 & RGCLD & $\mathbf{0.535}$ & 0.604 & 0.638 & 0.999 & 0.223 & 0.021 & -- \\
 & RelDiff & 0.514 & 0.804 & \underline{0.712} & 0.958 & 0.143 & \underline{0.458} & -- \\
 & SDV & 0.451 & \underline{0.978} & 0.094 & $\mathbf{1.000}$ & 0.012 & 0.352 & -- \\
 & REaLTabFormer & 0.511 & $\mathbf{0.999}$ & 0.053 & 0.978 & 0.026 & 0.279 & -- \\
\midrule
\multirow{7}{*}{\textbf{Walmart}} & ClavaDDPM & 0.594 & 0.578 & \underline{0.422} & $\mathbf{1.000}$ & \underline{0.406} & 0.662 & -- \\
 & RCTGAN & 0.467 & $\mathbf{0.939}$ & 0.200 & 0.818 & 0.078 & 0.658 & -- \\
 & RDBDiff & 0.589 & 0.578 & 0.400 & $\mathbf{1.000}$ & $\mathbf{0.417}$ & 0.662 & -- \\
 & RGCLD & 0.550 & 0.667 & 0.400 & $\mathbf{1.000}$ & 0.311 & \underline{0.671} & -- \\
 & RelDiff & \underline{0.611} & 0.622 & $\mathbf{0.622}$ & $\mathbf{1.000}$ & 0.139 & 0.596 & -- \\
 & SDV & 0.483 & \underline{0.856} & 0.000 & $\mathbf{1.000}$ & 0.061 & $\mathbf{0.733}$ & -- \\
 & REaLTabFormer & $\mathbf{0.662}$ & 0.338 & 0.000 & $\mathbf{1.000}$ & 0.331 & 0.400 & -- \\
\midrule
\multirow{7}{*}{\textbf{Freddie Mac}} & ClavaDDPM & $\mathbf{1.000}$ & 0.659 & 0.704 & $\mathbf{1.000}$ & 0.227 & $\mathbf{0.691}$ & -- \\
 & RCTGAN & $\mathbf{1.000}$ & 0.757 & 0.836 & 0.971 & 0.027 & 0.538 & -- \\
 & RDBDiff & 0.543 & 0.619 & \underline{0.903} & $\mathbf{1.000}$ & $\mathbf{0.242}$ & 0.653 & -- \\
 & RGCLD & $\mathbf{1.000}$ & 0.619 & \underline{0.903} & 0.999 & \underline{0.240} & 0.654 & -- \\
 & RelDiff & $\mathbf{1.000}$ & 0.673 & $\mathbf{0.918}$ & $\mathbf{1.000}$ & 0.234 & \underline{0.688} & -- \\
 & SDV & $\mathbf{1.000}$ & \underline{0.975} & 0.851 & $\mathbf{1.000}$ & 0.011 & 0.644 & -- \\
 & REaLTabFormer & 0.485 & $\mathbf{0.996}$ & 0.456 & 0.850 & 0.103 & 0.597 & -- \\
\midrule
\multirow{7}{*}{\textbf{PTB-XL}} & ClavaDDPM & 0.506 & 0.629 & 0.402 & 0.996 & 0.400 & \underline{0.633} & -- \\
 & RCTGAN & \underline{0.624} & 0.112 & 0.076 & 0.991 & 0.055 & 0.121 & -- \\
 & RDBDiff & 0.505 & 0.618 & 0.330 & 0.996 & 0.393 & 0.622 & -- \\
 & RGCLD & 0.509 & 0.561 & 0.204 & 0.995 & 0.370 & 0.565 & -- \\
 & RelDiff & 0.505 & \underline{0.630} & 0.378 & 0.996 & \underline{0.401} & \underline{0.633} & -- \\
 & SDV & 0.507 & 0.591 & \underline{0.995} & $\mathbf{1.000}$ & 0.364 & 0.596 & -- \\
 & REaLTabFormer & $\mathbf{1.000}$ & $\mathbf{1.000}$ & $\mathbf{0.999}$ & $\mathbf{1.000}$ & $\mathbf{0.999}$ & $\mathbf{0.999}$ & -- \\
\midrule
\multirow{6}{*}{\textbf{Citi Bike}} & ClavaDDPM & 0.500 & $\mathbf{1.000}$ & 0.589 & $\mathbf{1.000}$ & -- & -- & \underline{0.759} \\
 & RCTGAN & 0.500 & $\mathbf{1.000}$ & \underline{0.628} & 0.995 & -- & -- & 0.703 \\
 & SDV & 0.500 & $\mathbf{1.000}$ & $\mathbf{0.727}$ & $\mathbf{1.000}$ & -- & -- & 0.682 \\
 & CPAR & \underline{0.789} & $\mathbf{1.000}$ & 0.583 & 0.622 & -- & -- & 0.612 \\
 & REaLTabFormer & 0.500 & $\mathbf{1.000}$ & 0.096 & 0.883 & -- & -- & $\mathbf{0.912}$ \\
 & TabularARGN & $\mathbf{0.997}$ & 0.003 & 0.002 & 0.003 & -- & -- & 0.000 \\
\midrule
\multirow{6}{*}{\textbf{C-MAPSS}} & ClavaDDPM & 0.502 & 0.648 & 0.307 & \underline{0.996} & 0.352 & 0.645 & -- \\
 & RCTGAN & \underline{0.724} & 0.072 & \underline{0.330} & \underline{0.996} & 0.026 & 0.069 & -- \\
 & SDV & 0.502 & \underline{0.671} & 0.087 & \underline{0.996} & 0.354 & \underline{0.660} & -- \\
 & CPAR & $\mathbf{1.000}$ & $\mathbf{1.000}$ & $\mathbf{0.836}$ & \underline{0.996} & $\mathbf{1.000}$ & $\mathbf{0.981}$ & -- \\
 & REaLTabFormer & 0.501 & 0.125 & 0.064 & $\mathbf{1.000}$ & 0.123 & 0.122 & -- \\
 & TabularARGN & 0.502 & 0.646 & 0.310 & \underline{0.996} & \underline{0.369} & 0.647 & -- \\
\midrule
\multirow{6}{*}{\textbf{Coupon}} & ClavaDDPM & 0.501 & 0.967 & 0.859 & $\mathbf{1.000}$ & -- & -- & 0.815 \\
 & RCTGAN & 0.501 & $\mathbf{0.979}$ & 0.851 & 0.961 & -- & -- & 0.792 \\
 & RDBDiff & \underline{0.819} & \underline{0.968} & \underline{0.881} & $\mathbf{1.000}$ & -- & -- & 0.810 \\
 & RGCLD & $\mathbf{0.821}$ & 0.306 & 0.740 & 0.644 & -- & -- & 0.517 \\
 & RelDiff & 0.512 & 0.960 & $\mathbf{0.947}$ & $\mathbf{1.000}$ & -- & -- & \underline{0.822} \\
 & SDV & 0.496 & 0.946 & 0.750 & $\mathbf{1.000}$ & -- & -- & $\mathbf{0.833}$ \\
\midrule
\multirow{6}{*}{\textbf{Google Cluster}} & ClavaDDPM & 0.565 & 0.674 & 0.452 & 0.931 & -- & -- & 0.398 \\
 & RCTGAN & 0.510 & \underline{0.894} & 0.281 & 0.886 & -- & -- & 0.301 \\
 & RDBDiff & 0.566 & 0.667 & 0.550 & $\mathbf{0.938}$ & -- & -- & 0.412 \\
 & RGCLD & $\mathbf{0.753}$ & 0.397 & \underline{0.566} & 0.849 & -- & -- & $\mathbf{0.683}$ \\
 & RelDiff & \underline{0.597} & 0.595 & $\mathbf{0.631}$ & \underline{0.932} & -- & -- & \underline{0.468} \\
 & SDV & 0.484 & $\mathbf{0.949}$ & 0.523 & 0.873 & -- & -- & 0.319 \\
\midrule
\multirow{6}{*}{\textbf{H\&M}} & ClavaDDPM & 0.500 & $\mathbf{0.928}$ & 0.644 & $\mathbf{1.000}$ & -- & -- & 0.325 \\
 & RCTGAN & $\mathbf{0.802}$ & 0.118 & $\mathbf{0.817}$ & 0.996 & -- & -- & $\mathbf{0.813}$ \\
 & RDBDiff & 0.505 & 0.918 & 0.650 & $\mathbf{1.000}$ & -- & -- & 0.335 \\
 & RGCLD & 0.502 & \underline{0.926} & \underline{0.776} & 0.985 & -- & -- & 0.332 \\
 & RelDiff & 0.501 & 0.903 & 0.684 & $\mathbf{1.000}$ & -- & -- & 0.350 \\
 & SDV & \underline{0.791} & 0.030 & 0.309 & $\mathbf{1.000}$ & -- & -- & \underline{0.777} \\
\midrule
\multirow{5}{*}{\textbf{Home Credit}} & ClavaDDPM & 0.505 & $\mathbf{0.875}$ & 0.561 & $\mathbf{1.000}$ & 0.293 & 0.502 & 0.661 \\
 & RCTGAN & \underline{0.513} & 0.807 & 0.598 & $\mathbf{1.000}$ & 0.289 & 0.354 & \underline{0.720} \\
 & RDBDiff & $\mathbf{0.525}$ & 0.806 & \underline{0.802} & $\mathbf{1.000}$ & \underline{0.359} & \underline{0.636} & 0.702 \\
 & RGCLD & 0.506 & 0.810 & $\mathbf{0.950}$ & $\mathbf{1.000}$ & $\mathbf{0.399}$ & $\mathbf{0.655}$ & 0.714 \\
 & RelDiff & 0.505 & \underline{0.873} & 0.717 & $\mathbf{1.000}$ & 0.307 & 0.546 & $\mathbf{0.747}$ \\
\bottomrule
\end{tabular}
\end{adjustbox}
\end{table*}


\begin{table*}[p]
\centering
\caption{Cross-sectional sub-metric results across datasets and models. Bold denotes the best score for each dataset--metric column, while underline denotes the second-best score only when the best score is unique.}
\Description{Cross-sectional sub-metric table comparing synthetic data generators across datasets. Rows correspond to dataset and model pairs, and columns report distributional and dependency-preservation sub-metrics computed at each temporal slice. Higher values indicate better fidelity.}
\label{tab:submetric-cs}
\small
\setlength{\tabcolsep}{3.0pt}
\renewcommand{\arraystretch}{0.9}
\begin{adjustbox}{max width=\linewidth}
\begin{tabular}{@{}llccccccc@{}}
\toprule
\textbf{Dataset} & \textbf{Model} & \makecell{\texttt{CS KS}\\\texttt{Complement}} & \makecell{\texttt{CS TV}\\\texttt{Complement}} & \makecell{\texttt{CS Statistic}\\\texttt{Similarity}} & \makecell{\texttt{CS Correlation}\\\texttt{Similarity}} & \makecell{\texttt{CS Range}\\\texttt{Coverage}} & \makecell{\texttt{CS Category}\\\texttt{Coverage}} & \makecell{\texttt{CS Contingency}\\\texttt{Similarity}} \\
\midrule
\multirow{8}{*}{\textbf{Rossmann}} & ClavaDDPM & \underline{0.951} & 0.547 & 0.775 & 0.960 & 0.990 & $\mathbf{0.556}$ & 0.269 \\
 & RCTGAN & 0.874 & 0.543 & 0.723 & 0.961 & 0.929 & $\mathbf{0.556}$ & 0.265 \\
 & RDBDiff & 0.948 & 0.548 & 0.774 & 0.961 & 0.990 & 0.555 & 0.269 \\
 & RGCLD & 0.940 & 0.547 & 0.765 & \underline{0.963} & 0.979 & $\mathbf{0.556}$ & 0.268 \\
 & RelDiff & 0.881 & 0.546 & 0.681 & 0.957 & 0.923 & $\mathbf{0.556}$ & 0.267 \\
 & SDV & 0.565 & 0.000 & 0.727 & 0.939 & 0.721 & 0.000 & 0.000 \\
 & REaLTabFormer & 0.904 & \underline{0.552} & \underline{0.875} & \underline{0.963} & \underline{0.992} & $\mathbf{0.556}$ & \underline{0.275} \\
 & TabDiT & $\mathbf{0.996}$ & $\mathbf{0.553}$ & $\mathbf{0.947}$ & $\mathbf{0.999}$ & $\mathbf{1.000}$ & $\mathbf{0.556}$ & $\mathbf{0.277}$ \\
\midrule
\multirow{8}{*}{\textbf{Berka}} & ClavaDDPM & 0.579 & \underline{0.692} & 0.837 & 0.818 & 0.721 & \underline{0.705} & 0.474 \\
 & RCTGAN & 0.540 & 0.599 & 0.818 & 0.807 & 0.702 & 0.609 & 0.370 \\
 & RDBDiff & $\mathbf{0.742}$ & $\mathbf{0.829}$ & $\mathbf{0.923}$ & $\mathbf{0.880}$ & $\mathbf{0.894}$ & $\mathbf{0.810}$ & $\mathbf{0.723}$ \\
 & RGCLD & \underline{0.672} & 0.599 & \underline{0.871} & 0.824 & \underline{0.788} & 0.562 & 0.473 \\
 & RelDiff & 0.642 & 0.529 & 0.840 & 0.809 & 0.703 & 0.585 & 0.362 \\
 & SDV & 0.429 & 0.388 & 0.720 & \underline{0.855} & 0.534 & 0.064 & 0.193 \\
 & REaLTabFormer & 0.609 & 0.644 & 0.818 & 0.809 & 0.717 & 0.630 & \underline{0.565} \\
 & TabDiT & 0.655 & 0.567 & 0.862 & 0.808 & 0.767 & 0.564 & 0.442 \\
\midrule
\multirow{7}{*}{\textbf{Fannie Mae}} & ClavaDDPM & $\mathbf{0.993}$ & -0.000 & $\mathbf{0.868}$ & $\mathbf{0.997}$ & $\mathbf{1.000}$ & 0.000 & -0.000 \\
 & RCTGAN & 0.895 & -0.000 & 0.451 & 0.980 & 0.998 & 0.000 & -0.000 \\
 & RDBDiff & 0.901 & 0.000 & 0.442 & \underline{0.984} & 0.998 & 0.000 & -0.000 \\
 & RGCLD & \underline{0.952} & $\mathbf{0.943}$ & \underline{0.476} & 0.977 & \underline{0.999} & $\mathbf{0.905}$ & $\mathbf{0.881}$ \\
 & RelDiff & 0.882 & 0.000 & 0.291 & 0.982 & 0.972 & 0.000 & 0.000 \\
 & SDV & 0.646 & 0.000 & 0.274 & 0.983 & 0.994 & 0.000 & -0.000 \\
 & REaLTabFormer & 0.828 & \underline{0.922} & 0.392 & 0.978 & 0.957 & \underline{0.505} & \underline{0.855} \\
\midrule
\multirow{7}{*}{\textbf{Walmart}} & ClavaDDPM & \underline{0.843} & 0.483 & \underline{0.580} & 0.912 & $\mathbf{0.968}$ & 0.500 & 0.000 \\
 & RCTGAN & 0.708 & 0.930 & 0.435 & 0.838 & 0.840 & $\mathbf{1.000}$ & 0.864 \\
 & RDBDiff & $\mathbf{0.852}$ & 0.478 & $\mathbf{0.604}$ & $\mathbf{0.921}$ & \underline{0.966} & 0.500 & 0.000 \\
 & RGCLD & 0.836 & $\mathbf{0.984}$ & 0.574 & \underline{0.917} & \underline{0.966} & $\mathbf{1.000}$ & $\mathbf{0.943}$ \\
 & RelDiff & 0.748 & \underline{0.978} & 0.456 & 0.838 & 0.883 & $\mathbf{1.000}$ & \underline{0.923} \\
 & SDV & 0.632 & 0.706 & 0.399 & 0.889 & 0.850 & 0.500 & -- \\
 & REaLTabFormer & 0.688 & 0.742 & 0.461 & 0.813 & 0.878 & 0.300 & 0.493 \\
\midrule
\multirow{7}{*}{\textbf{Freddie Mac}} & ClavaDDPM & 0.972 & 0.959 & 0.679 & 0.963 & $\mathbf{1.000}$ & 0.990 & 0.907 \\
 & RCTGAN & 0.855 & 0.905 & 0.418 & 0.951 & 0.985 & 0.976 & 0.813 \\
 & RDBDiff & $\mathbf{0.987}$ & $\mathbf{0.993}$ & $\mathbf{0.920}$ & $\mathbf{0.997}$ & $\mathbf{1.000}$ & $\mathbf{0.994}$ & $\mathbf{0.982}$ \\
 & RGCLD & \underline{0.976} & \underline{0.966} & \underline{0.686} & \underline{0.994} & $\mathbf{1.000}$ & 0.975 & \underline{0.912} \\
 & RelDiff & 0.869 & 0.901 & 0.366 & 0.980 & 0.996 & \underline{0.993} & 0.826 \\
 & SDV & 0.673 & 0.803 & 0.275 & 0.981 & 0.964 & 0.863 & 0.623 \\
 & REaLTabFormer & 0.835 & 0.945 & 0.423 & 0.960 & 0.995 & 0.384 & 0.900 \\
\midrule
\multirow{6}{*}{\textbf{Citi Bike}} & ClavaDDPM & \underline{0.957} & \underline{0.880} & \underline{0.895} & \underline{0.991} & 0.996 & 0.964 & \underline{0.902} \\
 & RCTGAN & 0.906 & 0.806 & 0.837 & 0.970 & $\mathbf{1.000}$ & $\mathbf{0.967}$ & 0.813 \\
 & SDV & 0.900 & 0.793 & 0.871 & 0.987 & $\mathbf{1.000}$ & $\mathbf{0.967}$ & 0.415 \\
 & CPAR & 0.436 & 0.863 & 0.236 & 0.953 & 0.332 & $\mathbf{0.967}$ & 0.509 \\
 & REaLTabFormer & $\mathbf{0.968}$ & $\mathbf{0.921}$ & $\mathbf{0.950}$ & $\mathbf{0.992}$ & 0.999 & 0.416 & $\mathbf{0.946}$ \\
 & TabularARGN & 0.853 & 0.535 & 0.758 & 0.930 & 0.999 & 0.396 & 0.829 \\
\midrule
\multirow{6}{*}{\textbf{Coupon}} & ClavaDDPM & $\mathbf{0.753}$ & $\mathbf{0.743}$ & $\mathbf{0.832}$ & \underline{0.908} & \underline{0.968} & \underline{0.878} & $\mathbf{0.528}$ \\
 & RCTGAN & \underline{0.751} & \underline{0.737} & \underline{0.822} & \underline{0.908} & 0.965 & $\mathbf{0.881}$ & \underline{0.523} \\
 & RDBDiff & 0.749 & 0.717 & 0.811 & $\mathbf{0.909}$ & $\mathbf{0.972}$ & 0.870 & 0.498 \\
 & RGCLD & 0.705 & 0.675 & 0.783 & 0.879 & 0.965 & 0.716 & 0.436 \\
 & RelDiff & 0.740 & 0.695 & 0.807 & 0.905 & 0.967 & 0.862 & 0.476 \\
 & SDV & 0.410 & 0.410 & 0.514 & 0.562 & 0.960 & 0.320 & 0.143 \\
\midrule
\multirow{6}{*}{\textbf{H\&M}} & ClavaDDPM & $\mathbf{0.871}$ & $\mathbf{0.766}$ & $\mathbf{0.824}$ & 0.968 & $\mathbf{0.998}$ & $\mathbf{0.740}$ & $\mathbf{0.571}$ \\
 & RCTGAN & 0.589 & 0.357 & 0.371 & 0.873 & 0.964 & 0.359 & 0.137 \\
 & RDBDiff & 0.808 & 0.732 & 0.485 & \underline{0.969} & 0.952 & 0.671 & 0.513 \\
 & RGCLD & \underline{0.849} & \underline{0.754} & \underline{0.803} & 0.966 & $\mathbf{0.998}$ & 0.697 & \underline{0.547} \\
 & RelDiff & 0.810 & 0.725 & 0.716 & 0.967 & 0.993 & \underline{0.699} & 0.510 \\
 & SDV & 0.313 & 0.342 & 0.157 & $\mathbf{0.978}$ & 0.493 & 0.003 & 0.141 \\
\bottomrule
\end{tabular}
\end{adjustbox}
\end{table*}


\begin{table*}[p]
\centering
\caption{Longitudinal sub-metric results across datasets and models. TT-Wasserstein is computed as $1-\mathrm{TT\text{-}W}$ so that higher values indicate better fidelity; it is reported only for Rossmann, Walmart, and PTB-XL. Bold denotes the best score for each dataset--metric column, while underline denotes the second-best score only when the best score is unique.}
\Description{Longitudinal sub-metric table comparing synthetic data generators across datasets. Rows correspond to dataset and model pairs, and columns report trajectory-level dynamic similarity metrics. TT-Wasserstein is reported as one minus the original TT-W distance and is shown only for Rossmann, Walmart, and PTB-XL; inapplicable datasets are marked with dashes.}
\label{tab:submetric-long}
\scriptsize
\setlength{\tabcolsep}{3.0pt}
\renewcommand{\arraystretch}{0.9}
\begin{adjustbox}{max width=\linewidth}
\begin{tabular}{@{}llcccc@{}}
\toprule
\textbf{Dataset} & \textbf{Model} & \makecell{\texttt{First Difference}\\\texttt{KS Complement}} & \makecell{\texttt{Transition Matrix}\\\texttt{TV Complement}} & \makecell{\texttt{AutoCorrelation}\\\texttt{Similarity}} & \makecell{\texttt{TT-Wasserstein}} \\
\midrule
\multirow{8}{*}{\textbf{Rossmann}} & ClavaDDPM & 0.975 & 0.958 & 0.928 & 0.905 \\
 & RCTGAN & 0.971 & 0.918 & 0.872 & 0.880 \\
 & RDBDiff & 0.945 & 0.954 & 0.926 & 0.921 \\
 & RGCLD & 0.960 & 0.947 & 0.924 & 0.899 \\
 & RelDiff & 0.966 & 0.920 & 0.898 & 0.880 \\
 & SDV & 0.900 & 0.868 & 0.675 & 0.880 \\
 & REaLTabFormer & \underline{0.978} & $\mathbf{0.981}$ & \underline{0.941} & \underline{0.946} \\
 & TabDiT & $\mathbf{0.998}$ & \underline{0.980} & $\mathbf{0.954}$ & $\mathbf{0.991}$ \\
\midrule
\multirow{8}{*}{\textbf{Berka}} & ClavaDDPM & 0.941 & 0.829 & 0.860 & -- \\
 & RCTGAN & 0.937 & 0.806 & 0.832 & -- \\
 & RDBDiff & 0.956 & $\mathbf{0.913}$ & $\mathbf{0.980}$ & -- \\
 & RGCLD & 0.945 & 0.823 & 0.884 & -- \\
 & RelDiff & 0.936 & 0.815 & 0.852 & -- \\
 & SDV & 0.482 & 0.572 & \underline{0.925} & -- \\
 & REaLTabFormer & \underline{0.965} & \underline{0.845} & 0.877 & -- \\
 & TabDiT & $\mathbf{0.969}$ & 0.827 & 0.871 & -- \\
\midrule
\multirow{8}{*}{\textbf{Airbnb}} & ClavaDDPM & $\mathbf{1.000}$ & 0.908 & 0.995 & -- \\
 & RCTGAN & $\mathbf{1.000}$ & 0.909 & 0.995 & -- \\
 & RDBDiff & $\mathbf{1.000}$ & 0.900 & $\mathbf{0.996}$ & -- \\
 & RGCLD & $\mathbf{1.000}$ & \underline{0.910} & 0.994 & -- \\
 & RelDiff & $\mathbf{1.000}$ & 0.908 & $\mathbf{0.996}$ & -- \\
 & SDV & -- & $\mathbf{0.939}$ & -- & -- \\
 & REaLTabFormer & $\mathbf{1.000}$ & 0.899 & 0.994 & -- \\
 & TabDiT & $\mathbf{1.000}$ & 0.903 & 0.994 & -- \\
\midrule
\multirow{7}{*}{\textbf{Fannie Mae}} & ClavaDDPM & 0.622 & 0.796 & $\mathbf{0.980}$ & -- \\
 & RCTGAN & 0.608 & 0.786 & 0.914 & -- \\
 & RDBDiff & 0.627 & 0.797 & 0.857 & -- \\
 & RGCLD & \underline{0.747} & $\mathbf{0.863}$ & 0.891 & -- \\
 & RelDiff & 0.646 & 0.812 & 0.909 & -- \\
 & SDV & 0.738 & 0.804 & \underline{0.932} & -- \\
 & REaLTabFormer & $\mathbf{0.789}$ & \underline{0.838} & 0.693 & -- \\
\midrule
\multirow{7}{*}{\textbf{Walmart}} & ClavaDDPM & \underline{0.883} & \underline{0.922} & $\mathbf{0.938}$ & 0.783 \\
 & RCTGAN & 0.862 & 0.900 & 0.837 & 0.772 \\
 & RDBDiff & 0.873 & 0.921 & \underline{0.860} & 0.792 \\
 & RGCLD & $\mathbf{0.891}$ & $\mathbf{0.927}$ & 0.853 & $\mathbf{0.797}$ \\
 & RelDiff & 0.876 & 0.918 & 0.798 & 0.786 \\
 & SDV & 0.860 & 0.610 & 0.800 & \underline{0.796} \\
 & REaLTabFormer & 0.852 & 0.744 & 0.437 & 0.729 \\
\midrule
\multirow{7}{*}{\textbf{Freddie Mac}} & ClavaDDPM & 0.584 & $\mathbf{0.849}$ & \underline{0.971} & -- \\
 & RCTGAN & 0.544 & 0.832 & 0.900 & -- \\
 & RDBDiff & $\mathbf{0.791}$ & 0.838 & $\mathbf{0.977}$ & -- \\
 & RGCLD & \underline{0.716} & 0.835 & 0.939 & -- \\
 & RelDiff & 0.651 & $\mathbf{0.849}$ & 0.905 & -- \\
 & SDV & 0.565 & 0.779 & 0.840 & -- \\
 & REaLTabFormer & 0.697 & 0.486 & 0.712 & -- \\
\midrule
\multirow{7}{*}{\textbf{PTB-XL}} & ClavaDDPM & 0.967 & 0.737 & \underline{0.848} & $\mathbf{0.965}$ \\
 & RCTGAN & 0.765 & 0.747 & 0.808 & 0.951 \\
 & RDBDiff & 0.918 & 0.771 & 0.842 & $\mathbf{0.965}$ \\
 & RGCLD & 0.934 & 0.772 & 0.819 & $\mathbf{0.965}$ \\
 & RelDiff & \underline{0.968} & 0.763 & 0.836 & 0.961 \\
 & SDV & 0.946 & $\mathbf{0.971}$ & $\mathbf{0.850}$ & 0.963 \\
 & REaLTabFormer & $\mathbf{0.982}$ & \underline{0.774} & 0.798 & $\mathbf{0.965}$ \\
\midrule
\multirow{6}{*}{\textbf{Citi Bike}} & ClavaDDPM & \underline{0.935} & -- & \underline{0.995} & -- \\
 & RCTGAN & 0.928 & -- & 0.992 & -- \\
 & SDV & 0.906 & -- & 0.994 & -- \\
 & CPAR & 0.628 & -- & 0.994 & -- \\
 & REaLTabFormer & $\mathbf{0.937}$ & -- & $\mathbf{0.997}$ & -- \\
 & TabularARGN & 0.876 & -- & 0.966 & -- \\
\midrule
\multirow{6}{*}{\textbf{C-MAPSS}} & ClavaDDPM & \underline{0.897} & -- & 0.902 & -- \\
 & RCTGAN & 0.760 & -- & \underline{0.907} & -- \\
 & SDV & 0.859 & -- & $\mathbf{0.924}$ & -- \\
 & CPAR & 0.809 & -- & 0.893 & -- \\
 & REaLTabFormer & $\mathbf{0.973}$ & -- & 0.853 & -- \\
 & TabularARGN & \underline{0.897} & -- & 0.780 & -- \\
\midrule
\multirow{6}{*}{\textbf{Coupon}} & ClavaDDPM & 0.889 & 0.852 & 0.787 & -- \\
 & RCTGAN & 0.879 & 0.853 & 0.755 & -- \\
 & RDBDiff & 0.874 & 0.845 & $\mathbf{0.933}$ & -- \\
 & RGCLD & $\mathbf{0.935}$ & $\mathbf{0.891}$ & 0.847 & -- \\
 & RelDiff & \underline{0.927} & 0.865 & \underline{0.932} & -- \\
 & SDV & 0.806 & \underline{0.880} & 0.825 & -- \\
\midrule
\multirow{6}{*}{\textbf{Google Cluster}} & ClavaDDPM & 0.721 & -- & 0.789 & -- \\
 & RCTGAN & 0.620 & -- & 0.795 & -- \\
 & RDBDiff & \underline{0.801} & -- & 0.786 & -- \\
 & RGCLD & $\mathbf{0.802}$ & -- & \underline{0.797} & -- \\
 & RelDiff & 0.668 & -- & 0.775 & -- \\
 & SDV & 0.661 & -- & $\mathbf{0.823}$ & -- \\
\midrule
\multirow{6}{*}{\textbf{H\&M}} & ClavaDDPM & 0.893 & $\mathbf{0.822}$ & \underline{0.966} & -- \\
 & RCTGAN & 0.782 & 0.565 & 0.598 & -- \\
 & RDBDiff & \underline{0.911} & 0.680 & 0.590 & -- \\
 & RGCLD & $\mathbf{0.925}$ & \underline{0.809} & $\mathbf{0.975}$ & -- \\
 & RelDiff & 0.891 & 0.759 & 0.694 & -- \\
 & SDV & 0.778 & -- & 0.582 & -- \\
\midrule
\multirow{5}{*}{\textbf{Home Credit}} & ClavaDDPM & \underline{0.940} & \underline{0.895} & 0.935 & -- \\
 & RCTGAN & 0.881 & 0.844 & 0.899 & -- \\
 & RDBDiff & 0.939 & 0.882 & \underline{0.945} & -- \\
 & RGCLD & $\mathbf{0.958}$ & $\mathbf{0.900}$ & $\mathbf{0.949}$ & -- \\
 & RelDiff & 0.935 & 0.883 & 0.898 & -- \\
\bottomrule
\end{tabular}
\end{adjustbox}
\end{table*}


\begin{table*}[p]
\centering
\caption{Structural sub-metric results across datasets and models. Bold denotes the best score for each dataset--metric column, while underline denotes the second-best score only when the best score is unique.}
\Description{Structural sub-metric table comparing synthetic data generators across datasets. Rows correspond to dataset and model pairs, and columns report sequence-length, cardinality-shape, and dynamic relational-structure similarities. Higher values indicate better fidelity.}
\label{tab:submetric-struct}
\scriptsize
\setlength{\tabcolsep}{3.0pt}
\renewcommand{\arraystretch}{0.9}
\begin{adjustbox}{max width=\linewidth}
\begin{tabular}{@{}llccc@{}}
\toprule
\textbf{Dataset} & \textbf{Model} & \makecell{\texttt{Sequence Length}\\\texttt{Similarity}} & \makecell{\texttt{Temporal Cardinality}\\\texttt{Shape Similarity}} & \makecell{\texttt{Dynamic K-Hop}\\\texttt{Correlation Similarity}} \\
\midrule
\multirow{8}{*}{\textbf{Rossmann}} & ClavaDDPM & $\mathbf{1.000}$ & -- & -- \\
 & RCTGAN & $\mathbf{1.000}$ & -- & -- \\
 & RDBDiff & $\mathbf{1.000}$ & -- & -- \\
 & RGCLD & $\mathbf{1.000}$ & -- & -- \\
 & RelDiff & $\mathbf{1.000}$ & -- & -- \\
 & SDV & $\mathbf{1.000}$ & -- & -- \\
 & REaLTabFormer & $\mathbf{1.000}$ & -- & -- \\
 & TabDiT & $\mathbf{1.000}$ & -- & -- \\
\midrule
\multirow{8}{*}{\textbf{Berka}} & ClavaDDPM & 0.993 & -- & -- \\
 & RCTGAN & 0.952 & -- & -- \\
 & RDBDiff & $\mathbf{1.000}$ & -- & -- \\
 & RGCLD & 0.993 & -- & -- \\
 & RelDiff & $\mathbf{1.000}$ & -- & -- \\
 & SDV & 0.814 & -- & -- \\
 & REaLTabFormer & 0.952 & -- & -- \\
 & TabDiT & 0.958 & -- & -- \\
\midrule
\multirow{8}{*}{\textbf{Airbnb}} & ClavaDDPM & 0.987 & -- & -- \\
 & RCTGAN & 0.924 & -- & -- \\
 & RDBDiff & $\mathbf{1.000}$ & -- & -- \\
 & RGCLD & 0.981 & -- & -- \\
 & RelDiff & $\mathbf{1.000}$ & -- & -- \\
 & SDV & 0.938 & -- & -- \\
 & REaLTabFormer & 0.631 & -- & -- \\
 & TabDiT & 0.912 & -- & -- \\
\midrule
\multirow{7}{*}{\textbf{Fannie Mae}} & ClavaDDPM & $\mathbf{0.998}$ & -- & -- \\
 & RCTGAN & 0.953 & -- & -- \\
 & RDBDiff & 0.822 & -- & -- \\
 & RGCLD & \underline{0.990} & -- & -- \\
 & RelDiff & 0.750 & -- & -- \\
 & SDV & 0.848 & -- & -- \\
 & REaLTabFormer & 0.817 & -- & -- \\
\midrule
\multirow{7}{*}{\textbf{Walmart}} & ClavaDDPM & 0.956 & 0.704 & 0.931 \\
 & RCTGAN & 0.844 & 0.547 & 0.801 \\
 & RDBDiff & $\mathbf{1.000}$ & \underline{0.709} & $\mathbf{0.953}$ \\
 & RGCLD & 0.967 & $\mathbf{0.729}$ & \underline{0.950} \\
 & RelDiff & $\mathbf{1.000}$ & 0.591 & 0.822 \\
 & SDV & 0.878 & 0.598 & 0.881 \\
 & REaLTabFormer & 0.322 & 0.420 & 0.811 \\
\midrule
\multirow{7}{*}{\textbf{Freddie Mac}} & ClavaDDPM & 0.995 & -- & -- \\
 & RCTGAN & 0.972 & -- & -- \\
 & RDBDiff & $\mathbf{1.000}$ & -- & -- \\
 & RGCLD & 0.993 & -- & -- \\
 & RelDiff & $\mathbf{1.000}$ & -- & -- \\
 & SDV & 0.965 & -- & -- \\
 & REaLTabFormer & 0.695 & -- & -- \\
\midrule
\multirow{7}{*}{\textbf{PTB-XL}} & ClavaDDPM & $\mathbf{1.000}$ & -- & -- \\
 & RCTGAN & $\mathbf{1.000}$ & -- & -- \\
 & RDBDiff & $\mathbf{1.000}$ & -- & -- \\
 & RGCLD & $\mathbf{1.000}$ & -- & -- \\
 & RelDiff & $\mathbf{1.000}$ & -- & -- \\
 & SDV & $\mathbf{1.000}$ & -- & -- \\
 & REaLTabFormer & 0.999 & -- & -- \\
\midrule
\multirow{6}{*}{\textbf{Citi Bike}} & ClavaDDPM & 0.908 & -- & -- \\
 & RCTGAN & $\mathbf{0.979}$ & -- & -- \\
 & SDV & $\mathbf{0.979}$ & -- & -- \\
 & CPAR & 0.271 & -- & -- \\
 & REaLTabFormer & 0.676 & -- & -- \\
 & TabularARGN & 0.154 & -- & -- \\
\midrule
\multirow{6}{*}{\textbf{C-MAPSS}} & ClavaDDPM & 0.709 & -- & -- \\
 & RCTGAN & 0.330 & -- & -- \\
 & SDV & $\mathbf{0.961}$ & -- & -- \\
 & CPAR & 0.836 & -- & -- \\
 & REaLTabFormer & 0.000 & -- & -- \\
 & TabularARGN & \underline{0.917} & -- & -- \\
\midrule
\multirow{6}{*}{\textbf{Coupon}} & ClavaDDPM & 0.990 & 0.998 & 0.917 \\
 & RCTGAN & 0.834 & 0.998 & 0.898 \\
 & RDBDiff & $\mathbf{1.000}$ & $\mathbf{0.999}$ & $\mathbf{0.956}$ \\
 & RGCLD & $\mathbf{1.000}$ & 0.995 & 0.920 \\
 & RelDiff & $\mathbf{1.000}$ & $\mathbf{0.999}$ & \underline{0.927} \\
 & SDV & 0.669 & $\mathbf{0.999}$ & 0.921 \\
\midrule
\multirow{6}{*}{\textbf{Google Cluster}} & ClavaDDPM & 0.970 & 0.984 & -- \\
 & RCTGAN & 0.691 & 0.725 & -- \\
 & RDBDiff & $\mathbf{1.000}$ & $\mathbf{1.000}$ & -- \\
 & RGCLD & 0.981 & 0.991 & -- \\
 & RelDiff & $\mathbf{1.000}$ & $\mathbf{1.000}$ & -- \\
 & SDV & 0.477 & 0.656 & -- \\
\midrule
\multirow{6}{*}{\textbf{H\&M}} & ClavaDDPM & \underline{0.640} & -- & -- \\
 & RCTGAN & 0.555 & -- & -- \\
 & RDBDiff & 0.638 & -- & -- \\
 & RGCLD & \underline{0.640} & -- & -- \\
 & RelDiff & \underline{0.640} & -- & -- \\
 & SDV & $\mathbf{0.648}$ & -- & -- \\
\midrule
\multirow{5}{*}{\textbf{Home Credit}} & ClavaDDPM & 0.965 & 0.940 & -- \\
 & RCTGAN & 0.842 & 0.898 & -- \\
 & RDBDiff & \underline{0.995} & \underline{0.998} & -- \\
 & RGCLD & 0.991 & 0.997 & -- \\
 & RelDiff & $\mathbf{1.000}$ & $\mathbf{1.000}$ & -- \\
\bottomrule
\end{tabular}
\end{adjustbox}
\end{table*}

\end{document}